\documentclass[
    11pt,
    a4paper,
    twoside,
    openright,
    headinclude,
    footinclude,
    BCOR=10mm,
    DIV=12
]{scrbook}

\usepackage[utf8]{inputenc}
\usepackage[T1]{fontenc}
\usepackage{microtype}

\usepackage{libertinus}
\usepackage[scaled=0.85]{beramono}

\usepackage[english]{babel}
\usepackage{csquotes}

\usepackage{xcolor}

\definecolor{TUMBlue}{HTML}{3070B3}
\definecolor{TUMBlueDark}{HTML}{072140}
\definecolor{TUMBlueLight}{HTML}{5E94D4}

\definecolor{TUMGray}{HTML}{EBECEF}
\definecolor{TUMGrayLight}{HTML}{475058}

\usepackage{geometry}
\usepackage[headsepline]{scrlayer-scrpage}
\clearpairofpagestyles

\automark[chapter]{chapter}
\ohead{\headmark}
\ofoot[\pagemark]{\pagemark}

\setkomafont{pageheadfoot}{\small\color{TUMGrayLight}}
\setkomafont{pagenumber}{\normalfont}

\addtokomafont{headsepline}{\color{TUMBlue}}
\KOMAoptions{headsepline=0.5pt}

\usepackage{titlesec}

\titleformat{\chapter}[display]
    {\normalfont\huge\bfseries}
    {\textcolor{TUMBlue}{\chaptertitlename\ \thechapter}}
    {20pt}
    {\Huge\color{TUMBlueDark}}

\titleformat{name=\chapter,numberless}[display]
    {\normalfont\huge\bfseries}
    {}
    {0pt}
    {\Huge\color{TUMBlueDark}}

\titlespacing*{\chapter}{0pt}{-30pt}{40pt}

\titleformat{\section}
    {\normalfont\Large\bfseries\color{TUMBlueDark}}
    {\thesection}
    {1em}
    {}

\titleformat{\subsection}
    {\normalfont\large\bfseries\color{TUMBlue}}
    {\thesubsection}
    {1em}
    {}

\titleformat{\subsubsection}
    {\normalfont\normalsize\bfseries}
    {\thesubsubsection}
    {1em}
    {}

\usepackage{tocbasic}
\DeclareTOCStyleEntry[
    beforeskip=1em,
    entryformat=\bfseries\color{TUMBlueDark},
    pagenumberformat=\bfseries
]{tocline}{chapter}

\DeclareTOCStyleEntry[
    indent=1.5em,
    entryformat=\color{black}
]{tocline}{section}

\DeclareTOCStyleEntry[
    indent=3.8em,
    entryformat=\color{TUMGray}
]{tocline}{subsection}

\usepackage{graphicx}
\usepackage{float}
\usepackage{subcaption}

\usepackage[
    font=small,
    labelfont={bf,color=TUMBlue},
    format=plain,
    margin=1cm
]{caption}

\usepackage{pdfpages}

\usepackage{booktabs}
\usepackage{array}
\usepackage{tabularx}
\usepackage{multirow}

\usepackage{colortbl}
\newcommand{\tumrule}{\arrayrulecolor{TUMBlue}\midrule\arrayrulecolor{black}}

\usepackage{amsmath}
\usepackage{amssymb}
\usepackage{amsthm}
\usepackage{mathtools}
\usepackage{bm}

\theoremstyle{definition}

\theoremstyle{remark}

\usepackage{listings}

\lstdefinestyle{modern}{
    backgroundcolor=\color{TUMGrayLight!30},
    basicstyle=\ttfamily\small,
    breaklines=true,
    captionpos=b,
    commentstyle=\color{TUMGray},
    keywordstyle=\color{TUMBlue}\bfseries,
    stringstyle=\color{TUMBlueDark},
    numberstyle=\tiny\color{TUMGray},
    numbers=left,
    numbersep=8pt,
    frame=single,
    framerule=0pt,
    rulecolor=\color{TUMBlue},
    xleftmargin=15pt,
    framexleftmargin=15pt,
    tabsize=2
}
\usepackage{hyperref}
\hypersetup{
    colorlinks=true,
    linkcolor=TUMBlue,
    citecolor=TUMBlue,
    urlcolor=TUMBlue,
    bookmarks=true,
    bookmarksnumbered=true,
    pdfauthor={Felix Köhler},
    pdftitle={From Numerical Simulators of PDEs to Neural Emulators and Back},
    pdfsubject={PhD Dissertation},
    pdfkeywords={Neural Networks, PDEs, Numerical Methods, Machine Learning}
}

\usepackage[capitalise,noabbrev,nameinlink]{cleveref}

\usepackage{tikz}
\usetikzlibrary{positioning, arrows.meta, fit, backgrounds, calc, shapes.misc, shapes.geometric, decorations.pathreplacing, decorations.pathmorphing, mindmap, patterns}

\usepackage{pgfplots}
\pgfplotsset{compat=1.18}
\usepgfplotslibrary{fillbetween}

\usepackage[
    backend=biber,
    style=authoryear-comp,
    sorting=nyt,
    maxbibnames=99,
    minbibnames=99,
    maxcitenames=2,
    mincitenames=1,
    giveninits=false,
    natbib=true,
    uniquelist=false,
    uniquename=false,
    doi=false,
    isbn=false,
    eprint=false,
]{biblatex}
\DeclareFieldFormat{citehyperref}{%
  \DeclareFieldAlias{bibhyperref}{noformat}%
  \bibhyperref{#1}}

\DeclareFieldFormat{textcitehyperref}{%
  \DeclareFieldAlias{bibhyperref}{noformat}%
  \bibhyperref{%
    #1%
    \ifbool{cbx:parens}
      {\bibcloseparen\global\boolfalse{cbx:parens}}
      {}}}

\savebibmacro{cite}
\savebibmacro{textcite}

\renewbibmacro*{cite}{%
  \printtext[citehyperref]{%
    \restorebibmacro{cite}%
    \usebibmacro{cite}}}

\renewbibmacro*{textcite}{%
  \ifboolexpr{
    ( not test {\iffieldundef{prenote}} and
      test {\ifnumequal{\value{citecount}}{1}} )
    or
    ( not test {\iffieldundef{postnote}} and
      test {\ifnumequal{\value{citecount}}{\value{citetotal}}} )
  }
    {\DeclareFieldAlias{textcitehyperref}{noformat}}
    {}%
  \printtext[textcitehyperref]{%
    \restorebibmacro{textcite}%
    \usebibmacro{textcite}}}

\DeclareFieldFormat{date}{\textbf{#1}}
\DeclareFieldFormat{year}{\textbf{#1}}
\DeclareFieldFormat[article,inproceedings,incollection,thesis]{title}{\mkbibquote{#1}}
\DeclareFieldFormat{url}{\url{#1}}
\DeclareFieldFormat{urldate}{(visited~#1)}

\usepackage{epigraph}
\begin{document}

% -----------------------------------------------------------------------------
% Front Matter
% -----------------------------------------------------------------------------
\frontmatter

% Official TUM title page
\includepdf[pages=1]{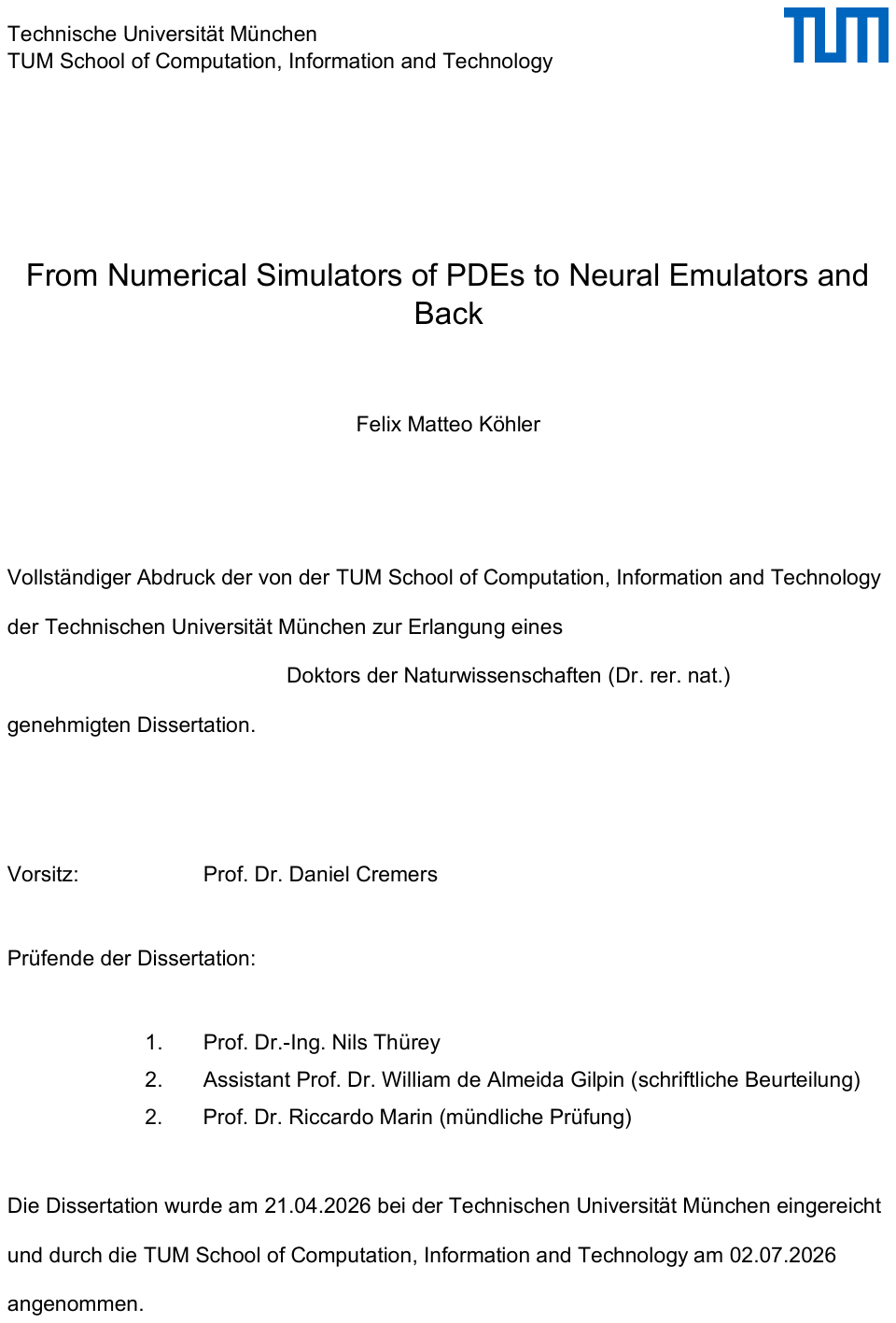}

% Blank page after title
\cleardoublepage

% Abstract
\chapter{Abstract}
\label{ch:abstract}

Simulation is central to modern engineering and science, but the cost of
numerical solvers for partial differential equations (PDEs) remains a bottleneck
whenever fast or many-query evaluations are required. Neural emulators trained on
solver-generated data promise significant speedups, yet they are usually framed
as opaque alternatives to the very methods that produce their training signal.
This thesis argues the two paradigms are more alike than different: neural
architectures mirror classical discretizations, their errors are amenable to the
same spectral analysis, and insight flows profitably in both directions. We
approach the relationship by disentangling the multiple roles a solver plays in
the emulator learning pipeline. Mode-wise Fourier analysis then provides a
common language in which solver errors, architectural inductive biases, and
training objectives can all be read off simultaneously. Taken together, this
allows synthesizing three contributions. (1) \textbf{APEBench}, a comprehensive
benchmarking suite for autoregressive neural emulators of PDEs that uses fast
differentiable pseudo-spectral solvers in JAX. (2) \textbf{Progressively Refined
Differentiable Physics}, an investigation of the effect of unconverged solvers
on surrogate training. (3) \textbf{Neural Emulator Superiority}, an analysis of
the influence of numerical errors and architectural inductive biases.

\cleardoublepage

% Zusammenfassung (German Abstract)
\chapter{Zusammenfassung}
\label{ch:zusammenfassung}

Simulationen sind ein zentrales Werkzeug in Ingenieurwesen und Wissenschaft.
Dennoch bleiben die Kosten numerischer Löser für partielle
Differentialgleichungen (PDGl) ein limitierender Faktor, insbesondere wenn
schnelle oder vielfach wiederholte Auswertungen erforderlich sind. Neuronale
Emulatoren, die auf den synthetischen Daten solcher Simulatoren trainiert
werden, versprechen erhebliche Geschwindigkeitsgewinne. Sie werden jedoch häufig
als undurchsichtige Alternative zu genau jenen Verfahren dargestellt, aus denen
ihr Trainingssignal stammt. In dieser Arbeit argumentieren wir, dass beide
Paradigmen einander ähnlicher sind, als es zunächst scheint: Neuronale
Architekturen spiegeln klassische Diskretisierungsverfahren wider, ihre Fehler
lassen sich mit derselben spektralen Analyse untersuchen, und Erkenntnisse
fließen gewinnbringend in beide Richtungen. Wir nähern uns dieser Verbindung,
indem wir die vielfältigen Rollen, die ein numerischer Löser im Trainingsprozess
eines Emulators einnimmt, klar voneinander trennen. Eine modenweise
Fourier-Analyse bietet dann eine gemeinsame Sprache, in der sich Löserfehler,
architekturelle induktive Biases und Lernziele gleichzeitig ablesen lassen.
Dieses Gerüst ermöglicht die Synthese dreier wissenschaftlicher Arbeiten.
(1)~\textbf{APEBench}, eine umfassende Benchmark-Suite zur Evaluierung
autoregressiver neuronaler Emulatoren von PDGl, basierend auf schnellen,
differenzierbaren pseudo-spektralen Lösern in JAX. (2)~\textbf{Progressively
Refined Differentiable Physics}, eine Untersuchung der Auswirkung nicht
vollständig konvergierter Löser auf das Emulatortraining. (3)~\textbf{Neural
Emulator Superiority}, eine Analyse des Einflusses numerischer Fehler und
architektureller induktiver Biases.

\cleardoublepage

% Acknowledgments
\chapter{Acknowledgments}
\label{ch:acknowledgments}

I am extremely grateful for the opportunity to have worked at the intersection
of two topics that equally fascinate me: Machine Learning and Numerical
Simulation. Over the course of my academic journey from a Bachelor degree in
mechanical engineering at TU Braunschweig, a Master degree in computational
science at TU Munich, and now a PhD in computer science, I have had the chance to
dive deep into topics that were close to my heart. Along the way, I had the
pleasure of working with many brilliant people who strongly influenced my
thinking and questioned my beliefs, and whom I want to acknowledge here.

First, I have to thank my doctoral advisor, Prof. Dr. Nils Thuerey. I am
especially grateful for his patience through the many early pivots before we
settled on the APEBench direction that anchored my doctorate. I am also
grateful for what he taught me about good scientific writing, a craft that will
serve me well beyond this dissertation, and for his open-mindedness toward the
emulator-superiority project.

I also want to thank Prof. Dr. William Gilpin for acting as the referee and
Prof. Dr. Daniel Cremers for chairing this PhD council.

I owe special appreciation to my collaborators. This starts with Simon
Niedermayr, whose expertise in computer vision and rendering not only
enlightened me but who also joined the APEBench effort and contributed an
amazing volume rendering tool that has become popular in the physics-based deep
learning community. I thank Prof. Dr. Rüdiger Westermann for his support and
for the computer graphics perspective on the APEBench paper.

My second paper of this PhD sprung out of the supervision of Kanishk Bhatia's
Master thesis. I appreciate Kanishk for taking on this deep dive into autodiff
fundamentals and bilevel optimization with me. I fondly remember our vivid
discussions.

I am grateful to Qiang Liu for his deep interest in the Exponax solver and for
translating it to PyTorch. Thank you also for identifying and pushing for the
importance of fast Fourier-spectral solvers that led to our Tadpole work. In
this light, I also thank Benjamin Holzschuh for collaborating on Tadpole and
for his deep expertise on 3D Transformer architectures and probabilistic
modeling.

A big thank you also goes to Paul Setinek for his interest in the holistic
perspective on neural surrogates that my work advocated for, and, on that basis,
for suggesting our collaboration on compute-optimal data generation. It was a
pleasure to exchange knowledge and I appreciate how our expertise combined to
build something great. In this setting, I also want to thank Dr.
Fabian Paischer and Prof. Dr. Johannes Brandstetter for working together on
this exciting project.

My time as a PhD student would have been way less fun and intellectually
stimulating if there were no colleagues. I am grateful for the fun times we
shared together, the conversations at Mensa (for which I motivated you to go
early :D, not to mention the ``Studitopf Fastenzeit''), and the sharing of
knowledge. Thank you Aleksandra, Benjamin, Bernhard, Björn, Chengyun, Georg,
Giacomo, Hao, Kanishk, Kevin, Liwei, Luca, Lukas, Luis, Mario, Marton, Mohammed,
Nilam, Patrick, Philipp, Qiang, Rene, Robin, Shuvayan, Timo, Yunjia Ziqi, and
also Christoph, David, Han, Josef, Junpeng, Kevin, Ludwic, Neel, Simon. Among
these wonderful colleagues, I especially want to thank Dr. Rene Winchenbach for
being my mentor, Dr. Luca Guastoni for acquiring compute resources that
benefited parts of my research, Patrick Schnell for being my autodiff and toy
problem buddy, Philipp Holl for stimulating discussions on (sparse) linear
solvers and differentiable simulation, Björn List for making me aware of Fourier
ETDRK methods that shaped large parts of my research, Georg Kohl for being an
early test user of Exponax (scraping terabytes of data for our foundation model
efforts), Kevin Höhlein for feedback on statistical rigor in APEBench, David
Berger for supporting us with transitioning the Game Physics programming
template to modern WebGPU, and Neel Kelkar for introducing me to modern CLI
coding agents.

Beyond Kanishk, I had the pleasure of supervising further students whose
curiosity and questions sharpened my own understanding. In particular, I want to
thank Stefan Jacob and Mohammed Abdul Moeed for the thoughtful discussions
during their theses and for building upon APEBench. I also want to thank
Constantin Le Clei for building upon Exponax and the wonderful discussions on
autoregressive error accumulation.

I am indebted to Susanne Weitz for her administrative support. I stopped
counting how often I consulted you for administrative questions. Thank you for
always being there and always having an open ear. Thanks also to Sebastian
Wohner for ensuring that our IT infrastructure ran flawlessly.

During my PhD, I had the opportunity to intern at Meta Reality Labs. This short
excursion into computer graphics simulations and unstructured surrogates gave me
a better understanding of neural emulators as a whole. For this, I thank Dr.
Ryan Goldade for hosting me, as well as Dr. Tuur Stuyck and Dr. Egor Larionov
for being my project peers. My time in Zurich was especially made wonderful by
my colleagues Artur, Cheeun, Christian, Fating, Jiaxi, Jose, Lukas (I and II),
and Philippe.

Throughout my doctorate I was fortunate to attend many conferences. I want to
thank the peers I met (repeatedly) with whom I enjoyed intellectually
stimulating discussions. Thank you Adrian Hill, Ali Can Bekar, Anthony Zhou,
Artur Toshev, Atul Agrawal, Bharat Srikishan, Dan MacKinlay, Daniel Musekamp,
Daniell Worrall, Edgar Torres, Fabian Roth, Fabian Waschkowski, Frank Schäfer,
Freja Petersen, Geoffrey Negiar, Jan Hagnberger, Leonard Storcks, Lucas Meyer,
Maximilian Herde, Michael McCabe, Peter Holderrieth, Rafael Pastrana, Rudy
Morel, Tobias Pfaff, and Zongyi Li.

My thanks go to Jousef Murad, Dr. Emmanuel Franck, Dr. Victor Michel-Dansac,
and Dr. Olof Mogren for inviting and hosting me for presentations and podcasts.

Moreover, I want to thank my peers at Pasteur Labs for intellectual exchange and
the combined effort to organize the EurIPS workshop. Thank you Alexander Lavin,
Dr. Dion Häfner, Dr. Andrei Paleyes, and Andrin Rehmann. I am additionally
grateful for Dion's feedback on the early iterations of Exponax.

I owe special appreciation to the mentors who have guided me through my academic
journey and broader career. I thank Prof. Dr. Dirk Langemann for motivating my
mathematical curiosity and for encouraging and supporting me to apply for
scholarships. I am grateful to Dr. Fabian Sewerin for inspiring my interest in
numerical methods through his FEM lectures, for supervising my Bachelor's
thesis, and for supporting me with scholarship applications. I also want to
greatly acknowledge Dr. Dirk Hartmann, whose mentorship and guidance began at
Siemens and have continued well beyond.

On a broader note, I want to acknowledge the Munich Center for Machine Learning
for funding parts of my research. I also want to thank the developers and
maintainers of the many open source libraries that made my research possible, in
particular around the JAX and Julia ecosystems. Among many others, Dr. Chris
Rackauckas and Dr. Patrick Kidger deserve thanks for pushing modern simulation
paradigms, democratizing a contemporary understanding of autodiff, and acting as
role models for maintaining scientific software.

Last, but not least, I want to thank my family and the friends I have made
across my years in Meiningen, Braunschweig, Munich, Stockholm, Zurich, and
beyond. Thank you for all the laughs, the great times, the philosophical
discussions, and for showing me a life outside of research. My deepest love and
appreciation go to Vivi, who supported me throughout my intellectual endeavors,
endured the foggy aftermath of stressful weeks, and was there for me in my
hardest times.

Thank you.
\begin{flushright}
Munich, April 2026
\end{flushright}

\cleardoublepage

% Table of Contents
{\hypersetup{linkcolor=TUMBlueDark}\tableofcontents}

\cleardoublepage

% List of Publications
\chapter{List of Publications}
\label{ch:publications}

% --- Publication card commands ---
% Venue badge
\newcommand{\venuebadge}[1]{%
    \tikz[baseline=(badge.base)]{%
        \node[fill=TUMBlue, text=white, rounded corners=2pt,
              inner xsep=7pt, inner ysep=3pt,
              font=\sffamily\small\bfseries] (badge) {#1};%
    }%
}

% Core publication card
\newcommand{\pubcard}[5]{%
    % #1 = number, #2 = title, #3 = authors, #4 = venue badge text, #5 = chapter ref
    \noindent\hspace{16mm}%
    \begin{tikzpicture}[baseline=(card.center)]
        \node[
            inner sep=0pt,
            outer sep=0pt,
        ] (card) {%
            \begin{minipage}{\dimexpr\textwidth-26mm\relax}
                \setlength{\leftskip}{4mm}%
                {\large\bfseries\color{TUMBlueDark}#2}\\[3pt]
                {\color{black!70}#3}\\[7pt]
                \venuebadge{#4}\hspace{8pt}\hfill%
                {\small\color{TUMGrayLight}#5}%
            \end{minipage}%
        };
        % Left accent bar
        \fill[TUMBlue] ([xshift=-1mm]card.north west) rectangle
            ([xshift=1.5mm]card.south west);
        % Number circle
        \node[
            circle, fill=TUMBlue, text=white,
            font=\sffamily\bfseries,
            minimum size=7mm, inner sep=0pt,
            anchor=east
        ] at ([xshift=-3mm]card.west) {\small#1};
    \end{tikzpicture}%
}

% Additional publication card (lighter accent)
\newcommand{\pubcardlight}[4]{%
    % #1 = number, #2 = title, #3 = authors, #4 = status
    \noindent\hspace{16mm}%
    \begin{tikzpicture}[baseline=(card.center)]
        \node[
            inner sep=0pt,
            outer sep=0pt,
        ] (card) {%
            \begin{minipage}{\dimexpr\textwidth-26mm\relax}
                \setlength{\leftskip}{4mm}%
                {\bfseries\color{TUMBlueDark}#2}\\[3pt]
                {\color{black!70}#3}\\[3pt]
                {\small\itshape\color{TUMGrayLight}#4}%
            \end{minipage}%
        };
        % Left accent bar (lighter)
        \fill[TUMBlueLight] ([xshift=-1mm]card.north west) rectangle
            ([xshift=1.5mm]card.south west);
        % Number circle
        \node[
            circle, fill=TUMBlueLight, text=white,
            font=\sffamily\bfseries,
            minimum size=7mm, inner sep=0pt,
            anchor=east
        ] at ([xshift=-3mm]card.west) {\small#1};
    \end{tikzpicture}%
}

% Software package item
\newcommand{\softwareitem}[2]{%
    % #1 = name, #2 = description
    \tikz[baseline=(name.base)]{%
        \node[fill=TUMBlue!10, rounded corners=2pt,
              inner xsep=5pt, inner ysep=2pt,
              font=\ttfamily\bfseries\color{TUMBlueDark}] (name) {#1};%
    }%
    \hspace{2pt}{\small #2}%
}

% Thesis card (no number circle, just accent bar)
\newcommand{\thesiscard}[3]{%
    % #1 = title, #2 = student name, #3 = thesis type
    \noindent\hspace{6mm}%
    \begin{tikzpicture}[baseline=(card.center)]
        \node[
            inner sep=0pt,
            outer sep=0pt,
        ] (card) {%
            \begin{minipage}{\dimexpr\textwidth-14mm\relax}
                \setlength{\leftskip}{4mm}%
                {\bfseries\color{TUMBlueDark}#1}\\[3pt]
                {\color{black!70}#2}\\[3pt]
                {\small\itshape\color{TUMGrayLight}#3}%
            \end{minipage}%
        };
        % Left accent bar
        \fill[TUMBlueLight] ([xshift=-1mm]card.north west) rectangle
            ([xshift=1.5mm]card.south west);
    \end{tikzpicture}%
}

% ---- Core Publications ----
% \vspace{0.2cm}
\noindent{\color{TUMBlueDark}\large\bfseries Core Publications}

\noindent The following publications form the basis of this dissertation.
Sole first authorship is indicated by a core factor of~1, shared first
authorship by a factor of~0.5.

\vspace{0.4cm}

\pubcard{1}%
    {APEBench: A Benchmark for Autoregressive Neural Emulators of PDEs}%
    {\textbf{Felix Koehler}*, Simon Niedermayr, R\"{u}diger Westermann, Nils Thuerey}%
    {NeurIPS 2024}%
    {Core factor: 1 \quad $\to$ \cref{sec:summary:apebench}}

\vspace{0.5cm}

\pubcard{2}%
    {PRDP: Progressively Refined Differentiable Physics}%
    {Kanishk Bhatia*, \textbf{Felix Koehler}*, Nils Thuerey}%
    {ICLR 2025}%
    {Core factor: 0.5 \quad $\to$ \cref{sec:summary:prdp}}

\vspace{0.5cm}

\pubcard{3}%
    {Neural Emulator Superiority: When Machine Learning for PDEs Surpasses its Training Data}%
    {\textbf{Felix Koehler}*, Nils Thuerey}%
    {NeurIPS 2025}%
    {Core factor: 1 \quad $\to$ \cref{sec:summary:superiority}}

\cleardoublepage

% -----------------------------------------------------------------------------
% Main Matter
% -----------------------------------------------------------------------------
\mainmatter

% =============================================================================
% CHAPTER 1: Introduction
% =============================================================================
\chapter{Introduction}
\label{ch:introduction}

\section{The Importance and Computational Challenge of Simulating Partial
Differential Equations}
\label{sec:intro:challenge}

Our understanding of nature and engineering increasingly depends on running
large-scale simulations of physical phenomena \citep{bauer2015quiet}. These
approximations to fundamental descriptions of reality allow gaining deep
insights into processes that would be hard to investigate with real-world
experiments. For example, it allows understanding how materials behave at a
subatomic level \citep{car1985unified}, how objects deform under load
\citep{zienkiewicz2013finite}, how aerodynamic a car design is
\citep{ferziger2020computational}, how to animate special effects in video games
or feature films realistically~\citep{bridson2015fluid}, or how to forecast the
weather \citep{bauer2015quiet}. Across all those scales, there is a fundamental
language that describes these processes, the one of partial differential
equations (PDEs). Those describe a constraint that a \emph{solution field},
e.g., a fluid velocity field, must fulfill to be in a valid state. Classically,
\emph{simulation} is the procedure of finding approximate solutions by
symbolically manipulating those continuous descriptions, deriving efficient
discretizations from first principles and implementing them as a computer program.

While first approaches can be traced back more than 100 years ago
\citep{richardson1922weather,courant1928partiellen}, modern simulation science
grew alongside the growth of digital computers
\citep{gustafsson2018scientific}\footnote{See also
\url{https://history.siam.org/}.}. Consistent with the increase of computational
power~\citep{strohmaier2015top500} and improvements in
algorithms~\citep{rude2018research}, ever larger and more sophisticated
simulations became feasible. Of particular interest is the simulation of
problems in a three-dimensional space with an additional time axis, as those
most closely resemble reality \citep{pope2001turbulent}. While there is an
ever-increasing set of problems we \emph{can} simulate, the required time for
doing so can be infeasibly high. Especially because the difficulty is not just
simulating a problem \emph{once}, but having the simulation be part of a larger
experimental loop. This could include a human design engineer who iteratively
tweaks parts of the geometry and requires feedback from the simulation. It could
also be an automated optimization process, which might require on the order of
thousands of simulations to converge \citep{forrester2008engineering}.

\section{Classical Numerical Methods and their Surrogation}
\label{sec:intro:paradigms}

To reduce the computational cost of simulations, researchers devised
\emph{surrogates} to full-fledged simulations. Oftentimes, they purposefully
leave out certain details that are deemed less important for usage in the
larger design loop. There are typically three types of building surrogates:
\begin{enumerate}
    \item One can use \textbf{simpler models}, e.g., finding a simplification in
    two or one dimension, or using a simpler constitutive model that is easier
    to integrate.
    \item One can \textbf{truncate numerics} of the simulation, e.g., by
    prematurely terminating an iterative process or under-resolving a phenomenon.
    \item One can devise \textbf{data-driven surrogates} that attempt to
    understand patterns from \emph{concrete examples} of the simulation.
\end{enumerate}

While in approaches (1) and (2), we keep control over numerical guarantees and
first principles, in (3) we often purposefully loosen them. Early approaches
following (3) were called Reduced-Order Models, oftentimes using Proper
Orthogonal Decomposition (POD\footnote{In the machine learning world, this is
typically referred to as principal component
analysis~\citep{bishop2006pattern}.}) \citep{benner2015survey}. In its core, this
approach tried to \emph{compress} a high-resolution state space into a more
manageable representation in which one can efficiently integrate in time. It
tried to approximate the \emph{physics' time derivative} but left the time
integration to classical approaches.

A recent trend is to replace \emph{both the physics and the numerics} by a
machine learning model \citep{guo2016convolutional,li2020fourier}. As such, one
finds surrogates that, in their most extreme forms, are fully data-driven, without
any numerical principles. Since such approaches are mostly based on deep
learning, we will refer to them as \emph{neural emulators}. Oftentimes, they
handle time sequentially, i.e., they predict next frames based on previous ones
autoregressively~\citep{brandstetter2021message}. Hence, a core theme of this
thesis will be \emph{autoregressive neural emulators}.

\section{Thesis Statement and Research Questions}
\label{sec:intro:thesis}

\begin{figure}
    \centering
    \begin{tikzpicture}[
        % Node styles
        sysnode/.style={
            rectangle, rounded corners=3pt,
            draw=TUMGrayLight, fill=TUMGray,
            minimum width=1.6cm, minimum height=0.75cm,
            font=\small\bfseries, align=center
        },
        solvernode/.style={
            rectangle, rounded corners=3pt,
            draw=orange!60!black, fill=orange!12,
            minimum width=1.6cm, minimum height=0.75cm,
            font=\small\bfseries, align=center
        },
        datanode/.style={
            rectangle, rounded corners=3pt,
            draw=TUMGrayLight, fill=TUMGray!50,
            minimum width=1.6cm, minimum height=0.75cm,
            font=\small\bfseries, align=center
        },
        emunode/.style={
            rectangle, rounded corners=3pt,
            draw=TUMBlue!80, fill=TUMBlue!12,
            minimum width=1.6cm, minimum height=0.75cm,
            font=\small\bfseries, align=center
        },
        icnode/.style={
            rectangle, rounded corners=2pt,
            draw=orange!40!black, fill=orange!6,
            minimum width=1.2cm, minimum height=0.55cm,
            font=\scriptsize, align=center
        },
        arr/.style={-{Stealth[length=4pt]}, thick, color=TUMGrayLight!80!black},
        lbl/.style={font=\scriptsize, text=TUMGrayLight!60!black},
    ]

    % Nodes
    \node[sysnode] (P) {$\mathbb{P}$};
    \node[solvernode, right=1.8cm of P] (Ph) {$\mathcal{P}_h$};
    \node[solvernode, right=1.4cm of Ph] (T) {$\mathcal{T}$};
    \node[datanode, right=1.4cm of T] (D) {$\mathcal{D}_h$};
    \node[emunode, right=1.8cm of D] (f) {$f_\theta$};

    % IC distribution above rollout
    \node[icnode, above=0.9cm of T] (IC) {$\mathcal{I}_h$};

    % Labels below nodes
    \node[lbl, below=2pt of P] {PDE System};
    \node[lbl, below=2pt of Ph] {Solver};
    \node[lbl, below=2pt of T] {Rollout};
    \node[lbl, below=2pt of D] {Data};
    \node[lbl, below=2pt of f] {Emulator};

    % Arrows
    \draw[arr] (P) -- (Ph)
        node[midway, above, font=\scriptsize] {$\mathbb{M}, N, \Delta t$};
    \draw[arr] (Ph) -- (T);
    \draw[arr] (IC) -- (T)
        node[midway, right, font=\scriptsize] {$\mathbf{u}_h^{[0]}$};
    \draw[arr] (T) -- (D);

    % Bidirectional train/evaluate arrows
    \draw[arr, bend left=18, color=orange!60!black]
        (D.north east) to
        node[midway, above, font=\scriptsize, text=orange!50!black] {train}
        (f.north west);
    \draw[arr, bend left=18, color=TUMBlue!80]
        (f.south west) to
        node[midway, below, font=\scriptsize, text=TUMBlue!80] {evaluate}
        (D.south east);

    \end{tikzpicture}
    \caption{%
        Overview of the emulation pipeline. A PDE system $\mathbb{P}$ is
        discretized into a solver $\mathcal{P}_h$ by choosing a numerical method
        $\mathbb{M}$, resolution $N$, and time step $\Delta t$. Initial
        conditions $\mathbf{u}_h^{[0]}$ sampled from a distribution
        $\mathcal{I}_h$ are rolled out via the trajectory operator $\mathcal{T}$
        to produce data trajectories $\mathcal{D}_h$, which are used to train
        and test the neural emulator $f_\theta$. }
    \label{fig:emulation-pipeline}
\end{figure}
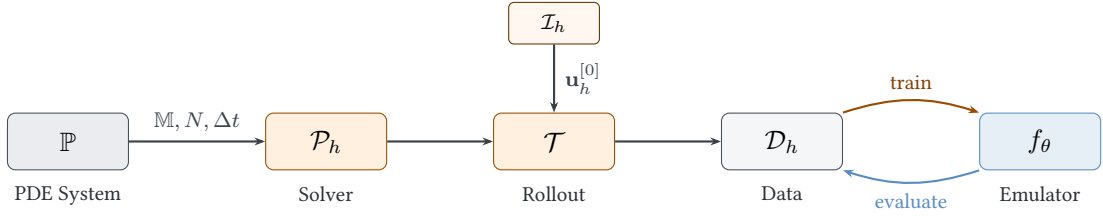

\begin{figure}
    \centering
    \begin{tikzpicture}[
        paper/.style={
            rectangle, rounded corners=3pt,
            minimum width=4.2cm, minimum height=0.75cm,
            font=\small\bfseries, align=center,
        },
        paradigm/.style={
            rectangle, rounded corners=3pt,
            draw=TUMGrayLight, fill=TUMGray,
            minimum width=2.8cm, minimum height=0.6cm,
            font=\small, align=center,
        },
        biarrow/.style={{Stealth[length=4pt]}-{Stealth[length=4pt]}, semithick},
    ]

    % === Top: paradigm framing ===
    \node[paradigm] (solvers) at (-2.3, 3.8) {Numerical Solvers};
    \node[paradigm] (emulators) at (2.3, 3.8) {Neural Emulators};
    \draw[biarrow, TUMGrayLight!80, thick]
        (solvers) -- (emulators)
        node[midway, above=2pt, font=\scriptsize, text=TUMGrayLight]
        {\emph{``\dots and Back''}};

    % === Paper nodes ===
    \node[paper, draw=purple!50, fill=purple!8] (sup) at (0, 2.2)
        {Emulator Superiority};
    \node[font=\scriptsize, text=purple!60!black, anchor=east, align=right]
        at (-4pt, 1.45) {Understanding \textbf{WHY}};

    \node[paper, draw=TUMBlue!60, fill=TUMBlue!8] (ape) at (0, 0)
        {APEBench};
    \node[font=\scriptsize, text=TUMBlue!80, anchor=east, align=right]
        at (-4pt, -0.75) {Measuring \textbf{WHAT}};

    \node[paper, draw=green!50!black, fill=green!8] (prdp) at (0, -2.2)
        {PRDP};
    \node[font=\scriptsize, text=green!40!black, anchor=east, align=right]
        at (-4pt, -2.95) {Training \textbf{HOW}};

    % === Edges between papers ===
    \draw[biarrow, purple!40!TUMBlue]
        (sup) -- (ape)
        node[midway, right=4pt, font=\scriptsize, text=TUMGrayLight, align=left]
        {framework enables\\systematic study};

    \draw[biarrow, TUMBlue!40!green!60!black]
        (ape) -- (prdp)
        node[midway, right=4pt, font=\scriptsize, text=TUMGrayLight, align=left]
        {benchmarks training\\strategies};

    % === Dashed links from paradigms ===
    \draw[TUMGrayLight!50, densely dashed, thin]
        (solvers.south) -- (sup.north west);
    \draw[TUMGrayLight!50, densely dashed, thin]
        (emulators.south) -- (sup.north east);

    \end{tikzpicture}
    \caption{%
        Overview of the three publications comprising this thesis. APEBench
        provides the benchmarking framework that enables systematic measurement
        of neural PDE emulators. In Emulator Superiority, we study the effect of
        numerical errors on surrogate quality with the surprising finding that they can
        sometimes exceed their training data fidelity. With PRDP, we demonstrate
        how to train emulators efficiently using progressively refined
        differentiable solvers. The bidirectional arc captures the thesis title:
        insights flow from numerical solvers to neural emulators and back. }
    \label{fig:thesis-overview}
\end{figure}
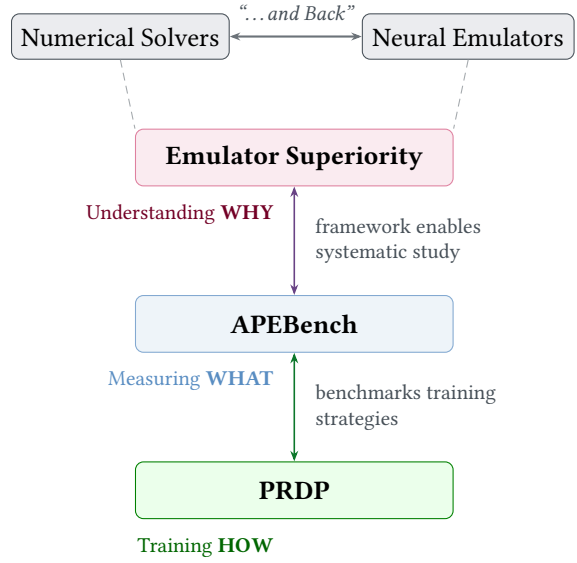

The prevalent theme of training autoregressive neural emulators is to use
synthetic data produced with conventional numerical methods; see
\cref{fig:emulation-pipeline} for an overview. This is different from many other
domains of machine learning that use large-scale \emph{empirically} obtained
datasets like annotated images or text corpora from the web
\citep{deng2009imagenet,brown2020language}. Since the ultimate goal of creating
surrogates is finding a tool that is more advantageously positioned on the
ever-present accuracy-speed trade-off in the approximate solutions of PDEs, this
creates an interesting interplay between emulators and simulators. Moreover, the
neural architectures being used as surrogates have structural and functional
similarities with certain numerical algorithms
\citep{long2018pde,alt2023connections}. These two aspects, the student-teacher
relationship and the functional similarity, raise interesting fundamental
questions that are central to this thesis, leading to the statement:
\begin{quote}
    Numerical simulators of PDEs and neural emulators share significant
    similarities and a deep understanding of the holistic surrogate-learning
    pipeline can inform good decisions in both directions.
\end{quote}
Concretely, we approach this from three angles. First, we use tools from
numerical analysis, in particular Fourier-spectral error analysis, to understand
\emph{why} certain emulators succeed or fail, and why they can sometimes exceed
the fidelity of the solver that generated their training data. Second, we show
\emph{how} the solver's role in the training loop can be exploited: full
convergence of iterative solvers is unnecessary for producing efficient
surrogates, enabling substantial computational savings. Third, we build a
benchmarking framework that treats the solver as a first-class component,
allowing systematic investigation of \emph{what} works across 46 PDE
configurations, multiple architectures, and diverse training strategies. We
restrict ourselves to semi-linear PDEs on periodic, uniform Cartesian grids, a
simplification that enables the Fourier-spectral analysis central to our
arguments while still covering a rich set of representative dynamics.
\Cref{fig:thesis-overview} visually summarizes the connection between the pieces
of research that make up this cumulative dissertation.

\section{Outline}
\label{sec:intro:outline}

\Cref{ch:foundations} reviews numerical methods for solving time-dependent PDEs
on uniform Cartesian grids, covering spatial and temporal discretization, the
spectral error analysis that underpins much of the later argumentation, and
iterative solvers relevant to differentiable physics. The chapter also presents
the JAX-based solver implementations (Exponax, Picardax, Chaotax) developed as
part of this thesis. \Cref{ch:emulators} introduces neural emulator
architectures and draws structural parallels with numerical methods:
convolutional networks and finite difference stencils, Fourier Neural Operators
and spectral methods, UNets and multigrid. We describe the data-driven training
pipeline for autoregressive emulators and discuss different aspects of
generalization. \Cref{ch:relationship} is the synthesis chapter. We identify
five roles that numerical solvers play in the emulator pipeline, then use this
framework to explain three phenomena: emulator superiority (proven via
Fourier-spectral analysis), computational savings from progressive solver
refinement (PRDP), and the benchmarking implications that follow from both.
\Cref{ch:apebench} describes the APEBench benchmarking suite, its software
ecosystem, and systematic experimental findings. \Cref{ch:outlook} distills
cross-cutting findings, discusses limitations and open questions, and outlines
future directions, including foundation models, compute-optimal data generation,
and the extension to unstructured meshes. \Cref{ch:publications_summary}
provides a concise summary of each publication. The full papers are reproduced
in the Appendix.

% =============================================================================
% CHAPTER 2: Foundations
% =============================================================================
\chapter{Foundations of Numerical Simulation}
\label{ch:foundations}

We want to solve time-dependent partial differential equations (PDEs) of the form
\begin{equation}\label{eq:general-pde}
    \frac{\partial u}{\partial t}
    =
    \mathcal{L} u
    +
    \mathcal{N}(u),
\end{equation}
in which $\mathcal{L}$ and $\mathcal{N}(\cdot)$ are linear and nonlinear
differential operators, respectively. The nonlinear operator can be
\emph{inhomogeneous}, i.e., it can include a forcing term. This equation is
solved for the solution function $u(t, \mathbf{x})$.
% If this function is vector-valued, i.e., if it constitutes multiple channels, we
% use bold-faced $\mathbf{u}(t, \mathbf{x})$ instead.
This function represents a \emph{field} on the $D$-dimensional domain $\Omega
\subseteq \mathbb{R}^D$ for a time $t \in [0, T)$. Typically, the domain is
bounded. Then, we additionally supply boundary conditions on $\partial \Omega$.
For the time axis, one has to prescribe an \emph{initial condition}, i.e., the
value of $u(t=0, \mathbf{x})$.

An example is the one-dimensional Burgers equation,
\begin{equation}\label{eq:burgers-equation-1d}
    \frac{\partial u}{\partial t}
    +
    \frac{1}{2}
    \frac{\partial u^2}{\partial x}
    =
    \nu \Delta u.
\end{equation}
Here, the linear operator is the Laplace operator scaled by the diffusivity
$\nu$, i.e., $\mathcal{L} = \nu \Delta$. The nonlinear operator is the negated
convection operator in conservative form, i.e., $\mathcal{N}(u) = - \frac{1}{2}
\frac{\partial u^2}{\partial x}$.

We refer to the combination of a PDE, its constitutive parameters (e.g., the
diffusivity $\nu$), and boundary conditions as a \emph{PDE system} $\mathbb{P}$ (also called an \emph{initial-boundary value problem}).
A PDE system defines a continuous problem, but does not prescribe how it is
solved on a computer. Choosing a numerical method $\mathbb{M}$ (e.g., finite
differences with Crank-Nicolson time stepping), a spatial resolution $N$, and
for time-dependent problems a time step $\Delta t$, yields a discrete solver
$\mathcal{P}_h$ (the subscript $h$ and variations of it are used to indicate
that this is a discrete approximation).
This distinction matters because multiple solvers $\mathcal{P}_h, \mathcal{P}_H,
\ldots$ can approximate the same system $\mathbb{P}$ at different
\emph{fidelities}, i.e., at different accuracies or with different properties.
\label{sec:foundations:fidelity}

\section{Numerical Methods for Time-Dependent PDEs}
\label{sec:foundations:numerical}

\begin{figure}[t]
    \centering
    \resizebox{\textwidth}{!}{%
    \begin{tikzpicture}[
        stagebox/.style={
            rectangle, rounded corners=3pt,
            minimum width=3.0cm, minimum height=1.7cm,
            font=\scriptsize, align=center, inner sep=6pt,
        },
        pipearrow/.style={-{Stealth[length=3pt]}, semithick, TUMBlueDark},
        annot/.style={font=\tiny, text=TUMGrayLight},
        seclabel/.style={font=\tiny\itshape, text=TUMBlue!70},
    ]

    % ================================================================
    % Stage 1: Continuous PDE
    % ================================================================
    \node[stagebox, draw=TUMGrayLight!60, fill=TUMGray!30]
        (pde) at (0, 0)
        {\textbf{Continuous PDE}\\[3pt]
         $\displaystyle\frac{\partial u}{\partial t}
         = \mathcal{L}u + \mathcal{N}(u)$\\[2pt]
         {\tiny on $\Omega \subseteq \mathbb{R}^D$}};

    % ================================================================
    % Stage 2: Semi-discrete ODE
    % ================================================================
    \node[stagebox, draw=TUMGrayLight!60, fill=TUMGray!30]
        (ode) at (5.0, 0)
        {\textbf{Semi-discrete ODE}\\[3pt]
         $\displaystyle\frac{d\mathbf{u}_h}{dt}
         = \mathcal{L}_h \mathbf{u}_h
         + \mathcal{N}_h(\mathbf{u}_h)$};

    % ================================================================
    % Stage 3: Evolution Operator
    % ================================================================
    \node[stagebox, draw=orange!60!black, fill=orange!10,
          minimum width=2.8cm]
        (evol) at (10.0, 0)
        {\textbf{Evolution Operator}\\[3pt]
         $\mathbf{u}_h^{[n+1]}
         = \mathcal{P}_h\!\left(\mathbf{u}_h^{[n]}\right)$};

    % ================================================================
    % Stage 4a: Initial Condition (sampled from a distribution)
    % ================================================================
    \node[stagebox, draw=TUMBlue!50, fill=TUMBlue!6,
          minimum width=2.8cm, minimum height=1.1cm]
        (ic) at (5.0, -3.2)
        {\textbf{Initial Condition}\\[3pt]
         $\mathbf{u}_h^{[0]} \sim \mathcal{I}_h$};

    % ================================================================
    % Stage 4b: Trajectory (placed below Evolution Operator)
    % ================================================================
    \node[stagebox, draw=TUMBlue!50, fill=TUMBlue!6,
          minimum width=3.2cm]
        (traj) at (10.0, -3.2)
        {\textbf{Discrete Trajectory}\\[3pt]
         $\mathbf{u}_h^{[0]}
         \!\xrightarrow{\;\mathcal{P}_h\;}
         \mathbf{u}_h^{[1]}
         \!\xrightarrow{\;\mathcal{P}_h\;}
         \cdots
         \!\xrightarrow{\;\mathcal{P}_h\;}
         \mathbf{u}_h^{[n]}$};

    % ================================================================
    % Arrows between stages
    % ================================================================
    % PDE → ODE
    \draw[pipearrow] (pde) -- (ode)
        node[midway, above, annot] {discretize space}
        node[midway, below, annot] {$\Delta x,\; N$};

    % ODE → Evolution operator
    \draw[pipearrow] (ode) -- (evol)
        node[midway, above, annot] {integrate time}
        node[midway, below, annot] {$\Delta t$};

    % Evolution operator → Trajectory (vertical)
    \draw[pipearrow] (evol.south) -- (traj.north)
        node[midway, right, annot] {apply repeatedly};

    % Initial Condition → Trajectory
    \draw[pipearrow] (ic.east) -- (traj.west)
        node[midway, above, annot] {sample};

    % ================================================================
    % Section references below
    % ================================================================
    \draw[TUMBlue!30, thin, decorate,
          decoration={brace, mirror, amplitude=3pt}]
        ([yshift=-14pt]pde.south west) -- ([yshift=-14pt]pde.south east)
        node[midway, below, seclabel, yshift=-2pt]
        {\cref{sec:foundations:numerical:spatial}};

    \draw[TUMBlue!30, thin, decorate,
          decoration={brace, mirror, amplitude=3pt}]
        ([yshift=-14pt]ode.south west) -- ([yshift=-14pt]ode.south east)
        node[midway, below, seclabel, yshift=-2pt]
        {\cref{sec:foundations:numerical:temporal}};

    \end{tikzpicture}%
    }
    \caption{%
        The method-of-lines pipeline for solving time-dependent PDEs.
        A continuous PDE is first discretized in space
        (\cref{sec:foundations:numerical:spatial}), yielding a
        semi-discrete system of ODEs. A time integration scheme
        (\cref{sec:foundations:numerical:temporal}) then defines the
        evolution operator~$\mathcal{P}_h$, which advances the discrete
        state~$\mathbf{u}_h^{[n]}$ by one time step~$\Delta t$.
        An initial condition~$\mathbf{u}_h^{[0]} \sim \mathcal{I}_h$
        is sampled from a prescribed distribution, and repeated
        application of~$\mathcal{P}_h$ produces a trajectory of states.
    }
    \label{fig:method-of-lines}
\end{figure}
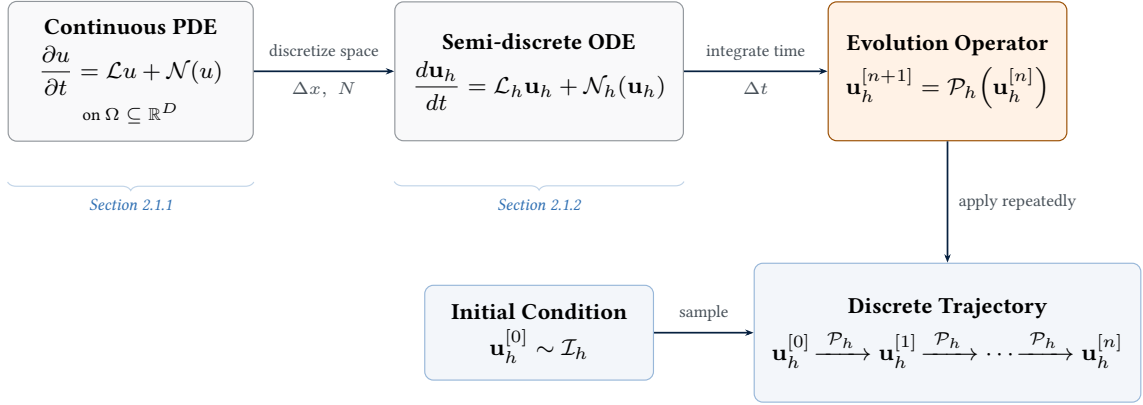

The goal of \emph{simulation} is to approximate the solution field $u(t,
\mathbf{x})$
on a computer. For this section, we will focus on \emph{time-dependent}
problems. However, parts of it are also relevant for stationary PDEs. A popular
strategy is to follow the \emph{method-of-lines} discretization
(\cref{fig:method-of-lines}) by finding a discrete representation of the
function in the domain $\Omega$ and then integrate the \emph{discrete state}
$\mathbf{u}_h$ in time. On structured grids, it is common to represent the state
using higher-order tensors, e.g., in two dimensions $\mathbf{u}_h \in
\mathbb{R}^{C \times N_x \times N_y}$\footnote{For consistency with deep
learning notation~\citep{goodfellow2016deep}, states always have a channel axis,
even if they only represent scalar fields. In the scalar case, the channel axis
is singleton, i.e., $C=1$.}. This representation is similar to images with $C$
channels (for a physical field, these channels represent the vector components).
Note that $\mathbf{u}_h$ must not necessarily
correspond to collocated grid values, but can also be coefficients of a
collection of basis functions, as described in
\cref{sec:foundations:numerical:spatial:spectral}.

For the remainder of this thesis, we will only consider \emph{uniform Cartesian
grids} that discretize the scaled unit cube $\Omega = (0, L)^D$ with $N$ degrees
of freedom per dimension. As such, a state in one, two, and three dimensions has
two, three, and four axes, respectively (due to the leading channel axis). The
last $D$ axes will all be of the same size $N$. While this is a significant
simplification, we will see in \cref{ch:apebench} that it still allows for a
rich set of PDEs that probe a wide range of representative emulation conditions. 

Given a state, we denote by $\mathbf{u}_h^{[n]}$ a state at \emph{time-level} $n
\in \mathbb{N}$. With a chosen time step size $\Delta t$, this corresponds to
physical time $t = n \Delta t$. Correspondingly, $\mathbf{u}_h^{[0]}$ is the
discretized initial condition. Then, we can step forward in time using the
evolution operator $\mathcal{P}_h$, i.e., $\mathbf{u}_h^{[n+1]} =
\mathcal{P}_h\left(\mathbf{u}_h^{[n]}\right)$. This operator absorbs the
evaluation of the spatial derivatives, a time integration scheme, and
potentially resolves implicit relations (see \cref{sec:foundations:iterative}).

For efficient execution, we require $\mathcal{P}_h$ to be \emph{batchable},
i.e., we would like to integrate multiple states $\mathbf{u}_h^{[n]}$ at the
same time. It should also run efficiently on modern hardware accelerators.
Moreover, we would like to differentiate over it, i.e., to use Jacobians
$J_{\mathcal{P}_h} = \partial \mathbf{u}_h^{[n+1]} / \partial
\mathbf{u}_h^{[n]}$ that are needed for design optimization or to build
hybrid emulators.

\subsection{Spatial Discretization}
\label{sec:foundations:numerical:spatial}

Spatial discretization approaches are characterized by (1) how they represent
the domain, (2) how they approximate spatial derivatives, and (3) how they
incorporate boundary conditions. For this thesis, we consider three (simplified)
types of boundary conditions: %(\cref{fig:boundary-grids}):
\begin{enumerate}
    \item Periodic boundaries: $u(t, \mathbf{x})\big|_{x_i=0} = u(t,
    \mathbf{x})\big|_{x_i=L}$
    \item Homogeneous Dirichlet boundaries: $u(t, \mathbf{x})\big|_{x_i \in \{0,
    L\}} = 0$
    \item Homogeneous Neumann boundaries: $\frac{\partial u}{\partial x_i}(t,
    \mathbf{x})\bigg|_{x_i \in \{0, L\}} = 0$
\end{enumerate}
Under periodic boundaries, we consider one end of the domain a degree of
freedom. In 1D, this yields the grid $\mathbf{x}_h^{\text{periodic}} = [0,
\Delta x, 2 \Delta x, \dots, (N -1) \Delta x]$ with $\Delta x = L / N$. For
Dirichlet and Neumann boundaries, the degrees of freedom only consider the
interior nodes. Hence, we would get $\mathbf{x}_h^{\text{dirichlet/neumann}} =
[\Delta x, 2 \Delta x, \dots, N \Delta x]$ with $\Delta x = 1 / (N+1)$. Higher
dimensional grids are built using tensor products.

\begin{figure}[t]
    \centering
    \begin{tikzpicture}[
        gridnode/.style={circle, fill=TUMGray, draw=TUMGrayLight!50,
                         minimum size=2.5mm, inner sep=0pt},
        fdactive/.style={circle, fill=orange!25, draw=orange!50,
                         minimum size=2.5mm, inner sep=0pt},
        spactive/.style={circle, fill=orange!15, draw=orange!40,
                         minimum size=2.5mm, inner sep=0pt},
        target/.style={circle, fill=TUMBlue, draw=TUMBlueDark,
                       minimum size=2.5mm, inner sep=0pt},
        freqnode/.style={circle, fill=orange!30, draw=orange!60,
                         minimum size=2mm, inner sep=0pt},
        arc/.style={-{Stealth[length=2pt]}, orange!50!black, thin},
    ]

    \def\colA{0}
    \def\colB{6.5}
    \def\ns{0.38}

    % =====================================================================
    % (a) Finite Differences
    % =====================================================================
    \begin{scope}[shift={(\colA,0)}]
        \node[font=\small\bfseries, text=TUMBlueDark] at (1.33,1.8) {(a) Finite Differences};
        \node[font=\scriptsize, text=TUMGrayLight] at (1.33,1.4) {Local, $\mathcal{O}((\Delta x)^p)$ accuracy};
        % Input row
        \foreach \i in {0,...,7} {
            \ifnum\i=2 \node[fdactive] (ai\i) at (\i*\ns, 0) {};
            \else\ifnum\i=3 \node[fdactive] (ai\i) at (\i*\ns, 0) {};
            \else\ifnum\i=4 \node[fdactive] (ai\i) at (\i*\ns, 0) {};
            \else \node[gridnode] (ai\i) at (\i*\ns, 0) {};
            \fi\fi\fi
        }
        % Output row
        \foreach \i in {0,...,7} {
            \ifnum\i=3 \node[target] (ao\i) at (\i*\ns, 0.85) {};
            \else \node[gridnode] (ao\i) at (\i*\ns, 0.85) {};
            \fi
        }
        % Arcs
        \draw[arc] (ai2.north) -- (ao3.south);
        \draw[arc] (ai3.north) -- (ao3.south);
        \draw[arc] (ai4.north) -- (ao3.south);
        % Stencil brace
        \draw[orange!50!black, thin, decorate,
              decoration={brace, mirror, amplitude=2pt}]
            ([yshift=-1.5mm]ai2.south) -- ([yshift=-1.5mm]ai4.south)
            node[midway, below, font=\tiny, yshift=-1.5pt,
                 text=orange!60!black] {$[1,\,-2,\,1]/(\Delta x)^2$};
        % Input/output labels
        \node[font=\tiny, text=TUMGrayLight, anchor=east] at (-0.3, 0) {$\mathbf{u}_h$};
        \node[font=\tiny, text=TUMGrayLight, anchor=east] at (-0.3, 0.85)
            {$(\partial_{xx})_h \mathbf{u}_h$};
    \end{scope}

    % =====================================================================
    % (b) Spectral Methods
    % =====================================================================
    \begin{scope}[shift={(\colB,0)}]
        \node[font=\small\bfseries, text=TUMBlueDark] at (1.33,1.8) {(b) Spectral (Fourier)};
        \node[font=\scriptsize, text=TUMGrayLight] at (1.33,1.4)
            {Global, exact for bandlimited};
        % Input row (bottom)
        \foreach \i in {0,...,7} {
            \node[spactive] (bi\i) at (\i*\ns, -0.15) {};
        }
        % Frequency row (middle)
        \foreach \i in {0,...,4} {
            \node[freqnode] (bf\i) at (0.38+\i*0.38, 0.4) {};
        }
        % Output row (top)
        \foreach \i in {0,...,7} {
            \ifnum\i=3 \node[target] (bo\i) at (\i*\ns, 0.95) {};
            \else \node[gridnode] (bo\i) at (\i*\ns, 0.95) {};
            \fi
        }
        % Trapezoidal FFT region
        \fill[orange!8, opacity=0.6]
            (bi0.north west) -- (bf0.south west) --
            (bf4.south east) -- (bi7.north east) -- cycle;
        % Trapezoidal iFFT region
        \fill[orange!8, opacity=0.6]
            (bf0.north west) -- (bo0.south west) --
            (bo7.south east) -- (bf4.north east) -- cycle;
        % Labels
        \node[font=\tiny, text=TUMGrayLight] at (-0.4, 0.12) {$\mathcal{F}_h$};
        \node[font=\tiny, text=TUMGrayLight] at (-0.5, 0.68) {$\mathcal{F}_h^{-1}$};
        % Multiply annotation
        \node[font=\tiny, text=orange!60!black, anchor=west] at (2.4, 0.4)
            {$\odot\; (ik\frac{2 \pi}{L})^2$};
        % Input/output labels
        \node[font=\tiny, text=TUMGrayLight, anchor=east] at (-0.3, -0.15) {$\mathbf{u}_h$};
        \node[font=\tiny, text=TUMGrayLight, anchor=east] at (-0.3, 0.95)
            {$(\partial_{xx})_h \mathbf{u}_h$};
    \end{scope}

    \end{tikzpicture}
    \caption{%
        Comparison of finite difference and Fourier-spectral spatial
        discretization. \textbf{(a)}~Finite differences approximate derivatives
        via a local stencil; only neighbouring grid points contribute to each
        output, yielding polynomial-order accuracy.
        \textbf{(b)}~Spectral methods transform to Fourier space
        ($\mathcal{F}_h$), apply an exact pointwise multiplier per wavenumber,
        and transform back ($\mathcal{F}_h^{-1}$); all grid points contribute
        globally, achieving exponential convergence for smooth, bandlimited
        functions on periodic domains.
    }
    \label{fig:fd-vs-spectral}
\end{figure}
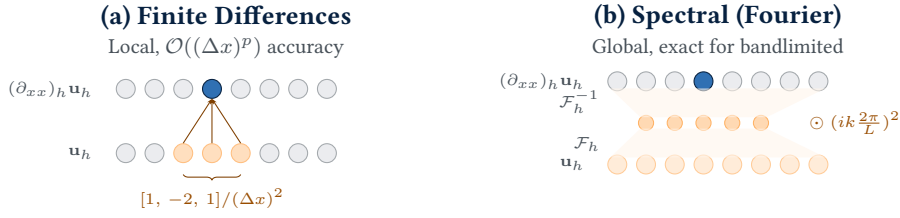

Below, we review two popular approaches for approximating spatial derivatives on
these grids: finite differences and Fourier spectral methods. See
\cref{fig:fd-vs-spectral} for a visual comparison of the two approaches.

\subsubsection{Finite Difference Methods}
\label{sec:foundations:numerical:spatial:fd}

Finite differences approximate derivatives using a Taylor series
expansion~\citep{leveque2007finite}. Computationally they can be expressed by a
cross-correlation\footnote{In a deep learning context, this is often also
loosely referred to as a convolution \citep{lecun1989backpropagation}.
Technically, a convolution would flip the filter prior to application.} ``$\star$''
with a filter $\mathbf{d}_h$. For example, to approximate the second derivative,
we could use the second-order approximation
\begin{equation}\label{eq:sample-three-point-stencil}
    \left(\frac{\partial^2 u}{\partial x^2}\right)_h
    =
    \frac{1}{(\Delta x)^2}
    [1, -2, 1]
    \star
    \mathbf{u}_h
    +
    \mathcal{O}\left((\Delta x)^2\right),
\end{equation}
which produces a \emph{spatial discretization error} that decays polynomially
quadratically with $\Delta x$. Note that this cross-correlation must be
understood to operate only on the spatial axis corresponding to spatial
dimension we are taking the derivative with respect to. All other axes are
broadcasted. There are different filters, also called stencils, for different
orders of derivatives with different orders of approximations
\citep{leveque2007finite}. Vector calculus operations like gradients or
divergence can be built by composing axis-wise derivatives. The approximation to
the nonlinear differential operator $\mathcal{N}_h(\cdot)$ is straightforwardly
implemented by combining derivatives and point-wise nonlinear evaluations.

Importantly, the cross-correlation (convolution) must respect the boundary
conditions. This is achieved by using \verb|"SAME"| \emph{padding}. The padding
mode has to match the boundary type. For periodic boundaries, we can use
circular padding. The homogeneous Dirichlet and Neumann boundaries can be
implemented by padding with constant zeros and reflect mode, respectively.

\subsubsection{Spectral Methods}
\label{sec:foundations:numerical:spatial:spectral}

Spectral methods use \emph{global} Ansatz functions and apply transformations
from collocated states $\mathbf{u}_h$ into a coefficient space
$\hat{\mathbf{u}}_h$ in which derivative operations can be evaluated
efficiently. In this thesis, we only consider Fourier-spectral methods that
exclusively work with periodic boundary conditions \citep{boyd2001chebyshev}.
Let $\mathcal{F}_h$ be the $D$-dimensional \emph{real-valued} fast Fourier
transform (in NumPy~\citep{harris2020array}, this corresponds to
\verb|np.fft.rfftn|). Without loss of generality, in 3D, this transforms our
real-valued state tensor $\mathbf{u}_h \in \mathbb{R}^{C \times N \times N
\times N}$ into a \emph{complex-valued coefficient tensor} $\hat{\mathbf{u}}_h
\in \mathbb{C}^{C \times N \times N \times (N//2 + 1)}$ with $N//2$ using
\emph{integer division}. Note that the real-valued Fourier transform
approximately halved the trailing axis. Effectively, $\hat{\mathbf{u}}_h$ still
occupies approximately the same memory since it is complex-valued. Moreover, let
us denote by $\mathcal{F}_h^{-1}$ the real-valued inverse fast Fourier
transform.

The entries in $\hat{\mathbf{u}}_h$ correspond to coefficients of specific modes
to which we can associate a wavenumber grid. For the right-most axis (that was
halved by the real-valued transform), we have $\tilde{\mathbf{k}}_h = [0, 1, 2,
\dots, N//2]$ (which has $N//2 + 1$ entries). For all other axes, we have
$\mathbf{k}_h = [-N//2, \dots, -2, -1, 0, 1, 2, \dots, N//2-1]$ with the full
$N$ entries. The complete wavenumber grid is built as a tensor product similarly
to how the spatial grid can be constructed.

Then, we can take the $d$-th derivative in Fourier space simply by
elementwise multiplying the coefficient state $\hat{\mathbf{u}}_h$ by $\left(i
\frac{2\pi}{L} \mathbf{k}_h\right)^{\circ d}$ (the $(\cdot)^{\circ d}$ shall denote
elementwise exponentiation). This allows for a spectral differentiation when
employing forward and reverse transformations, e.g., for the second derivative
in $x$-direction via
\begin{equation}\label{eq:fourier-derivative}
    \left(\frac{\partial^2 u}{\partial x^2}\right)_h
    =
    \mathcal{F}_h^{-1} \left(
        \left(i \frac{2 \pi}{L} \mathbf{k}_x \right)^{\circ 2}
        \odot
        \mathcal{F}_h \left(
            \mathbf{u}_h
        \right)
    \right).
\end{equation}
Here, $\odot$ denotes elementwise multiplication and $\mathbf{k}_x$ is the
wavenumber grid corresponding to the $x$-axis expanded with singleton axes,
i.e., in 3D it is $\mathbf{k}_x \in \mathbb{R}^{1 \times N \times 1 \times 1}$.

Spectral derivatives are \emph{exact} if the state is bandlimited
\citep{boyd2001chebyshev}. Bandlimited means that the state $\mathbf{u}_h$ does
not carry energy in modes higher than the Nyquist frequency $N//2$.
For discontinuous functions or ones with sharp gradients (i.e., states that are
spectrally rich), spectral differentation can be less efficient than finite
difference counterparts. Taking spectral derivatives is dominated by the cost of
the Fourier transform, which scales $\mathcal{O}\left(N^D \log(N)\right)$. All
operations in Fourier space are elementwise. Hence, they scale
$\mathcal{O}\left(N^D\right)$.

We can apply the Fourier transform to the PDE in \cref{eq:general-pde} and get
\begin{equation}\label{eq:PDE-in-spectral-space}
    \frac{\partial \hat{u}}{\partial t}
    =
    \hat{\mathcal{L}} \hat{u}
    +
    \hat{\mathcal{N}}(\hat{u}).
\end{equation}
The linear operator in Fourier space $\hat{\mathcal{L}}$ is straightforwardly
expressed by composing the elementwise operations. We can devise a
\emph{pseudo-spectral} evaluation strategy to approximate the nonlinear operator
in Fourier space $\hat{\mathcal{N}}_h$, which involves taking derivatives in
coefficient space, and evaluating nonlinearities point-wise in state space. This
requires employing appropriate dealiasing \citep{orszag1971elimination}. For the
convection nonlinearity of the Burgers equation in \cref{eq:burgers-equation-1d},
we get
\begin{equation}\label{eq:pseudo-spectral-evaluation}
    \hat{\mathcal{N}}_h(\hat{\mathbf{u}}_h)
    =
    -
    \left(i \frac{2\pi}{L}\right)
    \odot
    \mathbf{M}_h
    \odot
    \mathcal{F}_h \left(
        \left[
            \mathcal{F}_h^{-1}\left(
                \mathbf{M}_h \odot \mathbf{u}_h
            \right)
        \right]^{\circ 2}
    \right).
\end{equation}
Here, we employ a pre- and post-dealiasing strategy using the boolean mask
$\mathbf{M}_h$ that follows the 2/3 rule of \citet{orszag1971elimination}, i.e.,
it is true for all modes that are smaller than 2/3 of the Nyquist mode $N//2$.

\subsection{Temporal Discretization}
\label{sec:foundations:numerical:temporal}

Given the spatial discretization, we can write down the ordinary differential
equation (ODE) system for the state of degrees of freedom
\begin{equation}\label{eq:ODE-system}
    \frac{d \mathbf{u}_h}{d t}
    =
    \mathcal{L}_h \mathbf{u}_h
    +
    \mathcal{N}_h(\mathbf{u}_h),
\end{equation}
or, similarly, in spectral coefficient space as in
\cref{eq:PDE-in-spectral-space}. The goal of time integration is to advance from
the state at time level $n$ to time level $n+1$ by absorbing the discretized
spatial operators $\mathcal{L}_h$ and $\mathcal{N}_h(\cdot)$ and a time
integration method into the evolution operator $\mathcal{P}_h$.
For simplicity, we will here only consider approaches with a fixed $\Delta t$
and limit us to three classes of single-step methods.

\subsubsection{Explicit Schemes}
\label{sec:foundations:numerical:temporal:explicit}

Explicit methods only use information from time level $n$ to advance in time. In
its simplest form, we could use an \emph{explicit Euler integration},
\begin{equation}
    \mathbf{u}_h^{[n+1]} = \mathbf{u}_h^{[n]} + \Delta t \left(
        \mathcal{L}_h \mathbf{u}_h^{[n]}
        +
        \mathcal{N}_h\left(\mathbf{u}_h^{[n]}\right)
    \right).
\end{equation}
More sophisticated Runge-Kutta methods are possible~\citep{hairer1993solving}.
In most contexts, there is a stability requirement on $\Delta t$ which is
oftentimes coupled with $\Delta x$ and constitutive properties of the PDE.

\subsubsection{Implicit Schemes}
\label{sec:foundations:numerical:temporal:implicit}

Implicit methods use information from both time levels to advance in time. The
simplest form would be an \emph{implicit Euler scheme},
\begin{equation}
    \mathbf{u}_h^{[n+1]}
    -
    \Delta t \left(
        \mathcal{L}_h \mathbf{u}_h^{[n+1]}
        +
        \mathcal{N}_h\left(\mathbf{u}_h^{[n+1]}\right)
    \right)
    =
    \mathbf{u}_h^{[n]}.
\end{equation}
It is called an \emph{implicit} method because
% similar to the PDE in equation TODO,
it does not immediately induce a closed-form solution for
$\mathbf{u}_h^{[n+1]}$. Instead, it prescribes a constraint that
$\mathbf{u}_h^{[n+1]}$ must fulfill. This triggers linear and nonlinear solves
that we will discuss in section \cref{sec:foundations:iterative}. Those solvers
resolve the residuum $\mathcal{R}_h\left(\mathbf{u}_h^{[n+1]};
\mathbf{u}_h^{[n]}\right) \overset{!}{=} \mathbf{0}$ for $\mathbf{u}_h^{[n+1]}$.

Implicit schemes are often unconditionally stable in
theory~\citep{wanner1996solving}. However, they require resolving linear or
nonlinear systems of equations that may become ill-conditioned for larger time
steps. Hence, there are also practical limitations. Moreover, they tend to
introduce more numerical diffusion than explicit schemes, which is noticeable for
hyperbolic problems \citep{leveque2007finite}.

\subsubsection{Exponential Time Differencing (ETDRK)}
\label{sec:foundations:numerical:temporal:etdrk}

In general, the solution to an ordinary differential equation can be expressed
as
\begin{equation}
    \mathbf{u}_h^{[n+1]}
    =
    \mathbf{u}_h^{[n]} + \int_{n \Delta t}^{(n+1) \Delta t}
        \mathcal{L}_h \mathbf{u}_h(\tau)
        +
        \mathcal{N}_h\left(\mathbf{u}_h(\tau)\right)
    \mathrm{d} \tau.
\end{equation}
If the ODE was fully linear, i.e., $\mathcal{N}_h(\cdot) = 0$, we could solve
this integral in closed form and get $\mathbf{u}_h^{[n+1]} = e^{\Delta t
\mathcal{L}_h} \mathbf{u}_h^{[n]}$. This idea can be combined with a Runge-Kutta
approximation to the nonlinearity in the integral \citep{cox2002exponential}.
These methods introduce an integrating factor. The simplest form would be the
ETDRK1 scheme
\begin{equation}
    \mathbf{u}_h^{[n+1]}
    =
    e^{\Delta t \mathcal{L}_h} \mathbf{u}_h^{[n]}
    +
    \Delta t \varphi(\Delta t \mathcal{L}_h) \mathcal{N}_h(\mathbf{u}_h^{[n]}),
\end{equation}
in which $\varphi(z) = (e^z - 1)/z$. The key idea is to treat the linear part of
the PDE exactly by applying the integrating factor $e^{\Delta t \mathcal{L}_h}$
to the state at time level $n$. The nonlinear part is then integrated using a
suitable quadrature rule, which in the case of ETDRK1 is a simple forward Euler
step. Higher-order ETDRK methods, that use more sophisticated quadrature rules,
tend to achieve better accuracy \citep{montanelli2020solving}. For most
practical cases, evaluating the matrix exponential $e^{\Delta t \mathcal{L}_h}$
is not feasible. However, for a Fourier-spectral discretization, this
exponential is evaluated mode-wise, which is remarkably cheap. This builds the
basis for the \verb|Exponax| solvers (see
\cref{sec:foundations:implementation:exponax}) that underpins \verb|APEBench|
(\cref{ch:apebench}).

\section{Implicit Integrators, Stationary Problems and Iterative Solves}
\label{sec:foundations:iterative}

Implicit time integration requires solving linear or nonlinear systems of
equations. The same algorithmic machinery also applies to stationary
(steady-state) PDEs of the form
\begin{equation}\label{eq:stationary-pde}
    \mathcal{L} u
    +
    \mathcal{N}(u)
    =
    0.
\end{equation}
A canonical example is the Poisson equation $\Delta u = - f$
(corresponding to $\mathcal{L}=\Delta$ and $\mathcal{N}(u) = f$).
This reveals a natural connection between implicit time integration and
stationary solvers: as the time step $\Delta t \to \infty$, any implicit time
integrator reduces to a solver for the stationary problem. Conversely, stationary
solvers can be viewed as time integrators that converge in a single, infinitely
large step.

In both settings, the task reduces to finding a state $\mathbf{u}_h^*$ at which
a \emph{residual operator} $\mathcal{R}_h$ vanishes:
$\mathcal{R}_h(\mathbf{u}_h^*) = \mathbf{0}$. The concrete form of
$\mathcal{R}_h$ depends on the problem. For stationary PDEs, it encodes the
discretized PDE constraint. In the case of implicit time integration, it depends
on the particular scheme and typically involves the previous state. In the
purely linear case, the residual reduces to $\mathcal{R}_h(\mathbf{u}_h) =
\mathbf{A}_h \mathbf{u}_h - \mathbf{b}_h$.

While direct methods exist for certain problems, most practical scenarios
require \emph{iterative approaches}. Let $\Phi_h(\cdot)$ be an iteration
operator that produces a sequence of states $\mathbf{u}_h^{(k+1)} =
\Phi_h\left(\mathbf{u}_h^{(k)}\right)$ starting from an initial guess $\mathbf{u}_h^{(0)}$.
We use superscript parentheses $(k)$ to denote iterates and superscript brackets
$[n]$ for time steps. In the case of implicit time integration, these are
nested: $\mathbf{u}_h^{[n+1],(k)}$ denotes the $k$-th iterate when resolving time
step $n+1$.

Ideally, the iterates converge to the solution $\mathbf{u}_h^*$. Convergence can
be measured by the (relative) \emph{optimality error}
${\left\|\mathbf{u}_h^{(k)} -
\mathbf{u}_h^*\right\|}/{\left\|\mathbf{u}_h^*\right\|}$. Since the exact
solution is generally unavailable, a practical proxy is the (relative)
\emph{residual error}
${\left\|\mathcal{R}_h(\mathbf{u}_h^{(k)})\right\|}/{\left\|\mathcal{R}_h(\mathbf{u}_h^{(0)})\right\|}$.
The iteration terminates when the relative residual drops below a prescribed
tolerance $\epsilon$ (typically $\epsilon = 10^{-5}$ for single-precision
arithmetic) or a maximum number of iterations $K$ is reached. We denote by
$K_\epsilon$ the number of iterations required to achieve tolerance $\epsilon$.

\subsection{Linear Solvers}
\label{sec:foundations:iterative:linear}

Linear systems $\mathbf{A}_h \mathbf{u}_h = \mathbf{b}_h$ arise when using
implicit time integration on \emph{linear PDEs} (i.e., $\mathcal{N}(\cdot) =
0$), when resolving \emph{linear stationary problems} (such as the Poisson
equation), or as a substep within nonlinear solvers. For high-rank tensor state
representations, the matrix-vector product $\mathbf{A}_h \mathbf{u}_h$ must be
understood as a linear map. Equivalently, one could reshape $\mathbf{u}_h$ into
a vector in $\mathbb{R}^{CN^D}$ with $\mathbf{A}_h \in \mathbb{R}^{CN^D \times
CN^D}$. In most cases, $\mathbf{A}_h$ is \emph{sparse}, with only
$\mathcal{O}(CN^D)$ non-zero entries out of $(CN^D)^2$.

Oftentimes, there are assembly routines $\Lambda(\cdot)$ and $\beta(\cdot)$ that
build system matrix and right-hand side. For example, for a Poisson equation,
$\beta$ would build the right-hand side $\mathbf{b}_h$ out of the forcing term
$f$. For time stepping, the previous state might affect how matrix and
right-hand side are built. More details on their specifics can be found in the
appendix of the PRDP publication (see \cref{paper:prdp}).

Given $\mathbf{A}_h \mathbf{u}_h = \mathbf{b}_h$, several families of solvers exist,
each with different trade-offs in computational cost, convergence behavior, and
applicability.

\begin{description}
    \item[Direct solvers] \label{sec:foundations:iterative:linear:direct}
        employ matrix decompositions (e.g., LU or Cholesky)
        combined with forward and backward substitution \citep{golub2013matrix}.
        They are best suited for small or moderately sized, oftentimes dense,
        problems. Although, sparse variants do also exist.
    \item[Relaxation methods] \label{sec:foundations:iterative:linear:relaxation}
        use additive splittings of $\mathbf{A}_h$ and
        preferentially damp high-frequency error components. Their convergence
        is exponentially linear, with a rate governed by the spectral radius of
        the iteration matrix. Examples include Jacobi, Gauss-Seidel, and SOR
        methods \citep{saad2003iterative}. They are frequently used as smoothers
        within multigrid methods.
    \item[Krylov methods] \label{sec:foundations:iterative:linear:krylov}
        construct iterates in expanding Krylov subspaces~\citep{saad2003iterative}.
        Preconditioners are typically employed to accelerate convergence.
    \item[Multigrid methods] \label{sec:foundations:iterative:linear:multigrid}
        exploit the smoothing property of relaxation
        methods by operating across a hierarchy of resolutions: either
        geometrically via coarser meshes or algebraically via projected
        operators. They can achieve optimal $\mathcal{O}(N^D)$ complexity for
        elliptic problems such as the Poisson equation, where information must
        propagate across the entire domain \citep{briggs2000multigrid}.
\end{description}

For linear systems with non-singular $\mathbf{A}_h$, the solution
$\mathbf{u}_h^*$ is unique and any initial guess $\mathbf{u}_h^{(0)}$ converges
to the same result. In practice, one therefore typically initializes with
$\mathbf{u}_h^{(0)} = \mathbf{0}$. A common strategy is to combine solver
paradigms. For instance, a Krylov method preconditioned by multigrid (which
itself employs relaxation sweeps) or by a sparse incomplete factorization. This
hybridization of complementary solver paradigms foreshadows the neural-hybrid
methods discussed in \cref{sec:emulators:hybrid}.

\subsection{Nonlinear Solvers}
\label{sec:foundations:iterative:nonlinear}

For nonlinear problems, the goal is to find a root of the \emph{nonlinear}
residual operator, $\mathcal{R}_h(\mathbf{u}_h) \overset{!}{=} \mathbf{0}$. A
straightforward approach is to recast this as a minimization of
$\|\mathcal{R}_h(\mathbf{u}_h)\|_2^2$, leading to a gradient-descent-type
iteration $\Phi_h\left(\mathbf{u}_h^{(k)}\right) = \mathbf{u}_h^{(k)} - \mu\,
\mathcal{R}_h(\mathbf{u}_h^{(k)})$, analogous to a Richardson iteration for
linear systems~\citep{saad2003iterative}. While simple, this approach typically
converges slowly. This subsection introduces two popular strategies for faster
root finding.

Unlike linear solvers, the initial guess matters for both convergence speed and
which root is found. For time-stepping problems, a natural choice is to
warm-start from the previous time step, $\mathbf{u}_h^{[n+1],(0)} =
\mathbf{u}_h^{[n]}$, which places the initial guess close to the solution and
enables fast convergence.

\subsubsection{Newton-Raphson}
\label{sec:foundations:iterative:nonlinear:newton}
A more efficient method than the aforementioned gradient descent approach
linearizes the residual via a first-order Taylor expansion, yielding the
iteration
\begin{equation}
    \Phi_h(\mathbf{u}^{(k)}_h)
    =
    \mathbf{u}^{(k)}_h
    -
    \left[
        \partial_0 \mathcal{R}_h(\mathbf{u}_h^{(k)})
    \right]^{-1}
    \mathcal{R}_h(\mathbf{u}_h^{(k)}).
\end{equation}
Each Newton step requires solving a linear system with the Jacobian $\mathbf{J}
= \partial_0 \mathcal{R}_h(\mathbf{u}_h^{(k)})$. This matrix often need not be
explicitly assembled: automatic differentiation (e.g., \verb|jax.linearize|) can
provide a matrix-free Jacobian-vector product, sufficient for Krylov-type linear
solvers (Jacobian-free Newton-Krylov, JFNK)~\citep{blondel2021efficient}. Close
to the root $\mathbf{u}_h^*$, Newton's method achieves quadratic convergence.
Quasi-Newton variants such as Broyden's method approximate the Jacobian update,
trading quadratic convergence for reduced per-iteration cost, analogous to
(L-)BFGS in nonlinear optimization~\citep{nocedal2006numerical}.

\subsubsection{Picard Iteration}
\label{sec:foundations:iterative:nonlinear:picard}
Certain residual operators admit a \emph{quasilinear} structure
\begin{equation}
    \mathcal{R}_h(\mathbf{u}_h)
    =
    \Lambda(\mathbf{u}_h)\,
    \mathbf{u}_h
    -
    \beta(\mathbf{u}_h),
\end{equation}
where the assembly routines $\Lambda(\cdot)$ and $\beta(\cdot)$ depend
nonlinearly on the state but the state itself enters linearly. This suggests an
iteration that alternates between assembly and linear solve,
\begin{equation}
    \Phi_h\left(\mathbf{u}_h^{(k)}\right)
    =
    \left[
        \Lambda\left(\mathbf{u}_h^{(k)}\right)
    \right]^{-1}
    \beta\left(\mathbf{u}_h^{(k)}\right).
\end{equation}
This Picard iteration can be viewed as a simplification of Newton-Raphson that
drops certain Jacobian terms \citep{turek1999efficient}. Its advantage is
the explicit control over the matrix assembly process, including its sparsity
pattern which is something less straightforward when using automatic
differentiation naively. The trade-off is a reduction from quadratic to linear
convergence.

\subsection{Differentiating Through Iterative Solvers}
\label{sec:foundations:iterative:differentiation}

Explicit time integrators
(\cref{sec:foundations:numerical:temporal:explicit})
compose differentiable operations. Hence, automatic differentiation engines like
JAX or PyTorch can propagate gradients directly. For iterative solvers, the
situation is more nuanced.
One option is to \emph{unroll} the computation graph of $K$ iterations of
$\Phi_h$. Given contractivity and smoothness of the iterator, the Jacobian
estimate converges alongside the primal iterates~\citep{gilbert1992automatic}.
However, from a practical perspective, it is often preferable to differentiate
\emph{over} the converged solve rather than \emph{through} the individual
iterations~\citep{blondel2021efficient,blondel2024elements}. This is
particularly important when the solver is provided by a third-party library
without native autodiff support.

For a nonlinear solve, let $\mathcal{R}_h(\mathbf{u}_h;\mathbf{g})
\overset{!}{=} \mathbf{0}$ be the optimality condition, parameterized by
$\mathbf{g}$ (e.g., the previous time step $\mathbf{u}_h^{[n]}$). Implicit
differentiation of this condition yields the Jacobian
\begin{equation}
    \frac{\partial \mathbf{u}_h}{\partial \mathbf{g}}
    = -\left[
        \partial_0 \mathcal{R}_h(\mathbf{u}_h;\mathbf{g})
    \right]^{-1}
    \partial_1 \mathcal{R}_h(\mathbf{u}_h;\mathbf{g}),
\end{equation}
which requires solving a linear solve, regardless of how many iterations
the forward solve took. This can be used to implement custom forward- or
reverse-mode differentiation rules\footnote{Interactive reference tables for
explicit and implicit autodiff primitive rules are available online. See
\url{https://fkoehler.site/autodiff-table/} and
\url{https://fkoehler.site/implicit-autodiff-table/}.}. For more details, we
refer to the Appendix of the PRDP publication in \cref{paper:prdp}.

\section{Spectral Error Analysis}
\label{sec:foundations:spectral}

The consistency error of the finite difference derivative approximation in
\cref{eq:sample-three-point-stencil} used $\mathcal{O}((\Delta x)^2)$ to bound
the value. Due to spatial mixing, finding a closed-form expression of the
approximation error is challenging. Fortunately, if we limit ourselves to
periodic boundary conditions, we can use a Fourier-based analysis. By the
convolution theorem \citep{boyd2001chebyshev}, the kernel $\mathbf{d}_h$  acts
\emph{element-wise} in Fourier space, leading to mode-wise error analysis. This
allows not only to investigate the \emph{spatial discretization error} but also
the mode-wise error if we combine it with a temporal integration scheme. As
such, we can precisely identify what errors the evolution operator
$\mathcal{P}_h$ introduces, and how constitutive parameters (like diffusivity
$\nu$), numerical scheme $\mathbb{M}$, and discretization choces $N$ and $\Delta
t$ affect them. Moreover, we can use exact spectral differentiation with exact
exponential time differnece to obtain a \emph{reference} for what each mode
multiplier should be. It is this analytical tractability that motivates the
restriction to periodic, uniform Cartesian grids.

For simplicity, we will here only consider the linear one-dimensional diffusion
equation equation $\partial_t u = \nu \Delta u$ on a periodic domain. This
discussion is extended in the emulator superiority paper of
\cref{paper:superiority}. Let $\phi = m / N \in [0, 1/2]$ denote the relative
wavenumber (associated with the mode index $m$) and $\gamma_2 = 2 \nu N^2 \Delta
t/L^2$ the dimensionless diffusion parameter (a CFL-like number). Each time-stepping
scheme maps a Fourier mode $\hat{u}_m^{[n]}$ to $\hat{u}_m^{[n+1]}$ by
multiplying with the multiplier (amplification factor) relevant for index $m$.
Below we list the multipliers for three different schemes. First, we find the
exact or \emph{analytical} multiplier\footnote{Interestingly, since
Fourier-spectral ETDRK methods can solve (bandlimited) linear PDEs on periodic
grids \emph{exactly}, the analytical multiplier in
\cref{eq:diffusion-multiplier-exact} is precisely the one realized by the
Exponax solver of \cref{sec:foundations:implementation:exponax}.}
$\hat{\alpha}_\phi$
\begin{equation}\label{eq:diffusion-multiplier-exact}
    \hat{\alpha}_\phi
    = \exp\!\left(- \frac{\gamma_2}{2} (2\pi\phi)^2 \right).
\end{equation}
The FTCS scheme (forward Euler in time, central differences in space) yields
\begin{equation}\label{eq:diffusion-multiplier-ftcs}
    \hat{e}_\phi
    = 1 + \gamma_2 \left(\cos(2\pi\phi) - 1\right),
\end{equation}
which is stable only for $\gamma_2 \leq 1$. The implicit BTCS scheme (backward
Euler in time, central differences in space) gives
\begin{equation}\label{eq:diffusion-multiplier-btcs}
    \hat{\iota}_\phi
    = \frac{1}{1 - \gamma_2 \left(\cos(2\pi\phi) - 1\right)},
\end{equation}
which is unconditionally stable.%

Comparing $\hat{e}_\phi$ and $\hat{\iota}_\phi$ against the analytical
multiplier $\hat{\alpha}_\phi$ reveals a characteristic pattern. The BTCS scheme
always \emph{under}-attenuates, i.e., it diffuses less than the analytical
solution. Both of its error sources (the implicit Euler step and the central
difference stencil) act in the same direction. For the FTCS scheme, the two
sources compete: the forward Euler \emph{over}-attenuates, while the central
difference stencil under-attenuates. Hence, the net sign depends on $\gamma_2$
and the considered mode $\phi$. Both schemes are exact at $\phi = 0$ (the mean
mode) and deviate most strongly near the Nyquist frequency $\phi = 1/2$. The
discrepancy between the numerical and analytical multipliers constitutes the
\emph{discretization error} per mode. This mode-wise perspective will be central
to understanding the emulator superiority phenomenon in
\cref{sec:relationship:superiority}, where we extend the same analysis to
learned emulators and show that their inductive biases can implicitly correct
for these structured numerical errors (see \cref{fig:multiplier-curves} for a
combined visualization of the multipliers across a range of $\phi$ values).

\section{Modern Implementations: Solvers in Machine Learning Frameworks}
\label{sec:foundations:implementation}

Classically, numerical solvers were implemented in fast and compiled languages
like Fortran or C++. The advent of multicore CPUs and multi-node clusters
required extending these codes with parallelization frameworks like OpenMP or
MPI, giving rise to well-established suites such as Eigen~\citep{eigenweb},
PETSc~\citep{petsc-web-page}, or OpenFOAM~\citep{weller1998tensorial}.
Meanwhile, most deep learning research is conducted in Python using machine
learning frameworks like PyTorch~\citep{paszke2019pytorch} or
JAX~\citep{jax2018github}. These frameworks build on the tradition of MATLAB and
NumPy: wrapping efficient linear algebra routines in a high-level interface.
Additionally, they abstract hardware execution across CPUs and accelerators like
GPUs, and provide automatic differentiation engines that form the core pillar of
gradient-based learning~\citep{baydin2018automatic}. Many also ship optimizing
compilers (e.g., XLA for JAX) that produce efficient fused kernels for diverse
devices.

A natural approach to data generation for physics-based deep learning is to
employ proven solvers, write datasets to disk, and load them in the machine
learning framework of choice. However, this gives rise to a \emph{two-language
problem}: one cannot query a solver on-the-fly, nor differentiate through it to
build neural-hybrid surrogates.%
\footnote{Efforts exist to bridge this gap. For example, Enzyme is an autodiff
engine operating on the LLVM intermediate representation, enabling programs to
exchange gradient information even across different frontend languages
\citep{moses2020instead}. \citet{aehle2022ForwardModeAD} presents an effort to
apply automatic differentiation to compiled programs. Tesseract provides a
standardized interface for composing differentiable components across frameworks
\citep{haefner2025tesseract}. In a classical multi-physics setting, preCICE is a
coupling library for simulations across different codes and languages
\citep{chourdakis2022precice}.}

An alternative is to \emph{reimplement} solvers directly within machine learning
frameworks or modern languages like Julia~\citep{bezanson2017julia}. This
approach unlocks several benefits, including backend-agnostic execution across
CPUs, GPUs, and other accelerators with minimal boilerplate. Moreover, native
autodiff compatibility enables sensitivity analysis, design optimization, chaos
studies (e.g., Lyapunov exponents), neural-hybrid architectures, and using
physics Jacobians as regularizers. Such reimplementations also demonstrate that
classical solvers can be surprisingly competitive on modern
accelerators~\citep{mcgreivy2024weak}.

Challenges remain, however. Wrapping existing third-party solvers (and making
them differentiable) is nontrivial, and certain algorithms (like multigrid on
unstructured grids) do not map straightforwardly to the tensor-style programming
models of current ML frameworks. The broader ecosystem of established simulation
codes (FEniCS~\citep{logg2012automated}, deal.II~\citep{arndt2021dealII},
SU2~\citep{economon2016su2}, Dedalus~\citep{burns2020dedalus}, among others)
represents decades of development that cannot easily be replicated.

Still, there is a growing list of solvers in modern ML frameworks. Most
prominently, this includes \verb|PhiFlow| \citep{holl2024phiflow} that initially
translated computer graphics fluid simulation \citep{stam1999stable} techniques
to TensorFlow, and then generalized to a wider array of physics phenomena and
simulation techniques. \citet{kochkov2021machine} introduced \verb|jax-cfd|,
which has later been ported to spectral methos by \citet{dresdner2022learning}.
\citet{bezgin2023jaxfluids} contributed \verb|JAX-Fluids| which is a
higher-order finite volume solver that has been shown to scale to 1024 TPU cores
\citep{bezgin2025jaxfluids2}. The work of \citet{xue2023jaxfem} with
\verb|JAX-FEM| is a first approach to porting the finite element machinery to
JAX. PICT~\citep{franz2025pict} is a GPU-accelerated PISO solver with a direct
and differentiable interface to PyTorch. \citet{winchenbach2026diffsph}
presented a differentiable SPH-simulator in PyTorch.

\subsection{Why JAX?}
\label{sec:foundations:implementation:jax}

JAX's functional programming model is a natural fit for numerical time steppers
$\mathcal{P}_h$, which are inherently stateless maps from one state to the next.
Its \verb|vmap| transformation enables automatic vectorization: one writes a
solver for a single instance and batches it over thousands of initial conditions
or parameter configurations with a single line of code. Similarly,
\verb|lax.scan| provides a clean abstraction for time-stepping loops. JAX's
composable higher-order derivatives make Hessians straightforward, which is
valuable for PDE-constrained optimization and sensitivity analysis. Its explicit
handling of pseudo-random number generation ensures bitwise reproducibility
across runs, an important property for scientific computing. From a performance
perspective, JAX's XLA compiler fuses kernel operations automatically, producing
efficient code for CPUs, GPUs, and TPUs without manual optimization.

\subsection{Exponax: Differentiable Spectral Solvers}
\label{sec:foundations:implementation:exponax}

Exponax is a JAX library implementing ETDRK solvers (
\cref{sec:foundations:numerical:temporal:etdrk}) for semi-linear PDEs on periodic
domains. By relying exclusively on Fourier spectral methods for spatial
discretization, it maps efficiently to GPUs and TPUs by using highly optimized
FFT routines. When using substeps, it can also handle moderately nonlinear
systems such as the Navier-Stokes equations at low-to-medium Reynolds numbers.
For more details, see the appendix of the APEBench in \cref{paper:apebench}.

The key practical advantage is speed. Together with JAX's \verb|vmap| for
automatic vectorization over initial conditions, Exponax can generate large
datasets in seconds on modern accelerators. Since no implicit solves are
involved, automatic differentiation is unambiguous, i.e., there are no iterative
convergence tolerances that could affect gradient quality (see
\cref{sec:foundations:iterative:differentiation,sec:relationship:fidelity:prdp}).
Exponax is heavily inspired by the Chebfun library \citep{driscoll2014chebfun},
which is a reliable implementation of (Fourier-)spectral ETDRK methods in MATLAB.
Chebfun served as an early data generator for works such as the original
PINN~\citep{raissi2019physics} and FNO~\citep{li2020fourier} publications.

\subsection{Picardax: Differentiable Finite Difference Solvers}
\label{sec:foundations:implementation:picardax}

Picardax implements time-implicit solvers in JAX~\citep{koehler2026picardax}. It
supports $\theta$-method time stepping, encompassing explicit Euler
($\theta=0$), implicit Euler ($\theta=1$), and Crank-Nicolson ($\theta=1/2$)
\citep{turek1999efficient}. Using a general linearized form, the library
automatically extracts the system matrix assembly $\Lambda$ and right-hand side
assembly $\beta$ from the PDE discretization, then dispatches to a fixed-point
(Picard) iterator with configurable convergence criteria. It is built around the
idea of differentiable \emph{implicit} time integrators (see
\cref{sec:foundations:iterative:differentiation}).

Beyond its role as a solver, Picardax serves as a research tool for studying the
interplay between iterative solver convergence and gradient quality. This
capability directly underpins the PRDP work (full publication in \cref{paper:prdp}): by
controlling the number of solver iterations, one can investigate how
under-converged solves affect the training signal. Moreover, Picardax enables
the study of selective gradient cuts. For instance, deactivating backpropagation
through the assembly routine $\Lambda$ in a Picard-style iteration, a practice
common in CFD applications~\citep{list2022learned}. In this sense, Picardax
connects PDE simulation to the broader bilevel optimization literature.

\subsection{Chaotax: Inferring Dynamical System Properties with Autodiff in JAX}
\label{sec:foundations:implementation:chaotax}

The Lyapunov exponent $\lambda$ quantifies the exponential error divergence of a
dynamical system via
\begin{equation}
    \left\|\mathbf{u}_h^{[n]} - \mathbf{u}_h^{[n],ref}\right\|
    \approx
    e^{\lambda n \Delta t} 
    \left\|\mathbf{u}_h^{[0]} - \mathbf{u}_h^{[0],ref}\right\|.
\end{equation}
If $\lambda$ is positive, the dynamical system (like in the case of the
integration of a time-dependent PDE) is considered
\emph{chaotic}~\citep{gilpin2021chaos}. As such, time integration via
$\mathcal{P}_h$ is sensitive to small changes in the initial condition. Small
errors inevitably grow exponentially. We can approximate the largest Lyapunov
exponent numerically by integrating the \emph{tangent-linear} system for a
perturbation alongside the integration of the primal system. The primal system
is integrated by repeatedly calling $\mathcal{P}_h$. We can use
\emph{forward-mode automatic differentiation} for the Jacobian-vector-product
effect of $J_{\mathcal{P}_h}$ \citep{balcerzak2025lyapunov}.

Similarly, one can obtain the full Lyapunov spectrum $\{\lambda_1, \lambda_2,
..., \}$ of $\mathcal{P}_h$ with an automatic differentiation approach
\citep{benettin1980lyapunov,balcerzak2025lyapunov}. It requires integrating a
collection of tangent linear systems next to the primal one (i.e., thereby
combining both the automatic differentiation and automatic vectorization of
JAX). A full spectrum is helpful to categorize dynamics into chaotic,
dissipative, or hyperbolic and to find intrinsic properties of chaotic
attractors like the Kaplan-Yorke dimension \citep{gilpin2021chaos}.

Since the solvers \verb|Exponax| and \verb|Picardax| are implemented in JAX,
using autodiff to evaluate corresponding Lyapunov properties is straightforward.
With \verb|Chaotax| we provide a lightweight and efficient
implementation~\citep{koehler2026chaotax}. Beyond dynamics categorization,
Lyapunov exponents inform meaningful evaluation horizons for autoregressive
emulators, as discussed in
\cref{sec:relationship:benchmarking:rollout} and in \citet{mikhaeil2022chaotic}.

\chapter{Neural Emulators for PDEs}
\label{ch:emulators}

In the previous chapter, we abstracted the solution process of a partial
differential equation using the evolution operator $\mathcal{P}_h$ (either as a
time stepper or the solver to a stationary problem). This chapter considers the
neural emulator $f_\theta$ that is trained to mimic the simulator's behavior,
i.e., $f_\theta \approx \mathcal{P}_h$. This requires constructing a functional
form, i.e., a neural architecture, and a training procedure to select ``optimal''
$\theta$.

Crucially, we only consider neural surrogates which, similar to $\mathcal{P}_h$,
have all the modeling and discretization choices \emph{baked into them}. In
other words, they operate for one PDE, under one parameterization, a single type
of boundary conditions, a fixed resolution $N$, and, for temporal problems, a
fixed $\Delta t$. Many works try to be more general and, e.g., allow the model
to be \emph{conditioned} on constitutive parameters or identify them based on a
history input. This can be generalized to allow for a model to solve many
systems $\left\{\mathbb{P}_i\right\}_{i=1}^P$ and smoothly interpolate between
them \citep{herde2024poseidon}. We will discuss this more in
\cref{sec:outlook:future:foundation}.

By design, we restrict ourselves to full-field surrogates meant to
\emph{replace} the numerical solver $\mathcal{P}_h$ entirely.
Several orthogonal directions exist. \emph{Physics-Informed Neural Networks}
(PINNs) \citep{raissi2019physics} learn the solution function $u(t,
\mathbf{x})$ of \cref{eq:general-pde} directly by optimizing residual-based
losses and \emph{can} work without supervision. Machine learning components can
also replace classical heuristic elements in simulation pipelines, such as flux
limiters \citep{discacciati2020controlling}.

\section{Neural Architectures as Learned Discretizations}
\label{sec:emulators:neural}

\begin{figure}[t]
    \centering
    \begin{tikzpicture}[
        % Global node styles
        gridnode/.style={circle, fill=TUMGray, draw=TUMGrayLight!50,
                         minimum size=2.5mm, inner sep=0pt},
        active-strong/.style={circle, fill=TUMBlueLight!70, draw=TUMBlue!80,
                              minimum size=2.5mm, inner sep=0pt},
        active/.style={circle, fill=TUMBlueLight!40, draw=TUMBlue!60,
                       minimum size=2.5mm, inner sep=0pt},
        active-weak/.style={circle, fill=TUMBlueLight!20, draw=TUMBlue!30,
                            minimum size=2.5mm, inner sep=0pt},
        target/.style={circle, fill=TUMBlue, draw=TUMBlueDark,
                       minimum size=2.5mm, inner sep=0pt},
        freqnode/.style={circle, fill=orange!30, draw=orange!60,
                         minimum size=2mm, inner sep=0pt},
        arc/.style={-{Stealth[length=2pt]}, TUMBlue!70, thin},
    ]

    % Column x-positions
    \def\colA{0}
    \def\colB{3.85}
    \def\colC{7.7}
    \def\colD{11.55}
    % Row y-positions
    \def\rowA{0}
    \def\rowB{-2.9}
    % Node spacing
    \def\ns{0.35}

    % =====================================================================
    % (a) ConvNet
    % =====================================================================
    \begin{scope}[shift={(\colA,\rowA)}]
        \node[font=\small\bfseries, text=TUMBlueDark] at (1.225,1.4) {(a) ConvNet};
        % Input row
        \foreach \i in {0,...,7} {
            \ifnum\i=2 \node[active] (ai\i) at (\i*\ns, 0) {};
            \else\ifnum\i=3 \node[active] (ai\i) at (\i*\ns, 0) {};
            \else\ifnum\i=4 \node[active] (ai\i) at (\i*\ns, 0) {};
            \else \node[gridnode] (ai\i) at (\i*\ns, 0) {};
            \fi\fi\fi
        }
        % Output row
        \foreach \i in {0,...,7} {
            \ifnum\i=3 \node[target] (ao\i) at (\i*\ns, 0.8) {};
            \else \node[gridnode] (ao\i) at (\i*\ns, 0.8) {};
            \fi
        }
        % Arcs
        \draw[arc] (ai2.north) -- (ao3.south);
        \draw[arc] (ai3.north) -- (ao3.south);
        \draw[arc] (ai4.north) -- (ao3.south);
        % Kernel label
        \draw[TUMBlue!50, thin, decorate, decoration={brace, mirror, amplitude=2pt}]
            ([yshift=-1.5mm]ai2.south) -- ([yshift=-1.5mm]ai4.south)
            node[midway, below, font=\tiny, yshift=-1.5pt, text=TUMBlue!70] {$k{=}3$};
    \end{scope}

    % =====================================================================
    % (b) SWIN  [local windowed attention]
    % =====================================================================
    \begin{scope}[shift={(\colB,\rowA)}]
        \node[font=\small\bfseries, text=TUMBlueDark] at (1.225,1.4) {(b) SWIN};
        % Input row
        \foreach \i in {0,...,7} {
            \ifnum\i=2 \node[active] (gi\i) at (\i*\ns, 0) {};
            \else\ifnum\i=3 \node[active] (gi\i) at (\i*\ns, 0) {};
            \else\ifnum\i=4 \node[active] (gi\i) at (\i*\ns, 0) {};
            \else\ifnum\i=5 \node[active] (gi\i) at (\i*\ns, 0) {};
            \else \node[gridnode] (gi\i) at (\i*\ns, 0) {};
            \fi\fi\fi\fi
        }
        % Output row
        \foreach \i in {0,...,7} {
            \ifnum\i=3 \node[target] (go\i) at (\i*\ns, 0.8) {};
            \else \node[gridnode] (go\i) at (\i*\ns, 0.8) {};
            \fi
        }
        % Window backgrounds
        \begin{scope}[on background layer]
            % Window 1: nodes 0-3
            \fill[TUMGray!50, rounded corners=1.5pt]
                (-0.1, -0.15) rectangle (3*\ns+0.1, 0.15);
            \fill[TUMGray!50, rounded corners=1.5pt]
                (-0.1, 0.65) rectangle (3*\ns+0.1, 0.95);
            % Window 2: nodes 4-7
            \fill[TUMGray!50, rounded corners=1.5pt]
                (4*\ns-0.1, -0.15) rectangle (7*\ns+0.1, 0.15);
            \fill[TUMGray!50, rounded corners=1.5pt]
                (4*\ns-0.1, 0.65) rectangle (7*\ns+0.1, 0.95);
        \end{scope}
        % Local attention within window containing target
        \foreach \i in {0,...,3} {
            \draw[arc, TUMBlue!40] (gi\i.north) -- (go3.south);
        }
        % Annotation
        \node[font=\tiny, text=TUMGrayLight, align=center] at (1.225, -0.4)
            {windows shift\\[-1pt]between layers};
    \end{scope}

    % =====================================================================
    % (c) Dilated Conv
    % =====================================================================
    \begin{scope}[shift={(\colC,\rowA)}]
        \node[font=\small\bfseries, text=TUMBlueDark] at (1.225,1.4) {(c) Dilated ResNet};
        % Layer 0 (bottom, input)
        \foreach \i in {0,...,7} {
            \ifnum\i=1 \node[active] (ci\i) at (\i*\ns, -0.1) {};
            \else\ifnum\i=3 \node[active] (ci\i) at (\i*\ns, -0.1) {};
            \else\ifnum\i=5 \node[active] (ci\i) at (\i*\ns, -0.1) {};
            \else\ifnum\i=7 \node[active] (ci\i) at (\i*\ns, -0.1) {};
            \else \node[gridnode] (ci\i) at (\i*\ns, -0.1) {};
            \fi\fi\fi\fi
        }
        % Layer 1 (middle)
        \foreach \i in {0,...,7} {
            \ifnum\i=2 \node[active] (cm\i) at (\i*\ns, 0.4) {};
            \else\ifnum\i=3 \node[active] (cm\i) at (\i*\ns, 0.4) {};
            \else\ifnum\i=4 \node[active] (cm\i) at (\i*\ns, 0.4) {};
            \else \node[gridnode] (cm\i) at (\i*\ns, 0.4) {};
            \fi\fi\fi
        }
        % Layer 2 (top, output)
        \foreach \i in {0,...,7} {
            \ifnum\i=3 \node[target] (co\i) at (\i*\ns, 0.9) {};
            \else \node[gridnode] (co\i) at (\i*\ns, 0.9) {};
            \fi
        }
        % Arcs: layer 1 -> layer 2 (d=1, kernel=3)
        \draw[arc] (cm2.north) -- (co3.south);
        \draw[arc] (cm3.north) -- (co3.south);
        \draw[arc] (cm4.north) -- (co3.south);
        % Arcs: layer 0 -> layer 1 (d=2)
        \draw[arc] (ci1.north) -- (cm3.south);
        \draw[arc] (ci3.north) -- (cm3.south);
        \draw[arc] (ci5.north) -- (cm3.south);
        % Dilation labels
        \node[font=\tiny, text=TUMGrayLight] at (-0.4, 0.15) {$d{=}2$};
        \node[font=\tiny, text=TUMGrayLight] at (-0.4, 0.65) {$d{=}1$};
    \end{scope}

    % =====================================================================
    % (d) UNet  [hierarchical multi-scale]
    % =====================================================================
    \begin{scope}[shift={(\colD,\rowA)}]
        \node[font=\small\bfseries, text=TUMBlueDark] at (1.225,1.4) {(d) UNet};
        % Input row (bottom, full resolution)
        \node[active-weak] (di0) at (0*\ns, -0.1) {};
        \node[active] (di1) at (1*\ns, -0.1) {};
        \node[active-strong] (di2) at (2*\ns, -0.1) {};
        \node[active-strong] (di3) at (3*\ns, -0.1) {};
        \node[active-strong] (di4) at (4*\ns, -0.1) {};
        \node[active-strong] (di5) at (5*\ns, -0.1) {};
        \node[active] (di6) at (6*\ns, -0.1) {};
        \node[active-weak] (di7) at (7*\ns, -0.1) {};
        % Coarse row (middle, 4 nodes)
        \node[active] (dc0) at (0.5*\ns, 0.4) {};
        \node[active] (dc1) at (2.5*\ns, 0.4) {};
        \node[active] (dc2) at (4.5*\ns, 0.4) {};
        \node[active] (dc3) at (6.5*\ns, 0.4) {};
        % Output row (top, full resolution)
        \foreach \i in {0,...,7} {
            \ifnum\i=3 \node[target] (do\i) at (\i*\ns, 0.9) {};
            \else \node[gridnode] (do\i) at (\i*\ns, 0.9) {};
            \fi
        }
        % Trapezoidal regions (pooling / upsampling)
        \fill[TUMBlue, opacity=0.06]
            (di0.north west) -- (dc0.south west) -- (dc3.south east) -- (di7.north east) -- cycle;
        \fill[TUMBlue, opacity=0.06]
            (dc0.north west) -- (do0.south west) -- (do7.south east) -- (dc3.north east) -- cycle;
        % Skip connections
        \draw[TUMBlue!40, densely dashed, thin]
            (di1.north) to[bend left=35] (do1.south);
        \draw[TUMBlue!40, densely dashed, thin]
            (di6.north) to[bend right=35] (do6.south);
        % Labels
        \node[font=\tiny, text=TUMGrayLight] at (-0.5, 0.15) {pool};
        \node[font=\tiny, text=TUMGrayLight] at (-0.7, 0.65) {upsample};
    \end{scope}

    % =====================================================================
    % (e) ViT  [patch-based global]
    % =====================================================================
    \begin{scope}[shift={(\colA,\rowB)}]
        \node[font=\small\bfseries, text=TUMBlueDark] at (1.225,1.4) {(e) ViT};
        % Input row with patch grouping
        \foreach \i in {0,...,7} {
            \ifnum\i=2 \node[active] (fi\i) at (\i*\ns, 0) {};
            \else\ifnum\i=3 \node[active] (fi\i) at (\i*\ns, 0) {};
            \else \node[active-weak] (fi\i) at (\i*\ns, 0) {};
            \fi\fi
        }
        % Output row
        \foreach \i in {0,...,7} {
            \ifnum\i=3 \node[target] (fo\i) at (\i*\ns, 0.8) {};
            \else\ifnum\i=2 \node[target] (fo\i) at (\i*\ns, 0.8) {};
            \else \node[gridnode] (fo\i) at (\i*\ns, 0.8) {};
            \fi\fi
        }
        % Patch backgrounds
        \begin{scope}[on background layer]
            \foreach \p in {0,1,2,3} {
                \pgfmathsetmacro{\xl}{\p*2*\ns - 0.1}
                \pgfmathsetmacro{\xr}{\p*2*\ns + \ns + 0.1}
                \fill[TUMGray!50, rounded corners=1.5pt]
                    (\xl, -0.15) rectangle (\xr, 0.15);
                \fill[TUMGray!50, rounded corners=1.5pt]
                    (\xl, 0.65) rectangle (\xr, 0.95);
            }
        \end{scope}
        % Inter-patch attention arcs (all patches to target patch)
        \foreach \p in {0,1,2,3} {
            \pgfmathsetmacro{\src}{\p*2*\ns + 0.5*\ns}
            \draw[arc, TUMBlue!40] (\src, 0.15) -- (2.5*\ns, 0.65);
        }
    \end{scope}

    % =====================================================================
    % (f) FNO  [spectral global]
    % =====================================================================
    \begin{scope}[shift={(\colB,\rowB)}]
        \node[font=\small\bfseries, text=TUMBlueDark] at (1.225,1.4) {(f) FNO};
        % Input row (bottom)
        \foreach \i in {0,...,7} {
            \node[active-weak] (bi\i) at (\i*\ns, -0.15) {};
        }
        % Frequency row (middle)
        \foreach \i in {0,...,4} {
            \node[freqnode] (bf\i) at (0.35+\i*0.35, 0.4) {};
        }
        % Output row (top)
        \foreach \i in {0,...,7} {
            \ifnum\i=3 \node[target] (bo\i) at (\i*\ns, 0.95) {};
            \else \node[gridnode] (bo\i) at (\i*\ns, 0.95) {};
            \fi
        }
        % Trapezoidal regions (FFT / iFFT)
        \fill[TUMBlue, opacity=0.06]
            (bi0.north west) -- (bf0.south west) -- (bf4.south east) -- (bi7.north east) -- cycle;
        \fill[TUMBlue, opacity=0.06]
            (bf0.north west) -- (bo0.south west) -- (bo7.south east) -- (bf4.north east) -- cycle;
        % Labels
        \node[font=\tiny, text=TUMGrayLight] at (-0.35, 0.12) {FFT};
        \node[font=\tiny, text=TUMGrayLight] at (-0.4, 0.68) {iFFT};
    \end{scope}

    % =====================================================================
    % (g) Transformer  [fully global]
    % =====================================================================
    \begin{scope}[shift={(\colC,\rowB)}]
        \node[font=\small\bfseries, text=TUMBlueDark] at (1.225,1.4) {(g) Transformer};
        % Input row (intensity encodes typical learned attention decay)
        \foreach \i in {0,...,7} {
            \ifnum\i=3 \node[active-strong] (ei\i) at (\i*\ns, 0) {};
            \else\ifnum\i=2 \node[active] (ei\i) at (\i*\ns, 0) {};
            \else\ifnum\i=4 \node[active] (ei\i) at (\i*\ns, 0) {};
            \else \node[active-weak] (ei\i) at (\i*\ns, 0) {};
            \fi\fi\fi
        }
        % Output row
        \foreach \i in {0,...,7} {
            \ifnum\i=3 \node[target] (eo\i) at (\i*\ns, 0.8) {};
            \else \node[gridnode] (eo\i) at (\i*\ns, 0.8) {};
            \fi
        }
        % All-to-one arcs
        \foreach \i in {0,...,7} {
            \draw[arc, TUMBlue!40] (ei\i.north) -- (eo3.south);
        }
    \end{scope}

    % =====================================================================
    % (h) Legend
    % =====================================================================
    \begin{scope}[shift={(\colD,\rowB)}]
        \node[font=\small\bfseries, text=TUMBlueDark] at (1.225,1.4) {Legend};
        % Legend entries (shifted up by 0.4 to vertically align with neighbouring panels)
        \node[target] (lt) at (0.3, 1.15) {};
        \node[font=\scriptsize, right=3pt of lt, anchor=west] {output (target)};
        \node[active-strong] (las) at (0.3, 0.85) {};
        \node[font=\scriptsize, right=3pt of las, anchor=west] {strong contribution};
        \node[active] (la) at (0.3, 0.55) {};
        \node[font=\scriptsize, right=3pt of la, anchor=west] {moderate contribution};
        \node[active-weak] (law) at (0.3, 0.25) {};
        \node[font=\scriptsize, right=3pt of law, anchor=west] {weak contribution};
        \node[gridnode] (lg) at (0.3, -0.05) {};
        \node[font=\scriptsize, right=3pt of lg, anchor=west] {inactive position};
        \draw[arc] (0.15, -0.32) -- (0.45, -0.32);
        \node[font=\scriptsize, right=3pt, anchor=west] at (0.45, -0.32) {information flow};
        \node[freqnode] (lf) at (0.3, -0.59) {};
        \node[font=\scriptsize, right=3pt of lf, anchor=west] {frequency mode};
    \end{scope}

    \end{tikzpicture}
    \caption{%
        Connectivity patterns of neural architectures used as PDE emulators,
        shown as 1D schematics. Each panel highlights which input positions
        contribute to a single output position (dark blue) in one layer or block;
        shading intensity indicates contribution strength.
        Panels are ordered by locality of the receptive field:
        (a)~ConvNet ($\approx$~finite differences) uses a fixed local stencil;
        (b)~SWIN attends within shifted windows;
        (c)~Dilated ResNet grows the receptive field
        through dilation;
        (d)~UNet ($\approx$~multigrid) processes a hierarchy of resolutions with
        skip connections;
        (e)~ViT applies global attention
        between patches;
        (f)~FNO ($\approx$~spectral methods) mixes all positions through the
        Fourier domain;
        (g)~Transformer uses full self-attention
        over every position.
    }
    \label{fig:architecture-paradigms}
\end{figure}
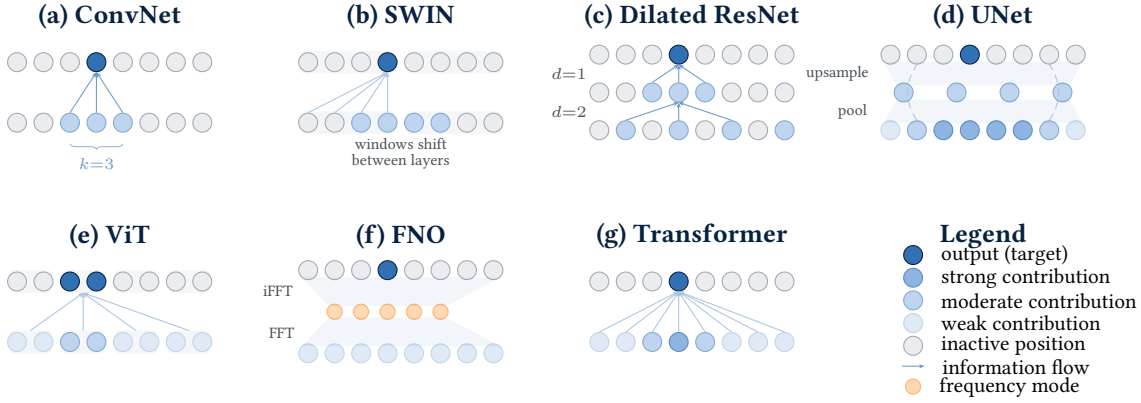

In this section, we will introduce the most popular architectures for neural
emulators on structured Cartesian grids. All of them first lift the input state
$\mathbf{u}_h$ in the channel axis, e.g., in 1D $\mathbf{u}_h \in \mathbb{R}^{C
\times N} \mapsto \mathbf{u}_h^{[l=0]} \in \mathbb{R}^{C_{\text{lifted}} \times
N}$. Afterward, this state is processed.
%  either sequentially, hierarchically, or with residual computations.
Ultimately, it is projected back to the output shape. In our case, we often have
that the output has the same number of channels as the input since the models
operate autoregressively. In \cref{fig:architecture-paradigms}, we visually
compare how the different architectures process spatial dependencies.

\subsection{Convolutional Networks and Finite Difference Stencils}
\label{sec:emulators:neural:conv}

A convolutional layer applies a learned filter via cross-correlation. This is
the same operation used to approximate spatial derivatives in finite difference methods                                                                                                                                                    
(\cref{sec:foundations:numerical:spatial}). In fact, a single convolutional layer                                                                                                                                                   
with kernel size $K=3$, no bias, and no nonlinearity is \emph{exactly} a finite                                                                                                                                                     
difference stencil with learned coefficients \citep{long2018pde}. Convolutional
Networks (ConvNets or CNNs) apply a sequence of convolutional layers on the
lifted representation \citep{lecun1989backpropagation}. What a convolutional
network adds beyond a single stencil is depth,                                                                                                                                                                 
channel mixing, and nonlinearity. Each layer has
$\mathcal{O}(C_{\text{lifted}}^2 K^D)$ parameters and
$\mathcal{O}(C_{\text{lifted}}^2 K^D N^D)$ compute. Channel mixing enables the
network to couple multiple physical quantities (e.g., velocity components and
pressure), much as a system of PDEs couples multiple fields. Pointwise
nonlinearities after each layer allow the network to represent nonlinear
operators. Stacking $L$ layers with kernel size $K=3$ yields a receptive field
of $L$ grid points per direction per dimension.
Residual connections \citep{he2016resnet} and pre-layer normalization
\citep{he2016identity} are essential for training deep networks and can be
interpreted as a learned identity-plus-perturbation structure, echoing the
$\mathbf{I} + \Delta t \, \mathbf{A}$ form of explicit time steppers.

\subsection{Fourier Neural Operators and Spectral Methods}
\label{sec:emulators:neural:fno}

Where convolutional networks mirror finite difference stencils, Fourier Neural
Operators (FNOs) mirror (Fourier) spectral methods. Recall that spectral
differentiation amounts to elementwise multiplication by wavenumber-dependent
coefficients in Fourier space (\cref{eq:fourier-derivative}). An FNO layer does
the same, but with \emph{learned} complex multipliers: the input is transformed
via the FFT, multiplied mode-wise by a learnable weight tensor, and transformed
back \citep{li2020fourier}. Where the spectral method prescribes the multipliers
from the PDE, the FNO learns them from data.

In 1D on a discrete representation with $N$ degrees of freedom, a full
parameterization would require $N//2 + 1$ complex multipliers per channel pair,
which is substantially more parameters than a spatial convolution with kernel
size $K \ll N$. In practice, FNOs retain only the lowest $M \approx 12 \sim 20$
modes and set higher-mode multipliers to zero. This acts as an inherent low-pass
filter. \citet{li2020fourier} argue that residual connections and stacking
multiple Fourier layers with pointwise nonlinearities can recover some
higher-frequency content, but the low-pass character remains a defining trait.
Channel mixing happens in Fourier space, hence, each layer has
$\mathcal{O}(C_{\text{lifted}}^2 M^D)$ parameters. The compute per layer is
$\mathcal{O}(C_{\text{lifted}} N^D \log N + C_{\text{lifted}}^2 M^D)$, typically
dominated by the FFT for multidimensional problems.

A known limitation is that the Fourier basis assumes periodic boundary
conditions. Despite this, FNOs are increasingly applied to non-periodic
problems. Here, the mode truncation appears to reduce the impact of the Gibbs
phenomenon \citep{boyd2001chebyshev}. Alternative strategies include
zero-padding the domain to approximate a periodic extension\footnote{See, e.g.,
the NVIDIA Neural Operator library at
\url{https://github.com/NVIDIA/neuraloperator}.}. Architectures like GINO
address unstructured domains by first mapping to a uniform grid and then
applying FNO processing~\citep{li2023gino}.

Several extensions reduce the cost or increase the flexibility of FNOs.
Factorized FNOs \citep{tran2021factorized} apply the FFT independently per
spatial dimension. Other approaches parameterize the multipliers as continuous
functions of the mode index \citep{xiao2024amortized}.

\subsection{UNets as Multigrid Methods}
\label{sec:emulators:neural:unet}

To overcome the slow growth of receptive field in feedforward ConvNets,
\citet{ronneberger2015u} used convolutions on a \emph{hierarchy} of grid
resolutions with appropriate downsampling and upsampling operations. Given a
kernel size of $3$, a single convolution at hierarchy level $l$ covers $2^l$
grid points per direction at the original resolution. For example, when
coarsening the resolution $l=4$ times, one convolution already spans $16$ grid
points. In contrast, a flat ConvNet would need $16$ layers to achieve this.

This hierarchical structure closely resembles geometric multigrid
(\cref{sec:foundations:iterative:linear}). In multigrid, relaxation sweeps
smooth the error on the current grid, a \emph{restriction} operator projects the
residual to a coarser grid where long-range components are corrected cheaply,
and a \emph{prolongation} operator interpolates the correction back
\citep{briggs2000multigrid}. The UNet follows the same pattern: convolution
blocks at each level act as learned smoothers, downsampling as learned
restriction, upsampling as learned prolongation, and skip connections carry
fine-scale information across the hierarchy much like the coarse-grid
correction in multigrid.

In practice, several choices exist for the downsampling and upsampling
operations. The ``modern UNet'' for PDE emulation \citep{gupta2022towards}
replaces the max-pooling of the original architecture with strided convolutions
for downsampling. An increasingly popular alternative is PixelUnshuffle
(space-to-depth) for downsampling and PixelShuffle (depth-to-space) for
upsampling \citep{shi2016real}: each $r \times r$ spatial patch is rearranged
into $r^2$ channels (or vice versa), followed by a learned linear projection.
An alternative modernization replaces the residual blocks with ConvNeXt blocks
\citep{liu2022convnet}, which separate spatial and channel mixing via depthwise
convolutions with large kernels and use an inverted bottleneck design. This
ConvNeXt-based UNet emerged as the strongest single baseline across the diverse
PDE problems in The Well benchmark \citep{ohana2024well}. In principle, the
hierarchical structure is not limited to convolutional processing. Recent work
has incorporated Fourier layers or attention mechanisms into the UNet hierarchy
\citep{herde2024poseidon,holzschuh2025pde}, just as multigrid can employ
different smoothers at different levels.

\subsection{Transformer-based Architecture}
\label{sec:emulators:neural:transformer}

The previous architectures all encode spatial structure explicitly:
convolutions enforce locality, Fourier layers enforce global periodicity, and
UNets enforce a resolution hierarchy. Attention-based mechanisms take a
different approach. Originally developed for sequence modelling in natural
language processing \citep{bahdanau2015neural,vaswani2017attention}, they
provide a \emph{nonlocal} mechanism to mix in ``space'' (in a language modelling
task, this refers to the tokens of the sequence). As such, they have been
adapted to neural surrogates to replace the spatial or spectral
convolution-based processing. A vanilla transformer architecture can be
ported to work on uniform Cartesian states in 1D, 2D, or 3D by removing the
causal attention mask and employing appropriate positional embeddings using
tensor-product sinusoidal encodings \citep{vaswani2017attention} or the more
modern RoPE \citep{su2024roformer}. Unfortunately, the compute requirements of
attention scale quadratically in the number of tokens, i.e., in a vanilla
approach it scales quadratically with the \emph{total} number of degrees of
freedom, giving $\mathcal{O}(N^{2D})$.

Several strategies reduce this cost. Vision Transformers (ViTs) partition the
grid into non-overlapping patches and treat each patch as a single token, so
that self-attention operates over $\mathcal{O}((N/p)^D)$ tokens instead of $N^D$
\citep{dosovitskiy2021image}. Shifted-window (SWIN) attention restricts
attention to local windows and shifts the window boundaries between layers, so
that tokens at window edges interact in alternating layers, giving the
architecture global connectivity over depth \citep{liu2021swin}.
Factorization of attention across spatial dimensions further reduces cost
\citep{li2023factformer,mccabe2024multiple}. \citet{paischer2025gyroswin}
demonstrated that shifted-window attention can scale to five spatial dimensions.

Since attention does not assume any spatial structure, transformers extend
naturally to unstructured representations with appropriate positional
embeddings \citep{wu2024transolver,alkin2024universal,alkin2025ab,zhdanov2025erwin},
a direction we discuss further in \cref{sec:outlook:future:unstructured}.

Compared to convolutions or Fourier layers, transformers introduce very little
\emph{physical} inductive bias. They do not bake in locality, periodicity, or
translation equivariance. Modern variants like SWIN or hierarchical designs
reintroduce some of these priors, but the vanilla form is almost entirely
domain-agnostic. Despite this, there is growing momentum behind
transformer-based surrogates as the architecture of choice for foundation models
in scientific computing~\citep{ashton2025fluid}, echoing observations from
language modelling~\citep{brown2020language}. Whether physical inductive biases
for neural surrogates remain valuable at sufficient scale, or whether they are
subsumed by sheer model capacity and data diversity, is an open question.

\subsection{Other Types of Architectures}
\label{sec:emulators:neural:other}

Another popular mechanism to increase the receptive field is to \emph{dilate}
the convolutional kernels \citep{yu2016multi-scale,stachenfeld2022learned},
effectively inserting zeros between the kernel values. With a dilation factor
$d$, a kernel of size $K$ spans $d(K-1)+1$ grid points. By stacking layers with
exponentially growing dilation rates, one can cover the full domain in
logarithmically many layers rather than linearly many.

Neural surrogates can be outfitted with an internal iteration. Oftentimes, this
comes in the form of diffusion models or flow matching approaches
\citep{flowsanddiffusions2026}. This seems natural when solving inherently stiff
problems like directly predicting into the turbulent manifold
\citep{lienen2024from,lino2025diffusion}. \citet{kohl2026benchmarking}
investigated whether autoregressive diffusion models can improve long-term
accuracy and stability while \citet{lippe2023pde} argue that a diffusion-based
iterative refinement helps with high-frequency accuracy and potentially
mitigates distribution shift.

On unstructured meshes, Graph Neural Networks (GNNs) exchange messages along
edges of the mesh graph \citep{sanchez2020learning,pfaff2020learning}. Each
message-passing step aggregates information from neighboring nodes. This is a local
operation, much like a convolutional layer on a structured grid. Stacking
multiple layers grows the receptive field analogously to stacking convolutions.
The same idea of hierarchical processing from \cref{sec:emulators:neural:unet}
has been adapted to graphs by introducing coarsened mesh levels
\citep{grigorev2023hood,cao2023bistride}.

\section{Neural-Hybrid Approaches}
\label{sec:emulators:hybrid}

Instead of fully replacing the numerical solver $\mathcal{P}_h$ by $f_\theta$,
we can introduce a coarse simulator $\mathcal{P}_H$ (note the capital $H$ to
indicate it being a coarser form of the fine simulator) and have a neural component
$c_\theta$ \emph{correct} its output or interact with it. The appeal is that
decades of expertise have gone into building numerical solvers
(\cref{ch:foundations}), and a hybrid approach can leverage that structure rather than
re-learning it from scratch. At the same time, operating at the state-discrete
level is more flexible than replacing a single algorithmic component like a flux
limiter \citep{discacciati2020controlling}.

In the simplest case this could be sequential, i.e., $f_\theta = c_\theta \circ
\mathcal{P}_H$ (or $f_\theta = \mathcal{P}_H \circ c_\theta$)
\citep{um2020solver,list2022learned}\footnote{See
\url{https://fkoehler.site/corrector-configurations/} for an overview of
possible ways to combine coarse solvers and correctors including ways to cut
gradients over them.}. More complex interactions are possible, e.g., by having
the correction term interact with the solver's internal state or by having the
correction term interact with the solver's numerics, e.g., a learned finite
volume interpolation \citep{kochkov2021machine} or as a learned forcing term
\citep{wei2025inc}. For example, hybrid approaches have been used in building
fast weather and climate emulators \citep{kochkov2024neuralgcm}.

Given certain learning setups (e.g., under an unrolled setting as we will
discuss in \cref{sec:emulators:training:data}), the combined component
$f_\theta$ is required to be \emph{output-input differentiable} (not just with
respect to the parameters $\theta$). Consequently, this requires the coarse
solver $\mathcal{P}_H$ to be differentiable. 
This introduces practical challenges: differentiable solvers can be expensive to
backpropagate through, and iterative solvers may not fully converge, raising
questions about gradient quality. In \cref{sec:relationship:fidelity}, we show
that full convergence of $\mathcal{P}_H$ is in fact unnecessary during training
and that this can be exploited for significant computational savings.

\section{Training and Testing of Autoregressive Neural Emulators}
\label{sec:emulators:training}

\subsection{Synthetic Data Generation and Training Configurations}
\label{sec:emulators:training:data}

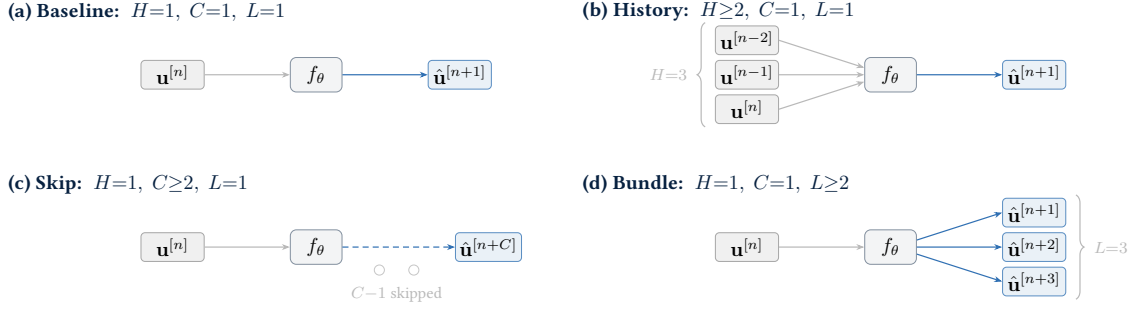
\begin{figure}
    \centering
    \resizebox{\textwidth}{!}{%
    \begin{tikzpicture}[
        % Node styles (matching fig:training-testing-pipeline conventions)
        state/.style={
            rectangle, rounded corners=2pt,
            minimum width=1.1cm, minimum height=0.5cm,
            font=\small, align=center, inner sep=2pt,
        },
        inputstate/.style={state, draw=gray!70, fill=gray!12},
        outputstate/.style={state, draw=TUMBlue!80, fill=TUMBlue!10},
        skipped/.style={circle, draw=gray!45, fill=none, minimum size=5pt, inner sep=0pt},
        netbox/.style={
            rectangle, rounded corners=3pt,
            draw=TUMBlueDark!50, fill=TUMBlueDark!5,
            minimum width=0.9cm, minimum height=0.6cm,
            font=\small, align=center, inner sep=2pt,
        },
        % Arrow styles
        arrow/.style={-{Stealth[length=4pt]}, semithick},
        inarrow/.style={arrow, gray!55},
        outarrow/.style={arrow, TUMBlue},
        % Decorations
        mybrace/.style={
            decorate, decoration={brace, amplitude=4pt, raise=2pt},
            semithick, gray!55,
        },
        panellabel/.style={font=\small\bfseries, text=TUMBlueDark, anchor=west},
        annot/.style={font=\scriptsize, text=gray!55},
    ]

    % Shared x-positions: left column (panels a, c) and right column (panels b, d)
    \def\inX{3.0}     % left-column input
    \def\netX{5.5}    % left-column f_θ
    \def\outX{8.0}    % left-column output
    \def\inXr{13.0}   % right-column input
    \def\netXr{15.5}  % right-column f_θ
    \def\outXr{18.0}  % right-column output

    % Panel y-centers (2x2 grid)
    \def\Ya{0}      \def\Yb{0}
    \def\Yc{-3.0}   \def\Yd{-3.0}

    % =================================================================
    % Panel (a): Baseline — H=1, C=1, L=1
    % =================================================================
    \node[panellabel] at (0, \Ya+1.1) {\textsf{(a)}~Baseline:\; $H{=}1,\; C{=}1,\; L{=}1$};

    \node[inputstate]  (a-in)  at (\inX,  \Ya) {$\mathbf{u}^{[n]}$};
    \node[netbox]      (a-net) at (\netX,  \Ya) {$f_\theta$};
    \node[outputstate] (a-out) at (\outX,  \Ya) {$\hat{\mathbf{u}}^{[n{+}1]}$};

    \draw[inarrow]  (a-in)  -- (a-net);
    \draw[outarrow] (a-net) -- (a-out);

    % =================================================================
    % Panel (b): History — H≥2, C=1, L=1
    % =================================================================
    \node[panellabel] at (10, \Yb+1.1) {\textsf{(b)}~History:\; $H{\geq}2,\; C{=}1,\; L{=}1$};

    % Three inputs stacked vertically
    \node[inputstate] (b-in1) at (\inXr, \Yb+0.6)  {$\mathbf{u}^{[n{-}2]}$};
    \node[inputstate] (b-in2) at (\inXr, \Yb)       {$\mathbf{u}^{[n{-}1]}$};
    \node[inputstate] (b-in3) at (\inXr, \Yb-0.6)   {$\mathbf{u}^{[n]}$};

    \node[netbox]      (b-net) at (\netXr, \Yb) {$f_\theta$};
    \node[outputstate] (b-out) at (\outXr, \Yb) {$\hat{\mathbf{u}}^{[n{+}1]}$};

    % Fan-in arrows to distinct border anchors
    \draw[inarrow] (b-in1.east) -- (b-net.165);
    \draw[inarrow] (b-in2.east) -- (b-net.180);
    \draw[inarrow] (b-in3.east) -- (b-net.195);
    \draw[outarrow] (b-net) -- (b-out);

    % Vertical brace left of inputs
    \draw[mybrace]
        ($(b-in3.south west)+(-0.1,-0.05)$) -- ($(b-in1.north west)+(-0.1,0.05)$)
        node[midway, left=7pt, annot] {$H{=}3$};

    % =================================================================
    % Panel (c): Skip — H=1, C≥2, L=1
    % =================================================================
    \node[panellabel] at (0, \Yc+1.1) {\textsf{(c)}~Skip:\; $H{=}1,\; C{\geq}2,\; L{=}1$};

    \node[inputstate]  (c-in)  at (\inX,     \Yc) {$\mathbf{u}^{[n]}$};
    \node[netbox]      (c-net) at (\netX,    \Yc) {$f_\theta$};
    \node[outputstate] (c-out) at (\outX+0.5, \Yc) {$\hat{\mathbf{u}}^{[n{+}C]}$};

    \draw[inarrow]  (c-in)  -- (c-net);
    \draw[outarrow, densely dashed] (c-net) -- (c-out);

    % Skipped time levels below the dashed arrow
    \node[skipped] at (6.6, \Yc-0.4) {};
    \node[skipped] at (7.2, \Yc-0.4) {};
    \node[annot, anchor=north] at (6.9, \Yc-0.55) {$C{-}1$ skipped};

    % =================================================================
    % Panel (d): Bundle — H=1, C=1, L≥2
    % =================================================================
    \node[panellabel] at (10, \Yd+1.1) {\textsf{(d)}~Bundle:\; $H{=}1,\; C{=}1,\; L{\geq}2$};

    \node[inputstate] (d-in)  at (\inXr,  \Yd)      {$\mathbf{u}^{[n]}$};
    \node[netbox]     (d-net) at (\netXr, \Yd)      {$f_\theta$};

    % Three outputs stacked vertically
    \node[outputstate] (d-out1) at (\outXr, \Yd+0.6) {$\hat{\mathbf{u}}^{[n{+}1]}$};
    \node[outputstate] (d-out2) at (\outXr, \Yd)      {$\hat{\mathbf{u}}^{[n{+}2]}$};
    \node[outputstate] (d-out3) at (\outXr, \Yd-0.6)  {$\hat{\mathbf{u}}^{[n{+}3]}$};

    \draw[inarrow] (d-in) -- (d-net);
    % Fan-out arrows from distinct border anchors
    \draw[outarrow] (d-net.15)  -- (d-out1.west);
    \draw[outarrow] (d-net.0)   -- (d-out2.west);
    \draw[outarrow] (d-net.-15) -- (d-out3.west);

    % Vertical brace right of outputs
    \draw[mybrace]
        ($(d-out1.north east)+(0.1,0.05)$) -- ($(d-out3.south east)+(0.1,-0.05)$)
        node[midway, right=7pt, annot] {$L{=}3$};

    \end{tikzpicture}%
    }

    \caption[Temporal axis taxonomy of autoregressive emulators.]{%
        \textbf{The $(H, C, L)$ temporal axis taxonomy.}
        An autoregressive emulator $f_\theta$ maps $H$ input states to $L$ output
        states, potentially skipping $C{-}1$ intermediate time levels.
        \textbf{(a)}~The standard Markovian baseline ($H{=}C{=}L{=}1$).
        \textbf{(b)}~Using $H{\geq}2$ past states as input enables higher-order
        temporal information, analogous to multi-step numerical methods.
        \textbf{(c)}~A skip length $C{\geq}2$ coarsens the effective time step;
        skipped time levels (hollow circles) are not predicted.
        \textbf{(d)}~Temporal bundling ($L{\geq}2$) produces multiple future states
        simultaneously.
    }
    \label{fig:temporal-axis}
\end{figure}

Following the pipeline of \cref{fig:emulation-pipeline}, we fix a PDE system
$\mathbb{P}$ and choose a numerical method $\mathbb{M}$, resolution $N$, and
time step $\Delta t$ to obtain a discrete solver $\mathcal{P}_h$. Training data
is generated by sampling initial conditions
$\mathbf{u}_h^{[0]} \sim \mathcal{I}_h$ from a prescribed distribution and
rolling them out via the trajectory operator $\mathcal{T}$, producing data
trajectories
\begin{equation}
    \mathcal{D}_h
    =
    \mathcal{T}_{\mathcal{P}_h}(\mathbf{u}_h^{[0]})
    =
    \{\mathbf{u}_h^{[0]}, \mathbf{u}_h^{[1]}, \dots,
    \mathbf{u}_h^{[T]}\},
    \quad
    \mathbf{u}_h^{[n+1]} = \mathcal{P}_h(\mathbf{u}_h^{[n]}).
\end{equation}
Several common modifications apply to the raw trajectories. First, one may
subsample to only retain every $C$-th time step, recording states at a coarser
temporal cadence (see the skip parameter in \cref{fig:temporal-axis}c). Second,
one may discard the first $W$ \emph{warmup} steps and only use
$\{\mathbf{u}_h^{[W]}, \mathbf{u}_h^{[W+1]}, \dots\}$ for training. This ensures
that the training data lies on a desired physically meaningful regime, e.g., a
chaotic attractor.

\begin{figure}
    \centering
    \begin{tikzpicture}[
        % Node styles
        state/.style={
            rectangle, rounded corners=2pt,
            minimum width=1.2cm, minimum height=0.55cm,
            font=\small, align=center, inner sep=2pt,
        },
        solverstate/.style={state, draw=orange!60!black, fill=orange!10},
        emustate/.style={state, draw=TUMBlue!80, fill=TUMBlue!10},
        evalstate/.style={state, draw=green!50!black, fill=green!8},
        icdist/.style={
            ellipse, draw=#1!60!black, densely dashed,
            fill=#1!6, minimum width=1.4cm, minimum height=0.7cm,
            font=\scriptsize, align=center, inner sep=1pt,
        },
        arrow/.style={-{Stealth[length=4pt]}, semithick},
        solverarrow/.style={arrow, orange!70!black},
        emuarrow/.style={arrow, TUMBlue},
        evalarrow/.style={arrow, green!50!black},
        lossarrow/.style={densely dashed, TUMGrayLight, -{Stealth[length=3pt]}},
        errorarrow/.style={densely dotted, green!40!black, -{Stealth[length=3pt]}},
    ]

    % =====================================================================
    % Panel (a): Training
    % =====================================================================
    % --- Solver row (top) ---
    \node[icdist=orange] (icTrain) at (0, 0.8) {$\mathcal{I}_h$};

    \node[font=\small\bfseries, text=TUMBlueDark] at (0, -0.05) {(a) Training};

    % --- Solver row: step 1 full color, steps 2+ faded ---
    \node[solverstate] (s0) at (2.2, 0.8) {$\mathbf{u}_h^{[0]}$};
    \node[solverstate] (s1) at (4.4, 0.8) {$\mathbf{u}_h^{[1]}$};
    \node[solverstate, draw=orange!30!black, fill=orange!5,
          text=orange!40!black] (s2) at (6.6, 0.8) {$\mathbf{u}_h^{[2]}$};
    \node[solverstate, draw=orange!30!black, fill=orange!5,
          text=orange!40!black] (s3) at (8.8, 0.8) {$\mathbf{u}_h^{[3]}$};

    \draw[solverarrow] (icTrain) -- (s0) node[midway, above, font=\scriptsize] {sample};
    \draw[solverarrow] (s0) -- (s1) node[midway, above, font=\scriptsize] {$\mathcal{P}_h$};
    \draw[solverarrow, orange!35!black] (s1) -- (s2)
        node[midway, above, font=\scriptsize, text=orange!35!black] {$\mathcal{P}_h$};
    \draw[solverarrow, orange!35!black] (s2) -- (s3)
        node[midway, above, font=\scriptsize, text=orange!35!black] {$\mathcal{P}_h$};

    % Reference trajectory label
    \node[font=\scriptsize, text=orange!60!black, anchor=west] at (10.0, 0.8) {Reference};

    % --- Emulator row: step 1 full color, steps 2+ faded ---
    \node[emustate] (e0) at (2.2, -0.8) {$\mathbf{u}_h^{[0]}$};
    \node[emustate] (e1) at (4.4, -0.8) {$f_\theta\left(\mathbf{u}_h^{[0]}\right)$};
    \node[emustate, draw=TUMBlue!40, fill=TUMBlue!5,
          text=TUMBlue!50] (e2) at (6.6, -0.8) {$f_\theta^2\left(\mathbf{u}_h^{[0]}\right)$};
    \node[emustate, draw=TUMBlue!40, fill=TUMBlue!5,
          text=TUMBlue!50] (e3) at (8.8, -0.8) {$f_\theta^3\left(\mathbf{u}_h^{[0]}\right)$};

    \draw[emuarrow] (e0) -- (e1) node[midway, below, font=\scriptsize] {$f_\theta$};
    \draw[emuarrow, TUMBlue!40] (e1) -- (e2)
        node[midway, below, font=\scriptsize, text=TUMBlue!40] {$f_\theta$};
    \draw[emuarrow, TUMBlue!40] (e2) -- (e3)
        node[midway, below, font=\scriptsize, text=TUMBlue!40] {$f_\theta$};

    % Emulator label
    \node[font=\scriptsize, text=TUMBlue, anchor=west] at (10.0, -0.8) {Emulator};

    % Shared initial condition (connect solver to emulator start)
    \draw[TUMGrayLight, densely dotted, thick] (s0.south) -- (e0.north);

    % --- Boundary: T=1 vs T>1 ---
    \draw[TUMGrayLight!60, densely dashed, thin]
        (5.2, 1.4) -- (5.2, -1.5);
    \node[font=\tiny, text=TUMGrayLight, anchor=south] at (3.6, 1.4) {$T{=}1$};
    \node[font=\tiny, text=TUMGrayLight!70, anchor=south] at (7.0, 1.4)
        {$T{>}1$ (unrolled)};

    % --- Loss connections: step 1 full, steps 2+ faded ---
    \draw[lossarrow] (s1.south) -- (e1.north)
        node[midway, right, font=\scriptsize, text=TUMGrayLight] {$\zeta$};
    \draw[lossarrow, TUMGrayLight!40] (s2.south) -- (e2.north)
        node[midway, right, font=\scriptsize, text=TUMGrayLight!50] {$\zeta$};
    \draw[lossarrow, TUMGrayLight!40] (s3.south) -- (e3.north)
        node[midway, right, font=\scriptsize, text=TUMGrayLight!50] {$\zeta$};

    % --- Gradient flow: two segments ---
    % T>1 portion (faded purple)
    \draw[purple!35, thick, densely dashed]
        ([yshift=-8pt]e3.south) -- ([yshift=-8pt]e2.south) --
        ([yshift=-8pt]e1.south);
    % T=1 portion (strong purple, with arrowhead to f_theta)
    \draw[purple, thick, densely dashed, -{Stealth[length=4pt]}]
        ([yshift=-8pt]e1.south) -- (3.5, -1.08);
    \node[font=\scriptsize, text=purple, anchor=north] at (5.5, -1.5) {$\nabla_\theta L$};

    % =====================================================================
    % Panel (b): Testing / Evaluation
    % =====================================================================
    \def\panelBshift{-3.5}

    % --- Reference solver row (top) ---
    \node[icdist=green!60!black] (icTest) at (0, \panelBshift+0.8) {$\tilde{\mathcal{I}}_h$};

    \node[font=\small\bfseries, text=TUMBlueDark] at (0, \panelBshift-0.05) {(b) Testing};

    \node[evalstate] (ts0) at (2.2, \panelBshift+0.8) {$\mathbf{u}_h^{[0]}$};
    \node[evalstate] (ts1) at (4.0, \panelBshift+0.8) {$\mathbf{u}_h^{[1]}$};
    \node[font=\small, text=green!40!black] (tsdots) at (5.8, \panelBshift+0.8) {$\cdots$};
    \node[evalstate] (tsT1) at (7.6, \panelBshift+0.8) {$\mathbf{u}_h^{[T{-}1]}$};
    \node[evalstate] (tsT) at (10.0, \panelBshift+0.8) {$\mathbf{u}_h^{[T]}$};

    \draw[evalarrow] (icTest) -- (ts0) node[midway, above, font=\scriptsize] {sample};
    \draw[evalarrow] (ts0) -- (ts1) node[midway, above, font=\scriptsize] {$\tilde{\mathcal{P}}_h$};
    \draw[evalarrow] (ts1) -- (tsdots) node[midway, above, font=\scriptsize] {$\tilde{\mathcal{P}}_h$};
    \draw[evalarrow] (tsdots) -- (tsT1) node[midway, above, font=\scriptsize] {$\tilde{\mathcal{P}}_h$};
    \draw[evalarrow] (tsT1) -- (tsT) node[midway, above, font=\scriptsize] {$\tilde{\mathcal{P}}_h$};

    % --- Emulator row (bottom) ---
    \node[emustate] (te0) at (2.2, \panelBshift-0.8) {$\mathbf{u}_h^{[0]}$};
    \node[emustate] (te1) at (4.0, \panelBshift-0.8) {$f_\theta\left(\mathbf{u}_h^{[0]}\right)$};
    \node[font=\small, text=TUMBlue] (tedots) at (5.8, \panelBshift-0.8) {$\cdots$};
    \node[emustate] (teT1) at (7.6, \panelBshift-0.8) {$f_\theta^{T{-}1}\left(\mathbf{u}_h^{[0]}\right)$};
    \node[emustate] (teT) at (10.0, \panelBshift-0.8) {$f_\theta^T\left(\mathbf{u}_h^{[0]}\right)$};

    \draw[emuarrow] (te0) -- (te1) node[midway, below, font=\scriptsize] {$f_\theta$};
    \draw[emuarrow] (te1) -- (tedots) node[midway, below, font=\scriptsize] {$f_\theta$};
    \draw[emuarrow] (tedots) -- (teT1) node[midway, below, font=\scriptsize] {$f_\theta$};
    \draw[emuarrow] (teT1) -- (teT) node[midway, below, font=\scriptsize] {$f_\theta$};

    % Shared initial condition
    \draw[TUMGrayLight, densely dotted, thick] (ts0.south) -- (te0.north);

    % --- Error connections ---
    \draw[errorarrow] (ts1.south) -- (te1.north)
        node[midway, right, font=\scriptsize, text=green!40!black] {$e^{[1]}$};
    \draw[errorarrow] (tsT1.south) -- (teT1.north)
        node[midway, right, font=\scriptsize, text=green!40!black] {$e^{[T{-}1]}$};
    \draw[errorarrow] (tsT.south) -- (teT.north)
        node[midway, right, font=\scriptsize, text=green!40!black] {$e^{[T]}$};

    % --- Generalization annotations ---
    % Centered below the chains
    \node[rectangle, rounded corners=2pt, draw=TUMGrayLight, fill=TUMGray,
          font=\scriptsize, align=center, inner sep=5pt]
        (genbox) at (5.0, \panelBshift-2.0)
        {%
            \textbf{Generalization axes:}\quad
            \textcolor{purple}{Temporal:} $T \gg 1$\quad
            \textcolor{green!50!black}{State-space:} $\tilde{\mathcal{I}}_h \neq \mathcal{I}_h$\quad
            \textcolor{orange!70!black}{Solver:} $\tilde{\mathcal{P}}_h \neq \mathcal{P}_h$\quad
            \textcolor{TUMBlue}{Metric:} $\tilde{\zeta} \neq \zeta$%
        };

    \end{tikzpicture}

    \caption{%
        Training and evaluation pipeline for autoregressive neural emulators.
        \textbf{(a)}~During training, a numerical solver $\mathcal{P}_h$ produces
        reference trajectories from initial conditions sampled from $\mathcal{I}_h$.
        The emulator $f_\theta$ is trained to match consecutive states; gradient
        flow (purple) may extend through multiple autoregressive steps.
        \textbf{(b)}~At test time, the emulator is rolled out autoregressively for
        many steps and compared against a reference solver $\tilde{\mathcal{P}}_h$.
        Tildes indicate that the test setup may differ from training, probing
        temporal, state-space, solver, or metric generalization.
    }
    \label{fig:training-testing-pipeline}
\end{figure}
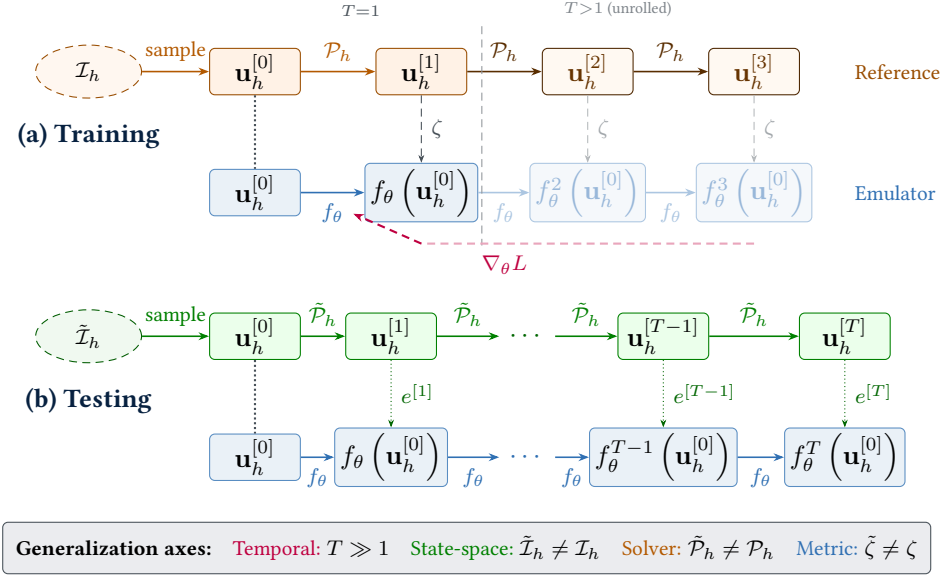

Given such trajectories, the standard one-step supervised training objective is
\begin{equation}\label{eq:one_step_supervised_objective}
    L(\theta) = \mathbb{E}_{
        \{\mathbf{u}_h^{[n]}, \mathbf{u}_h^{[n+1]}\} \sim
        \mathcal{D}_h
    }
    \left[
        \zeta \left(
            f_\theta\left(\mathbf{u}_h^{[n]}\right),
            \mathbf{u}_h^{[n+1]}
        \right)
    \right].
\end{equation}
Here, $\zeta$ is typically the mean squared error, though relative losses like
the relative $L^2$ norm are also common, especially for multi-channel systems
where physical quantities vary in magnitude.
The emulator $f_\theta$ is trained to match the solver $\mathcal{P}_h$
\emph{in expectation} over pairs of consecutive states from a trajectory
(\cref{fig:training-testing-pipeline}a). In practice, the expectation is
approximated using a finite dataset and stochastic minibatches. Then, the
optimization follows a standard deep learning pipeline with momentum-accelerated
stochastic gradient descent \citep{prince2023understanding}, with Adam
\citep{kingma2014adam} being the most common optimizer choice.

Many variations of this one-step supervised setting exist. Most prominently,
\emph{unrolled} training optimizes over trajectory mismatch by applying
$f_\theta$ autoregressively for more than one step and comparing the entire
predicted trajectory against the reference
(\cref{fig:training-testing-pipeline}a, light purple gradient flow). This can improve
long-term stability \citep{kohl2026benchmarking}, but requires backpropagation
through the autoregressive chain.
Additional modifications include weighting individual time steps differently in
the loss, and curriculum strategies that gradually increase the unroll length
during training \citep{grigorev2023hood}.
This exposes the model to its own prediction errors during training and helps
reduce the distribution shift encountered during autoregressive rollout.
Similarly, adding small amounts of noise to the training inputs
\citep{sanchez2020learning,pfaff2020learning} serves a comparable purpose by
perturbing the inputs away from the clean solver trajectory.

The temporal axis taxonomy of \cref{fig:temporal-axis} characterizes the
emulator's input-output structure along three dimensions: the history length
$H$, skip $C$, and bundle size $L$. The baseline setting ($H = C = L = 1$)
corresponds to a Markovian single-step emulator, mirroring one-step numerical
methods like Euler or Runge-Kutta. Using a history $H \geq 2$ is analogous to
multi-step methods (e.g., Adams-Bashforth) and can capture higher-order temporal
information. Interestingly, solid performance is often achieved with the simple
Markovian setting, suggesting that the dynamics of many PDE systems are well
approximated by single-step maps. This is consistent with the prevalence of one-step
methods for first-order PDEs in numerical practice \citep{ruiz2024benefits}.

%

%
% --- Figure: (H,C,L) Temporal Axis Taxonomy ---
%

\subsection{Different Aspects of Generalization}
\label{sec:emulators:training:generalization}

At test time, autoregressive emulators are evaluated by rolling out the model
for multiple time steps and comparing against a reference trajectory
(\cref{fig:training-testing-pipeline}b),
\begin{equation}\label{eq:rollout_test_metric}
    e^{[n]} = \mathbb{E}_{
        \mathbf{u}_h^{[0]} \sim \tilde{\mathcal{I}}_h
    }
    \left[
        \tilde{\zeta} \left(
        f_\theta^n(\mathbf{u}_h^{[0]}), \tilde{\mathcal{P}}_h^n(\mathbf{u}_h^{[0]})
        \right)
    \right].
\end{equation}
Here, \( f_\theta^n \) denotes applying the emulator \( n \) times
autoregressively. The test setup (denoted by tildes) might differ from the
training setup in several ways, probing different aspects of generalization.

\begin{description}
    \item[Temporal/Autoregressive generalization] \label{sec:emulators:training:generalization:temporal}
        involves evaluating long-term
        accuracy and stability by testing for \( n > 1 \), i.e., for more autoregressive
        applications than during training. Exponential error growth \( e^{[n]} \)
        usually signals instability.
    \item[State-space generalization] \label{sec:emulators:training:generalization:statespace}
        examines performance on different initial
        conditions \( \tilde{\mathcal{I}}_h \neq \mathcal{I}_h \) or over
        extended time horizons beyond those used for training data generation,
        where the physics might enter a regime that is not well-represented by
        the training data.
    \item[Metric generalization] \label{sec:emulators:training:generalization:metric}
        refers to evaluating the emulator using a
        different metric \(\tilde{\zeta}\) than the one used for training
        \(\zeta\). This allows for probing specific aspects of performance, such
        as accuracy in higher frequencies/modes.
    \item[Solver generalization] \label{sec:emulators:training:generalization:solver}
        evaluates against a different numerical
        simulator \( \tilde{\mathcal{P}}_h \neq \mathcal{P}_h \), where both
        solvers are consistent with the same underlying PDE
        \eqref{eq:general-pde} but employ different numerical implementations,
        varying in aspects such as scheme order or solver tolerance.
\end{description}

Emulators can be tested for a single type of generalization or a combination of
several. It is important to distinguish these axes clearly: for instance,
autoregressive generalization and state-space generalization are often conflated
in practice, since long rollouts naturally push the emulator into unseen regions
of state space. The generalization box in \cref{fig:training-testing-pipeline}b
summarizes these four axes. The emulator's behavior across these axes can often
be analyzed from a spectral perspective: similar to numerical methods (whose
error behavior is discussed in \cref{sec:foundations:spectral}),
emulators may perform better or worse depending on the spectral composition of
the state.

Beyond the four axes above, one may additionally probe generalization over
resolution, constitutive parameters, or even entirely different PDE
systems. These are questions that become central in the context of foundation models
(\cref{sec:outlook:future:foundation}).

% =============================================================================
% CHAPTER 4: The Solver--Emulator Relationship
% =============================================================================
\chapter{The Solver-Emulator Relationship}
\label{ch:relationship}

This chapter sets out to explain three observations about neural emulators of
PDEs that are difficult to reconcile with a simple data-driven view of
surrogation. First, an emulator trained on trajectories from a given numerical
solver can, at inference time, produce multi-step rollouts whose error against
a higher-fidelity---in some cases analytical---reference is strictly
\emph{smaller} than that of the solver which generated its training data.
Second, differentiable-physics training can reach the same final accuracy at a
fraction of the solver iteration budget that asymptotic convergence analysis
would demand. Third, two apparently identical benchmarks can yield opposing
verdicts about the same emulator under assumptions that are rarely stated
explicitly.

We will argue that all three observations share a common cause.
\Cref{ch:foundations} and \cref{ch:emulators} introduced numerical
solvers and neural emulators as separate paradigms. We already noted structural
parallels between specific architectures and classical methods, e.g., ConvNets
and finite differences (\cref{sec:emulators:neural:conv}), Fourier Neural
Operators and pseudo-spectral methods (\cref{sec:emulators:neural:fno}), UNets
and multigrid (\cref{sec:emulators:neural:unet}). This is in line with a broader
pattern documented in the
literature~\citep{toshev2023relationships,alt2023connections,brandstetter2021message,mccabe2023towards}.
The entanglement, however, runs deeper than architecture. At the level of the
training and evaluation \emph{pipeline} (\cref{sec:emulators:training}), a
``solver'' is not a single object but several: one generates the training data,
a (possibly different) one serves as the reference against which the emulator is
evaluated, a (possibly third) one may be embedded inside the training loop, and
so on. Conflating these silently
is what makes each of the three observations above appear puzzling.

\Cref{sec:relationship:roles} formalizes this picture by identifying five
distinct roles that numerical solvers occupy in the emulator pipeline and the
fidelity relationships among them. Each subsequent section then pins one of
the three observations to a specific role configuration: emulators exceeding
their training source (\cref{sec:relationship:superiority}), training-loop
solvers that need not converge (\cref{sec:relationship:fidelity}), and the
benchmarking consequences that follow from both
(\cref{sec:relationship:benchmarking}).

\section{The Multiple Roles of the Solver}
\label{sec:relationship:roles}

Recall from \cref{ch:foundations} that a single PDE system $\mathbb{P}$ does
not have a unique numerical counterpart. The same governing equations can be
discretized with different methods $\mathbb{M}$, resolutions $N$, and time
steps $\Delta t$. Each choice yields a distinct solver $\mathcal{P}_h$
with its own error structure and other properties. We introduced the notion of
\emph{fidelity}~(\cref{sec:foundations:fidelity}) to compare them. Two solvers
$\mathcal{P}_h$ and $\mathcal{P}_H$ approximating the same $\mathbb{P}$ need
not be equally accurate and can be ordered along a spectrum of fidelity, made
precise in the spectral sense of \cref{sec:foundations:spectral}.

\begin{figure}
    \centering
    \begin{tikzpicture}[
        % Node styles
        pde/.style={
            rectangle, rounded corners=3pt,
            draw=TUMGrayLight, fill=TUMGray,
            minimum width=2.8cm, minimum height=0.85cm,
            font=\small\bfseries, align=center
        },
        solver/.style={
            rectangle, rounded corners=3pt,
            draw=TUMBlue!60, fill=TUMBlue!8,
            minimum width=2.0cm, minimum height=0.7cm,
            font=\small, align=center
        },
        role/.style={
            rectangle, rounded corners=2pt,
            draw=#1!60!black, fill=#1!12,
            minimum width=2.8cm, minimum height=0.9cm,
            font=\scriptsize, align=center
        },
        emulator/.style={
            rectangle, rounded corners=3pt,
            draw=TUMBlue!80, fill=TUMBlue!12,
            minimum width=2.8cm, minimum height=0.85cm,
            font=\small\bfseries, align=center
        },
        arrow/.style={-{Stealth[length=6pt, width=5pt]}, color=TUMGrayLight!80, thick},
        sarrow/.style={-{Stealth[length=5pt, width=4pt]}, color=#1!60!black, semithick},
    ]

    % === Top: Continuous PDE ===
    \node[pde] (pde) at (0, 0)
        {$\mathbb{P}$\\[-1pt]{\scriptsize\mdseries Continuous PDE}};

    % === Discretization annotation ===
    \node[font=\scriptsize, text=TUMGrayLight!80, anchor=west]
        at (2.1, -0.9) {choose $\mathbb{M},\; N,\; \Delta t$};

    % === Solvers ===
    \node[solver] (sA) at (-3.0, -1.8) {$\mathcal{P}_h$};
    \node[solver] (sB) at ( 0.0, -1.8) {$\mathcal{P}_H$};
    \node[solver] (sC) at ( 3.0, -1.8) {$\mathcal{P}_{h'}$};

    \draw[arrow] (pde) -- (sA);
    \draw[arrow] (pde) -- (sB);
    \draw[arrow] (pde) -- (sC);

    % Dots indicating more solvers possible
    \node[font=\small, color=TUMGrayLight]
        at ($(sA)!0.5!(sB)$) {$\cdots$};
    \node[font=\small, color=TUMGrayLight]
        at ($(sB)!0.5!(sC)$) {$\cdots$};

    % === Separator: solvers -> roles ===
    \draw[TUMGrayLight!40, densely dotted]
        (-6.8, -2.6) -- (6.8, -2.6);
    \node[font=\scriptsize, text=TUMGrayLight, fill=white, inner sep=2pt]
        at (0, -2.6) {\emph{each may fill one or more roles}};

    % === Roles (evenly spaced row) ===
    \node[role=orange!80!red] (r1) at (-6.2, -3.8)
        {\textbf{(1)} Training Data\\Source};
    \node[role=green!60!black] (r2) at (-3.1, -3.8)
        {\textbf{(2)} Evaluation\\Reference};
    \node[role=TUMBlue] (r3) at (0, -3.8)
        {\textbf{(3)} Hybrid Training\\Component};
    \node[role=purple] (r4) at (3.1, -3.8)
        {\textbf{(4)} Structural\\Analogue};
    \node[role=TUMGrayLight] (r5) at (6.2, -3.8)
        {\textbf{(5)} Baseline\\Comparator};

    % === Emulator ===
    \node[emulator] (emu) at (0, -6.0)
        {Neural Emulator $f_\theta$};

    % === Role -> Emulator connections ===
    % Solid arrows: roles that feed into the emulator
    \draw[sarrow=orange!80!red]
        (r1.south) -- (emu.165)
        node[midway, left, xshift=-3pt, font=\scriptsize, text=orange!60!black] {trains};
    \draw[sarrow=TUMBlue]
        (r3.south) -- (emu.north)
        node[midway, right, font=\scriptsize, text=TUMBlue!80] {in loop};
    % Double-headed arrows: roles that compare against the emulator
    \draw[{Stealth[length=5pt, width=4pt]}-{Stealth[length=5pt, width=4pt]},
        semithick, color=green!40!black]
        (r2.south) -- (emu.150)
        node[midway, left, font=\scriptsize, text=green!40!black] {evaluates};
    \draw[{Stealth[length=5pt, width=4pt]}-{Stealth[length=5pt, width=4pt]},
        semithick, color=TUMGrayLight]
        (r5.south) -- (emu.15)
        node[midway, right, xshift=7pt, font=\scriptsize, text=TUMGrayLight] {compares};
    % Dashed arrow: conceptual influence
    \draw[sarrow=purple, densely dashed]
        (r4.south) -- (emu.30)
        node[midway, right, xshift=3pt, font=\scriptsize, text=purple!80] {inspires};

    \end{tikzpicture}

    \caption{%
        The multiple roles of numerical solvers in the neural emulator pipeline.
        A continuous PDE system $\mathbb{P}$ admits many valid discretizations
        $\mathcal{P}_h, \mathcal{P}_H, \mathcal{P}_{h'}, \ldots$ obtained by
        choosing a numerical method~$\mathbb{M}$, resolution~$N$, and time
        step~$\Delta t$. Each solver may fill one or more of the five roles
        shown, and different experimental setups activate different subsets
        (see \cref{tab:role-configurations}). Solid arrows denote data flow
        into the emulator, double-headed arrows denote comparison, and the
        dashed arrow denotes conceptual influence.
    }
    \label{fig:solver-roles}
\end{figure}
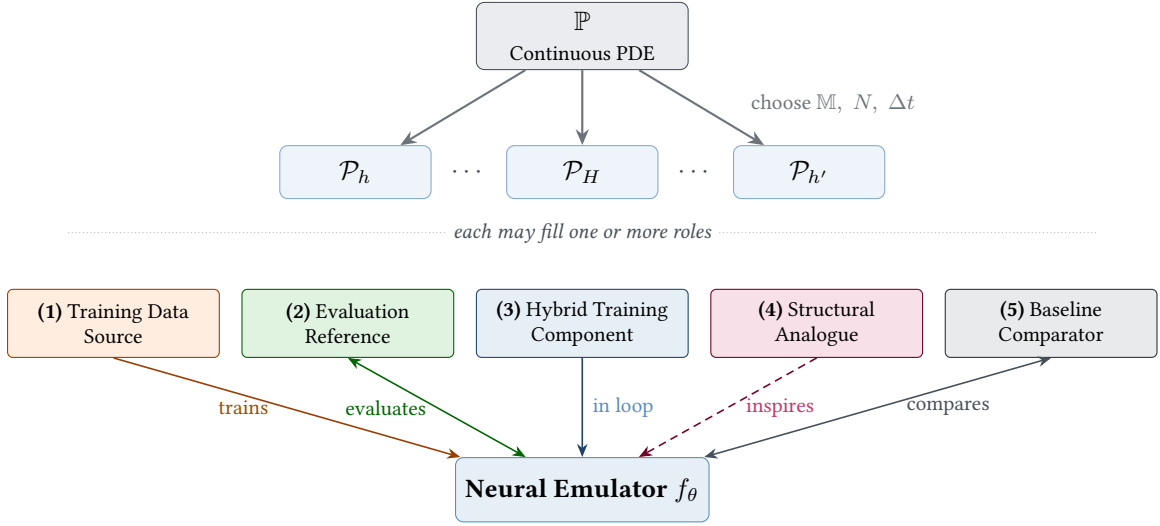

What is specific to neural emulation is that the emulator pipeline silently
involves \emph{multiple} such discretizations at once, at potentially different
fidelities, each performing a different job. We distinguish five roles
(\cref{fig:solver-roles}):

\begin{enumerate}
    \item \textbf{Training data source.} The solver whose trajectories
    $\mathcal{D}_h$ the network is trained on. Its numerical errors enter the
    emulator through the supervised loss. Its fidelity is essentially
    arbitrary: the emulator only sees what the dataset was generated with.
    \item \textbf{Evaluation reference.} The solver against which the emulator
    is scored at test time; this is what ``accuracy'' actually measures. This
    raises the question how its fidelity being different from (1) affects the
    benchmarking of a neural emulator.
    \item \textbf{Hybrid or in-the-loop component.} A differentiable solver
    called inside the training loop, either as part of a neural-hybrid
    architecture or as a physics-informed component through which gradients
    flow. Its fidelity can be chosen flexibly (often coarser than (1)) since its
    job is to work together with a neural component, not to produce reference
    trajectories.
    \item \textbf{Structural analogue.} The classical numerical method that the
    architecture structurally resembles. Unlike the other roles, (4) does not
    participate computationally in the pipeline. It provides an inductive bias.
    \item \textbf{Baseline comparator.} The solver against which the emulator is
    compared on the accuracy--speed trade-off. Its fidelity should match the
    claim being made: comparing against a very coarse solver benefits the
    emulator, against a very fine one penalizes it.
\end{enumerate}

Two observations about role~(4) deserve emphasis, since it is qualitatively
different from the others. First, it is conceptual rather than computational. 
No solver needs to exist in memory for role~(4) to be filled. It is enough
that the architecture is structurally analogous to one.
Second, role~(4) can be essentially \emph{vacant}: for architectures with no
clear classical counterpart (e.g., vanilla transformers are the canonical case)
no numerical method supplies such a bias, and any plausibility prior must come
from elsewhere. This can be the data distribution, the training objective, or
explicit physical constraints. Both observations will reappear in the
superiority analysis of \cref{sec:relationship:superiority}.

Not every experimental setup activates all five roles, and the fidelity
relationships among the active ones can vary. In the default setup of training
on trajectories and evaluating against held-out trajectories from the same
solver, roles~(1), (2), and~(5) collapse onto a single solver, role~(3) is
absent, and role~(4) is whatever the architecture provides. Superiority
investigations (\cref{sec:relationship:superiority}) deliberately break this
by introducing a strictly higher-fidelity reference~(2) distinct
from~(1). PRDP (\cref{sec:relationship:fidelity}) introduces
role~(3) with intentionally low and varying fidelity. Finally, an
\emph{unfair benchmark} arises when role~(2) is not separately available at
all. This is the case for most fixed, published benchmarks such as
PDEBench~\citep{takamoto2022pdebench}, where the data-generating solver is not
exposed as a configurable component. As such, roles~(1),~(2), and~(5) collapse
opaquely into the dataset itself and there is no way to tell whether the
emulator merely reproduces its training source's errors.
\Cref{tab:role-configurations} summarises these configurations alongside a
neural-hybrid variant and the symbolic conventions used in the rest of the
chapter.

\begin{table}[htpb]
    \centering
    \caption{%
        Which solver roles are active under different experimental setups.
        \checkmark~indicates the role is active and filled by a distinct
        solver; ``$=$\,(k)'' indicates the role collapses onto the solver
        filling role~(k); ``$<\!\dots\!<$'' indicates progressive refinement
        of fidelity during training; ``---'' indicates the role is absent;
        ``(hidden)'' indicates the role is not exposed in the setup.
    }
    \label{tab:role-configurations}
    {\footnotesize
    \begin{tabular}{@{}lccccc@{}}
        \toprule
        \textbf{Setup} & \textbf{(1)} & \textbf{(2)} & \textbf{(3)} & \textbf{(4)} & \textbf{(5)} \\
        \tumrule
        Default (\cref{sec:emulators:training:data})
                                     & \checkmark & $=$\,(1) & ---        & arch-dep & $=$\,(1) \\
        Neural-Hybrid                & \checkmark & $=$\,(1) & \checkmark & arch-dep & $=$\,(1) \\
        \midrule
        Superiority investigation    & \checkmark & \checkmark & ---     & aligned  & $=$\,(1) \\
        PRDP                         & \checkmark & $=$\,(1) & \checkmark\,$<\!\dots\!<$\,\checkmark & arch-dep & $=$\,(1) \\
        Unfair benchmark             & \checkmark & (hidden) & ---        & arch-dep & $=$\,(1) \\
        \bottomrule
    \end{tabular}
    }
\end{table}

\section{When Emulators exceed their Training Source}
\label{sec:relationship:superiority}

The conventional assumption is that a neural emulator trained on data from a
numerical solver can, at best, reproduce the accuracy of that solver. The
emulator superiority result challenges this: under certain conditions, the
emulator's inductive biases perform implicit regularization that corrects for
structured numerical artifacts in the training data. While this may not hold in
all practical scenarios, it is a call to be cautious about where training data
comes from and what one evaluates against.

\subsection{Spectral Error Analysis of Learned Emulators}
\label{sec:relationship:superiority:spectral}

\begin{figure}
    \centering
    \begin{tikzpicture}
    \begin{axis}[
        width=0.88\textwidth,
        height=6.5cm,
        xlabel={Relative wavenumber $\phi = m/N$},
        ylabel={Multiplier},
        xmin=0, xmax=0.5,
        ymin=0, ymax=1.15,
        % Grid
        grid=major,
        grid style={TUMGray!50},
        % Axis styling
        axis lines=left,
        tick label style={font=\small, /pgf/number format/fixed},
        label style={font=\small},
        % Legend
        legend style={
            at={(0.02,0.02)}, anchor=south west,
            font=\small,
            draw=TUMGrayLight,
            fill=white,
            fill opacity=0.9,
            text opacity=1,
            rounded corners=2pt,
            cells={anchor=west},
        },
    ]

    % === Exact/Analytical (black, thick) ===
    % exp(-gamma2/2 * (2*pi*phi)^2) with gamma2=0.5
    \addplot[black, thick, domain=0:0.5, samples=200]
        {exp(-0.25*(2*pi*x)^2)};
    \addlegendentry{Exact: $e^{-\frac{\gamma_2}{2}(2\pi\phi)^2}$}

    % === FTCS (orange, solid) ===
    % 1 + gamma2*(cos(2*pi*phi) - 1) with gamma2=0.5
    \addplot[orange!70!black, thick, domain=0:0.5, samples=200]
        {1 + 0.5*(cos(deg(2*pi*x)) - 1)};
    \addlegendentry{FTCS (explicit)}

    % === BTCS (orange, dashed) ===
    % 1/(1 - gamma2*(cos(2*pi*phi) - 1)) with gamma2=0.5
    \addplot[orange!70!black, thick, dashed, domain=0:0.5, samples=200,
        name path=btcs]
        {1/(1 - 0.5*(cos(deg(2*pi*x)) - 1))};
    \addlegendentry{BTCS (implicit)}

    % === Learned emulator trained on BTCS data (blue, thick) ===
    % Ansatz: 1 + theta*(cos(2*pi*phi) - 1)
    % theta = gamma2 / (2*gamma2*sin^2(pi*psi) + 1)
    % With gamma2=0.5, psi=0.15:
    %   theta = 0.5 / (2*0.5*sin(pi*0.15)^2 + 1) = 0.5 / (1 + sin(27deg)^2)
    %         = 0.5 / (1 + 0.2061) = 0.5 / 1.2061 ≈ 0.4145
    \addplot[TUMBlue, thick, domain=0:0.5, samples=200,
        name path=learned]
        {1 + (0.5/(2*0.5*sin(deg(pi*0.15))^2 + 1))*(cos(deg(2*pi*x)) - 1)};
    \addlegendentry{Learned (trained on BTCS at $\psi{=}0.15$)}

    % === Superiority shading between BTCS and Learned ===
    \addplot[green!15, forget plot]
        fill between[of=btcs and learned, soft clip={domain=0.15:0.5}];
    \node[font=\scriptsize, text=green!40!black] at (axis cs:0.38,0.72)
        {\emph{superiority}};

    % === Training mode marker ===
    \draw[TUMGrayLight, densely dashed, thin]
        (axis cs:0.15,-0.15) -- (axis cs:0.15,1.15);
    \node[font=\tiny, text=TUMGrayLight, anchor=south west]
        at (axis cs:0.155,1.0) {$\psi$};

    \end{axis}
    \end{tikzpicture}
    \caption{%
        Mode-wise multipliers for the one-dimensional diffusion equation with
        $\gamma_2 = 0.5$ (within the stability domain of the explicit scheme).
        The exact solution (black) decays exponentially with wavenumber. The
        explicit FTCS scheme (orange, solid) over-attenuates high frequencies,
        while the implicit BTCS scheme (orange, dashed) under-attenuates them. A learned
        emulator with FTCS-form inductive bias, trained on BTCS data at mode
        $\psi = 0.15$ (blue), achieves a multiplier that lies closer to the
        exact solution than its training source across a range of higher
        modes.}
    \label{fig:multiplier-curves}
\end{figure}
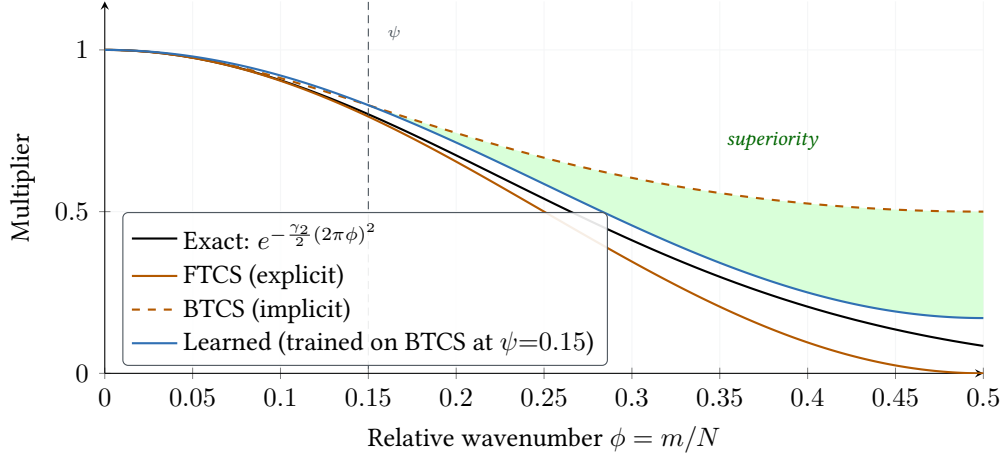

For simple linear PDEs, we can perform the same mode-wise analysis as in
\cref{sec:foundations:spectral}, but now for learned emulators. Consider an
emulator with the functional form of the FTCS scheme in
\cref{eq:diffusion-multiplier-ftcs} but with a free parameter $\theta$ replacing
$\gamma_2$,
\begin{equation}\label{eq:learned-ansatz-diffusion}
    \hat{q}_\phi(\theta) = 1 + \theta \left(\cos(2\pi\phi) - 1\right).
\end{equation}
This ansatz has the same inductive bias as the FTCS scheme. However, the
parameter $\theta$ is not derived from first principles about the PDE and its
constitutive parameter. Instead, it is fit from data by minimizing the squared
error at a training mode $\psi$\footnote{Crucially, for this proof we assume
that we only fit the ansatz at mode $M$ given relatively as $\psi = M/N$. As
shown in the full paper of \cref{paper:superiority}, this allows for a
closed-form optimizer $\theta^*$. In the paper, we extend the discussion to
richer states.}. Depending on which solver generates the training data, the
optimal $\theta^*$ differs.

If the training data comes from the FTCS scheme, the optimal fit recovers
$\theta = \gamma_2$ exactly. The ansatz reproduces its training source. If the
training data comes from the implicit BTCS of
\cref{eq:diffusion-multiplier-btcs} scheme, fitting at mode $\psi$ yields
\begin{equation}\label{eq:learned-theta-btcs}
    \theta^* = \frac{\gamma_2}{2\gamma_2 \sin^2(\pi\psi) + 1}.
\end{equation}
This is a ``corrected'' $\gamma_2$: it converges to $\gamma_2$ as $\psi \to 0$
but deviates for higher training modes.
Crucially, if we plug this $\theta^*$ back into the ansatz (see the paper in
\cref{paper:superiority} for the symbolic expression), this learned multiplier
can be \emph{closer} to the analytical multiplier
\eqref{eq:diffusion-multiplier-exact} than the BTCS training data was. As such,
the emulator has exceeded its training source in certain frequency bands. In
\cref{fig:multiplier-curves}, we plot a representative evaluation of all four
involved functions. Beyond the training mode $\psi$, the distance from the
learned multiplier to the analytical one is smaller than the distance from the
BTCS multiplier to it.
See the superiority paper for more interesting discussions of what
happens if we combinatorially change ansatz, training reference, baseline, and
even the PDE.

This analysis, while restricted to linear PDEs and linear ansätze, reveals an
interesting mechanism. The inductive bias of the ansatz (here, the FTCS
functional form) acts as implicit regularization, steering the learned
multiplier toward the analytical solution even when trained on low-fidelity
data.

\subsection{From State-Space to Autoregressive Superiority}
\label{sec:relationship:superiority:types}

To formalize what we have just observed, we define the \emph{superiority ratio}.
Given the rollout test metric of \cref{eq:rollout_test_metric}, we compare the
emulator's error against a high-fidelity reference $\tilde{\mathcal{P}}_h$
(role (2) in \cref{fig:solver-roles}) to the error of a \emph{baseline} solver
$\mathcal{P}_h^{\text{base}}$ (role (5) in \cref{fig:solver-roles}) against the
same reference:
\begin{equation}\label{eq:superiority-ratio}
    \xi^{[n]} = \frac{
        \mathbb{E}\left[
            \tilde{\zeta}\!\left(
                f_\theta^n\left(\mathbf{u}_h^{[0]}\right),\;
                \tilde{\mathcal{P}}_h^n\left(\mathbf{u}_h^{[0]}\right)
            \right)
        \right]
    }{
        \mathbb{E}\left[
            \tilde{\zeta}\!\left(
                \mathcal{P}_h^{\text{base},n}\left(\mathbf{u}_h^{[0]}\right),\;
                \tilde{\mathcal{P}}_h^n\left(\mathbf{u}_h^{[0]}\right)
            \right)
        \right]
    }.
\end{equation}
When $\xi^{[n]} < 1$, the emulator is closer to the high-fidelity reference
than the baseline solver, i.e., it has achieved \emph{superiority}. In the most
interesting case, the baseline is the training solver itself
($\mathcal{P}_h^{\text{base}} = \mathcal{P}_h$, i.e., roles (1) and (5) in
\cref{fig:solver-roles} coincide), meaning the emulator exceeds the very data
it was trained on.

The spectral analysis of \cref{sec:relationship:superiority:spectral} is an
example of \emph{state-space superiority} ($\xi^{[1]} < 1$): the emulator was
trained at mode $\psi$ but tested on other modes, so the improvement comes from
generalizing across the state space. Superiority is not automatic; it requires a
\emph{mismatch between the emulator's inductive biases and the solver's error
structure} (when the emulator has the same functional form as the training
solver, $\theta^* = \gamma_2$ exactly and no superiority occurs), a
\emph{mismatch between training and testing conditions} (here through different
spectral content), and a \emph{sufficiently inaccurate training solver} (if the
training data is already close to the analytical solution, there is little room
for improvement).

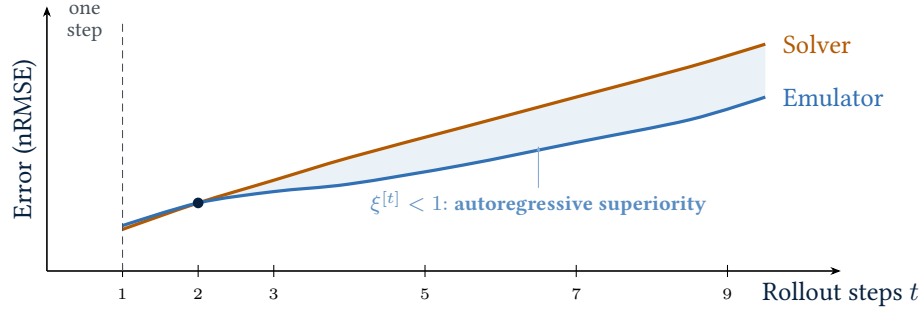
\begin{figure}
    \centering
    \begin{tikzpicture}[
        annot/.style={font=\scriptsize},
    ]

    % === Axes ===
    \draw[-{Stealth[length=4pt]}, semithick]
        (0, 0) -- (10.5, 0)
        node[below, font=\small, text=TUMBlueDark] {Rollout steps $t$};
    \draw[-{Stealth[length=4pt]}, semithick]
        (0, 0) -- (0, 3.5)
        node[at={(0, 1.75)}, font=\small, text=TUMBlueDark, rotate=90,
             anchor=south]
        {Error (nRMSE)};

    % Tick marks on x-axis
    \foreach \t/\lab in {1/1, 2/2, 3/3, 5/5, 7/7, 9/9} {
        \draw (\t, -0.08) -- (\t, 0.08);
        \node[font=\tiny, below] at (\t, -0.1) {$\lab$};
    }

    % === Shaded superiority region ===
    \fill[TUMBlue!10]
        plot[smooth, tension=0.55] coordinates {
            (2.0, 0.9) (3.0, 1.2) (4.0, 1.5) (5.5, 1.9)
            (7.0, 2.3) (8.5, 2.7) (9.5, 3.0)
        }
        -- plot[smooth, tension=0.55] coordinates {
            (9.5, 2.3) (8.5, 2.0) (7.0, 1.7) (5.5, 1.4)
            (4.0, 1.15) (3.0, 1.05) (2.0, 0.9)
        }
        -- cycle;

    % === Solver error curve (orange) ===
    \draw[orange!70!black, very thick]
        plot[smooth, tension=0.55] coordinates {
            (1.0, 0.55)
            (2.0, 0.9)
            (3.0, 1.2)
            (4.0, 1.5)
            (5.5, 1.9)
            (7.0, 2.3)
            (8.5, 2.7)
            (9.5, 3.0)
        };
    \node[font=\small, text=orange!70!black, anchor=west]
        at (9.6, 3.0) {Solver};

    % === Emulator error curve (blue) ===
    \draw[TUMBlue, very thick]
        plot[smooth, tension=0.55] coordinates {
            (1.0, 0.6)
            (2.0, 0.9)
            (3.0, 1.05)
            (4.0, 1.15)
            (5.5, 1.4)
            (7.0, 1.7)
            (8.5, 2.0)
            (9.5, 2.3)
        };
    \node[font=\small, text=TUMBlue, anchor=west]
        at (9.6, 2.3) {Emulator};

    % === Crossing point ===
    \fill[TUMBlueDark] (2.0, 0.9) circle (2pt);

    % === Vertical separator at t=1 ===
    \draw[TUMGrayLight, densely dashed, thin]
        (1.0, 0) -- (1.0, 3.3);

    % === Annotations ===
    \node[annot, text=TUMGrayLight, align=center, anchor=south]
        at (0.5, 2.9) {one\\[-1pt]step};

    % Superiority region label
    \node[annot, text=TUMBlue!80, align=center]
        at (6.5, 0.9)
        {$\xi^{[t]} < 1$: \textbf{autoregressive superiority}};
    \draw[-{Stealth[length=2pt]}, TUMBlue!60, thin]
        (6.5, 1.1) -- (6.5, 1.65);

    \end{tikzpicture}
    \caption{%
        Conceptual sketch of the emulator superiority phenomenon over
        autoregressive rollout steps. The solver (orange) and emulator (blue)
        both accumulate errors over time, but the emulator's more favorable
        error growth causes it to cross below the solver's error curve.
        In the shaded region, the superiority ratio $\xi^{[t]} < 1$,
        meaning the emulator is closer to the high-fidelity reference than
        the solver it was trained on. This \emph{autoregressive superiority}
        emerges even when the emulator is slightly less accurate at $t = 1$.
    }
    \label{fig:superiority-sketch}
\end{figure}

However, there is a second form that does not require a change in initial
condition distribution. \emph{Autoregressive superiority} ($\xi^{[1]} \geq 1$
but $\xi^{[n]} < 1$ for $n \geq 2$) occurs even when training and test
distributions are identical. The emulator accumulates errors more favorably than
the numerical solver over multiple rollout steps
(\cref{fig:superiority-sketch}).

In our experiments (see \cref{paper:superiority}), autoregressive superiority
appeared across all tested architectures (ConvNets, ResNets, FNOs, UNets,
Transformers). State-space superiority, on the other hand, was observed
primarily for convolutional architectures with local receptive fields.
Global-receptive-field architectures (FNOs, Transformers) did not achieve it
under the same distribution shift in our setup.
These findings extend to nonlinear PDEs; for the full empirical analysis, see
the publication in \cref{paper:superiority}.

Superiority is a consequence of the interplay between solver imperfections and
emulator structure. Since virtually all numerical solvers are imperfect (often
intentionally so, to remain computationally feasible), the phenomenon is more
broadly relevant than it might first appear. Whether superiority manifests in a
given scenario depends on the precise experimental configuration, underscoring
the importance of understanding solver roles and careful benchmarking practices
(\cref{sec:relationship:benchmarking}).

\section{Efficient Differentiable Physics Training via Progressive Refinement}
\label{sec:relationship:fidelity}

The previous section examined how solver properties affect emulators through
training data (role~(1) in \cref{fig:solver-roles}) as well as through
architectural analogy (role~(4)). A third interface is the coarse solver
$\mathcal{P}_H$ as a differentiable component for a neural-hybrid model
(role~(3)), see \cref{sec:emulators:hybrid}. Depending on the configuration,
gradients must flow through $\mathcal{P}_H$ either when training with the
one-step supervised objective of \cref{eq:one_step_supervised_objective} or
under unrolled training. Oftentimes, $\mathcal{P}_H$ contains an iterative
solver, e.g., for implicit time stepping
(\cref{sec:foundations:numerical:temporal:implicit}) or for constraint
enforcement (like the incompressibility projection
\citep{chorin1968numerical}). This creates a natural bilevel optimization
structure: the outer problem learns the network parameters $\theta$, while the
inner problem is the iterative solver reaching convergence. The key insight,
developed as part of \citet{bhatia2024unrolling_thesis} and, subsequently, the
PRDP paper (\cref{paper:prdp}), is that \emph{full solver convergence is
unnecessary for full network accuracy}. This enables progressive refinement
strategies that begin with coarse physics and adaptively increase fidelity
during training.

Note that this is distinct from manipulating the fidelity of the
\emph{data-generating} solver~(1), which we discuss as an open research
direction in \cref{sec:outlook:future:generation}.

\subsection{The Bilevel Optimization Perspective}
\label{sec:relationship:fidelity:bilevel}

The bilevel view connects differentiable physics training to a well-studied
problem in optimization. \citet{pedregosa2016hyperparameter} showed that when
the outer problem (here: learning $\theta$) is far from convergence, the inner
problem (here: the iterative solver) does not need to be solved to full
accuracy. Differentiable optimization layers \citet{amos2017optnet} formalize
this further: any rootfinding or linear solve within the forward pass can be
viewed as an inner optimization problem whose convergence tolerance is a free
parameter. In our setting, the inner problem is the iterative physics solver
with $K$ iterations, and the question is how $K$ relates to the quality of
the gradients $\nabla_\theta L$ that the outer optimizer receives.

\subsection{Progressive Refinement and Incomplete Convergence}
\label{sec:relationship:fidelity:prdp}

The gradient of the outer loss with respect to the network parameters depends on
the accuracy of the inner solver. A more converged forward solve yields a more
accurate solution, and a more converged backward solve (whether via unrolled or
implicit differentiation, see \cref{sec:foundations:iterative:differentiation})
yields a more accurate Jacobian. In principle, both must be fully converged for
exact gradients. In practice, two observations about neural network training
relax this requirement considerably.

\begin{description}
    \item[Early gradients can be coarse.] \label{sec:relationship:fidelity:prdp:coarse}
    At the beginning of training, the network parameters are far from any optimum
    and the loss landscape is navigated in broad strokes. Precise gradients are
    unnecessary when the optimizer is making large, exploratory steps. This mirrors
    the classical insight from bilevel optimization
    \citep{pedregosa2016hyperparameter}: when the outer problem is far from
    convergence, the inner problem does not need to be solved to full accuracy. In
    the differentiable physics setting, this means that starting with few solver
    iterations $K \ll K_\epsilon$ and progressively increasing $K$ as training
    matures can achieve the same final accuracy at lower total cost. We call the
    resulting savings \emph{progressive refinement} (PR) savings.

    \item[Full convergence is never needed.] \label{sec:relationship:fidelity:prdp:incomplete}
    Even late in training, neural networks do not require exact gradients. Several
    factors contribute: (i) stochastic minibatching introduces gradient noise that
    typically dominates solver-induced inaccuracy, (ii) momentum-based optimizers
    like Adam~\citep{kingma2014adam} normalize gradient magnitudes, reducing the
    importance of exact gradient norms, and (iii) neural networks are inherently
    approximative models whose final accuracy (on the order of $\text{nRMSE} \sim
    10^{-2}$) is far coarser than the tolerance $\epsilon$ required for solver
    convergence. Empirically, there exists a threshold $K_{\max} \ll K_\epsilon$
    beyond which further solver refinement yields no measurable improvement in
    network accuracy. Training at $K_{\max}$ throughout, rather than $K_\epsilon$,
    constitutes \emph{incomplete convergence} (IC) savings.
\end{description}

Taken together, PR and IC savings compound: starting coarse saves iterations
early, and stopping refinement at $K_{\max}$ saves iterations for the entire
remainder of training. We illustrate this in \cref{fig:prdp-schematic}. The
challenge is that neither the optimal refinement schedule nor $K_{\max}$ is
known a priori; both depend on the PDE, its discretization, the iterative
solver, and the learning dynamics. This motivates an adaptive algorithm.

\subsection{The PRDP Algorithm and Computational Savings}
\label{sec:relationship:fidelity:algorithm}

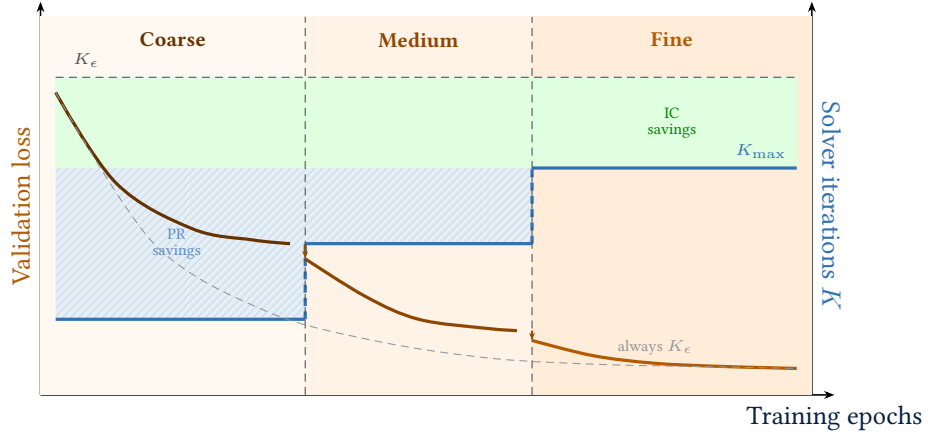
\begin{figure}
    \centering
    \begin{tikzpicture}[
        annot/.style={font=\scriptsize, text=TUMGrayLight},
    ]

    % === Coordinate helpers (equal-step staircase) ===
    \pgfmathsetmacro{\yKone}{1.0}     % step 1 (coarse)
    \pgfmathsetmacro{\yKtwo}{2.0}     % step 2 (medium)
    \pgfmathsetmacro{\yKmax}{3.0}     % step 3 = K_max (fine)
    \pgfmathsetmacro{\yKeps}{4.2}     % K_epsilon (full convergence)

    % === Left y-axis (validation loss) ===
    \draw[-{Stealth[length=4pt]}, semithick]
        (0, 0) -- (10.5, 0)
        node[below, font=\small, text=TUMBlueDark] {Training epochs};
    \draw[-{Stealth[length=4pt]}, semithick]
        (0, 0) -- (0, 5.2)
        node[at={(0, 2.5)}, font=\small, text=orange!70!black, rotate=90,
             anchor=south]
        {Validation loss};

    % === Right y-axis (solver iterations K, schematic) ===
    \draw[-{Stealth[length=4pt]}, semithick]
        (10.2, 0) -- (10.2, 5.2)
        node[at={(10.2, 2.5)}, font=\small, text=TUMBlue, rotate=-90,
             anchor=south]
        {Solver iterations $K$};

    % === Background stage shading ===
    \fill[orange!5] (0, 0) rectangle (3.5, 5.0);
    \fill[orange!10] (3.5, 0) rectangle (6.5, 5.0);
    \fill[orange!15] (6.5, 0) rectangle (10.2, 5.0);

    % Stage labels at top
    \node[font=\scriptsize\bfseries, text=orange!40!black]
        at (1.75, 4.7) {Coarse};
    \node[font=\scriptsize\bfseries, text=orange!55!black]
        at (5.0, 4.7) {Medium};
    \node[font=\scriptsize\bfseries, text=orange!70!black]
        at (8.35, 4.7) {Fine};

    % === IC savings region (horizontal band across ALL stages) ===
    \fill[green!12]
        (0.2, \yKmax) rectangle (10.0, \yKeps);

    % === PR savings region (staircase below K_max, hatched) ===
    \fill[TUMBlue!12]
        (0.2, \yKone) rectangle (3.5, \yKmax)
        (3.5, \yKtwo) rectangle (6.5, \yKmax);
    \fill[pattern=north east lines, pattern color=TUMBlue!25]
        (0.2, \yKone) rectangle (3.5, \yKmax)
        (3.5, \yKtwo) rectangle (6.5, \yKmax);

    % === K_epsilon reference line (full convergence) ===
    \draw[TUMGrayLight, densely dashed, thin]
        (0.2, \yKeps) -- (10.0, \yKeps);
    \node[font=\tiny, text=TUMGrayLight, anchor=south west]
        at (0.3, \yKeps) {$K_\epsilon$};

    % === PRDP staircase (solver iterations schedule) ===
    \draw[TUMBlue, very thick]
        (0.2, \yKone) -- (3.5, \yKone)
        (3.5, \yKtwo) -- (6.5, \yKtwo)
        (6.5, \yKmax) -- (10.0, \yKmax);

    % Vertical jumps at refinement points
    \draw[TUMBlue, very thick, densely dashed]
        (3.5, \yKone) -- (3.5, \yKtwo)
        (6.5, \yKtwo) -- (6.5, \yKmax);

    % K_max label
    \node[font=\tiny, text=TUMBlue, anchor=south east]
        at (10.0, \yKmax) {$K_{\max}$};

    % === Savings labels ===
    \node[font=\tiny, text=TUMBlue!80, align=center]
        at (1.8, {0.5*(\yKone + \yKmax)}) {PR\\[-1pt]savings};
    \node[font=\tiny, text=green!50!black, align=center]
        at (8.35, {0.5*(\yKmax + \yKeps)}) {IC\\[-1pt]savings};

    % === Validation loss curves (left y-axis) ===
    % Stage 1: rapid decrease then plateau
    \draw[orange!40!black, very thick]
        plot[smooth, tension=0.6] coordinates {
            (0.2, 4.0)
            (1.0, 2.8)
            (2.0, 2.2)
            (2.8, 2.05)
            (3.3, 2.0)
        };

    % Stage 2: drop at refinement, decrease, plateau
    \draw[orange!55!black, very thick]
        plot[smooth, tension=0.6] coordinates {
            (3.5, 1.8)
            (4.3, 1.3)
            (5.0, 1.0)
            (5.9, 0.88)
            (6.3, 0.85)
        };

    % Stage 3: drop at refinement, converge
    \draw[orange!70!black, very thick]
        plot[smooth, tension=0.6] coordinates {
            (6.5, 0.72)
            (7.3, 0.52)
            (8.1, 0.42)
            (8.9, 0.38)
            (9.6, 0.36)
            (10.0, 0.35)
        };

    % === ``Always fine'' reference curve (gray, thin) ===
    \draw[TUMGrayLight!60, thin, densely dashed]
        plot[smooth, tension=0.55] coordinates {
            (0.2, 4.0)
            (1.5, 2.0)
            (3.0, 1.1)
            (4.5, 0.7)
            (6.0, 0.48)
            (7.5, 0.40)
            (9.0, 0.37)
            (10.0, 0.35)
        };
    \node[font=\tiny, text=TUMGrayLight!60, anchor=south west]
        at (7.5, 0.40) {always $K_\epsilon$};

    % === Refinement trigger lines ===
    \draw[TUMGrayLight, densely dashed, thin]
        (3.5, 0) -- (3.5, 5.0);
    \draw[TUMGrayLight, densely dashed, thin]
        (6.5, 0) -- (6.5, 5.0);

    % Refine arrows on validation loss
    \draw[-{Stealth[length=3pt]}, orange!60!black, thick]
        (3.5, 2.0) -- (3.5, 1.8);
    \draw[-{Stealth[length=3pt]}, orange!60!black, thick]
        (6.5, 0.85) -- (6.5, 0.72);

    % === Monitoring ratio annotation ===
    % \node[annot, anchor=north west] at (0.2, -0.15)
    %     {plateau $\to$ refine: $r > \tau_{\text{step}}$};

    \end{tikzpicture}
    \caption{%
        Schematic of the PRDP progressive refinement strategy. \emph{Left
        axis (orange):} Training begins with a coarse physics solver and
        monitors the validation loss. When the loss plateaus, solver fidelity
        is increased. The dashed gray curve shows training with fully
        converged physics ($K_\epsilon$) throughout; both strategies converge
        to the same final accuracy. \emph{Right axis (blue):} The PRDP
        staircase schedule of solver iterations $K$. Two types of savings
        emerge: \emph{PR savings} (blue hatched) from using fewer solver
        iterations during the early, coarse stages of training, and
        \emph{IC savings} (green band spanning all stages) from the network
        achieving full accuracy at $K_{\max}$ well below the convergence
        threshold $K_\epsilon$.
    }
    \label{fig:prdp-schematic}
\end{figure}

The PRDP algorithm automates both progressive refinement and the detection of
$K_{\max}$ by monitoring a validation metric during training. Starting from a
coarse solver with $K_0$ iterations (typically $K_0 \approx 1$), the algorithm
periodically evaluates a validation loss and detects when it plateaus at the
current refinement level. A plateau signals that the solver fidelity, not the
network capacity, is the bottleneck, prompting an increase of $K$ by $\Delta K$
iterations. If a refinement step does not yield meaningful improvement over the
previous plateau, the algorithm concludes that $K_{\max}$ has been reached and
stops refining. This combination yields both PR savings (from starting coarse)
and IC savings (from stopping early), as illustrated in
\cref{fig:prdp-schematic}.

Across our experiments, PRDP achieves total savings of up to 86\% of cumulative solver
iterations compared to training with fully converged physics. These savings
translate directly into wall-clock time reductions, particularly for
higher-dimensional problems where iterative linear solves dominate the
computational cost.
The IC savings also carry over to inference: if the learned model is deployed
with the differentiable physics component at $K_{\max}$ rather than
$K_\epsilon$, the inference cost is reduced proportionally. For the full
algorithm in pseudocode and detailed experimental results, see
the full publication in \cref{paper:prdp}.

\section{Implications for Benchmarking}
\label{sec:relationship:benchmarking}

The insights from the preceding sections have direct consequences for how neural
PDE emulators should be evaluated. The superiority phenomenon
(\cref{sec:relationship:superiority}) showed that emulator accuracy is
entangled with the fidelity of the training solver. The spectral error analysis
(\cref{sec:foundations:spectral}) revealed that errors are scale-dependent and
that global metrics can mask important structure. And the bilevel perspective on
differentiable physics (\cref{sec:relationship:fidelity:bilevel}) demonstrated
that even the gradients used for training need not come from fully converged
solvers. Together, these observations motivate a set of desiderata for benchmarking
suites that go beyond simply providing fixed datasets and leaderboards. In the
language of \cref{fig:solver-roles}, a benchmark must be explicit about which
solver fills which role.

\begin{description}
    \item[High-Fidelity Reference Solutions.]
    \label{sec:relationship:benchmarking:reference}
        The superiority phenomenon makes one thing clear: when an emulator is
        evaluated against its own training solver, the result conflates the
        emulator's error with the solver's error. A benchmark that uses a
        low-fidelity solver to generate both training data and test references
        cannot distinguish a good emulator from one that has merely learned to
        reproduce numerical artifacts. This also raises a deeper question:
        should the emulator match $\mathcal{P}_h$ (as it is trained to) or
        resolve the underlying PDE $\mathbb{P}$ more accurately? The
        superiority results suggest the latter can happen, making high-fidelity
        references essential for detecting it.

    \item[Flexible Data Generators over Fixed Datasets.]
    \label{sec:relationship:benchmarking:generators}
        A benchmark can provide fixed datasets for the community to download and
        use. However, the particular relationship between simulators and
        emulators makes this insufficient. Unlike domains where training data
        comes from measurements with inherent noise and collection cost, PDE
        emulators are trained on synthetic data that can be generated on demand,
        cheaply (at least for canonical PDEs) and without measurement error. A
        benchmark should leverage this by emphasizing the \emph{data generator}
        rather than a fixed dataset.

    \item[Rollout Metrics over One-Step Accuracy.]
    \label{sec:relationship:benchmarking:rollout}
        Neural emulators are deployed autoregressively: the
        output of one step becomes the input to the next. Errors therefore
        compound over the rollout, and the \emph{rate} at which they grow
        matters more than their magnitude at any single step. An emulator with
        slightly larger one-step error but more favorable error accumulation can
        outperform a one-step-optimal model over longer horizons. In fact, the
        superiority phenomenon (\cref{sec:relationship:superiority})
        demonstrates that rollout behavior can even reverse the ranking between
        the emulator and its training solver. One-step metrics would entirely
        miss this.

        For chaotic systems, the connection between rollout errors and dynamical
        systems theory is even more direct. Nearby trajectories diverge at a
        rate governed by the system's largest Lyapunov exponent
        $\lambda$
        (cf.\ \cref{sec:foundations:implementation:chaotax}),
        providing a physics-informed lower bound on error growth: no emulator,
        however accurate, can beat $e^{[n]} \sim e^{\lambda_{\max} n \Delta t}$ in the
        pointwise sense once trajectories have decorrelated. This means that
        rollout metrics for chaotic dynamics should be interpreted relative to
        the Lyapunov time scale $1/\lambda$, and that comparisons at a
        fixed number of steps (e.g., $n = 50$) are physically meaningless
        without accounting for the underlying rate of chaos.

    \item[Precise Specification of the Emulated System.]
    \label{sec:relationship:benchmarking:normalized}
        A numerical simulator $\mathcal{P}_h$ is uniquely identified by a large
        number of parameters: the differential operators, their constitutive
        coefficients (e.g., diffusivity $\nu$, advection velocity $c$), the
        boundary conditions, the spatial resolution $N$, the time step $\Delta
        t$, the domain extent $L$, and the dimensionality $D$. It must be clear
        what constitutes the PDE system $\mathbb{P}$ and what constitutes the
        discretization choices. Changing a single physical parameter like $\nu$
        while holding $\Delta t$ and $L$ fixed simultaneously alters both the
        physical dynamics and the effective numerical difficulty, confounding any
        attempt to isolate the effect of one factor on emulation quality.

    \item[Unambiguous Metric Definitions.]
    \label{sec:relationship:benchmarking:reporting}
        The performance of an emulator depends not only on what we measure
        against but also on \emph{how} we measure. The spectral error analysis
        of \cref{sec:foundations:spectral} showed that errors are distributed
        unevenly across spatial scales. Global scalar metrics compress this
        structure into a single number. Even a seemingly standard term like
        ``nRMSE'' is ambiguous without specifying the order in which axes are
        reduced and whether normalization is applied per channel or globally.

    \item[Multi-Seed Statistical Rigor.]
    \label{sec:relationship:benchmarking:seeds}
        Neural network training is inherently stochastic: results depend on
        parameter initialization, data sampling order, and (for procedural data)
        the initial condition seeds. Single-seed comparisons between
        architectures or training strategies are unreliable, as the variance
        across seeds can exceed the differences between methods. A benchmark
        should run each configuration across multiple seeds and report
        distributional summaries (medians, interquartile ranges), enabling
        proper hypothesis testing to determine whether observed differences are
        statistically significant.

    \item[Systematic Training Methodology Coverage.]
    \label{sec:relationship:benchmarking:methodology}
        Many studies conflate architecture improvements with training
        improvements, making it unclear which factor drives a reported gain. A
        benchmark should allow testing architectures under the same training
        setup and vice versa. This requires a systematic taxonomy of training
        configurations.

    \item[Computational Efficiency Reporting.]
        \label{sec:relationship:benchmarking:efficiency} Accuracy alone is
        insufficient for evaluating emulators. Since the motivation for
        surrogates is often computational speedup, a benchmark should report
        accuracy relative to cost (wall-clock time, FLOPs, or parameter count)
        and compare the emulator against the solver it is meant to replace (role
        (5) in \cref{fig:solver-roles}). A recent meta-analysis by
        \citet{mcgreivy2024weak} documents that much of the ML-for-PDE
        literature reports speedups against weak or poorly tuned baselines, and
        that these speedups often shrink or vanish once role (5) is filled by a
        properly optimized classical solver. This reporting bias is precisely
        what the five-role decomposition is designed to surface: when roles (1),
        (2), and (5) collapse onto a single under-specified solver, there is no
        structural way to audit whether a claimed win reflects emulator quality
        or baseline weakness.

\end{description}

\chapter{Systematic Emulator Benchmarking}
\label{ch:apebench}

\begin{figure}[ht]
    \centering
    \includegraphics[width=\textwidth]{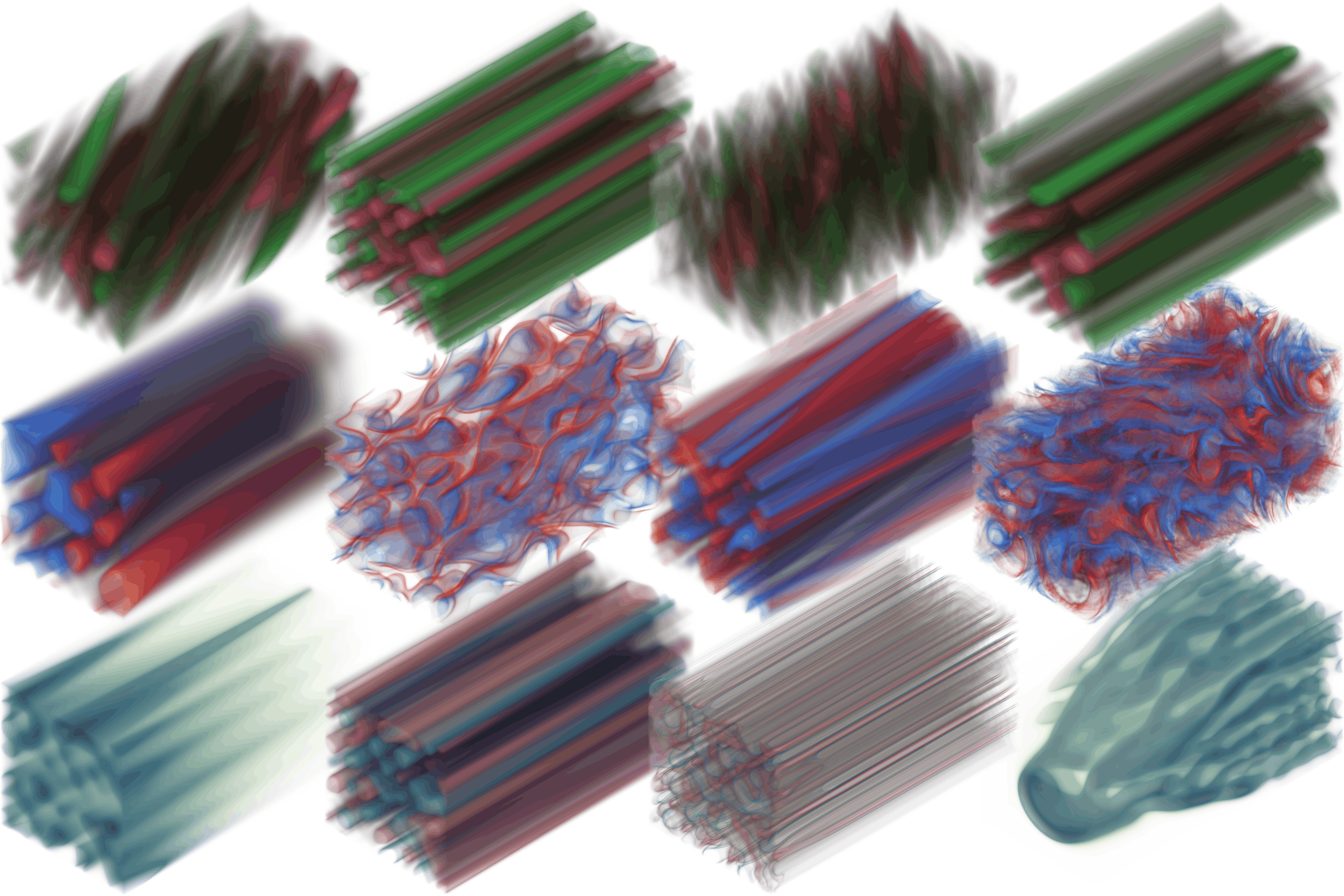}
    \caption{Qualitative overview of a subset of the PDE dynamics supported by
    APEBench, here in 2D displayed as volume-rendered spatiotemporal plots with
    time on the diagonal axis. Top row: linear dynamics (advection, diffusion,
    dispersion, hyper-diffusion). Middle row: nonlinear dynamics (Burgers,
    Kuramoto-Sivashinsky, Decaying Navier-Stokes, Kolmogorov Flow). Bottom row:
    reaction-diffusion dynamics (Fisher-KPP, Allen-Cahn, Gray-Scott,
    Swift-Hohenberg).}
    \label{fig:apebench-configurations}
\end{figure}

\section{The APEBench Framework}
\label{sec:apebench:framework}

\subsection{Design Principles}
\label{sec:apebench:framework:principles}

The desiderata of \cref{sec:relationship:benchmarking} call for a benchmark that
treats the solver as a first-class component rather than a black-box data source.
APEBench \citep{koehler2024apebench} is built around four design principles
that realize this vision (the full paper is in \cref{paper:apebench}).

\begin{description}
    \item[Procedural data generation.] \label{sec:apebench:framework:principles:procedural}
    Rather than distributing fixed datasets, APEBench generates all training and
    evaluation trajectories procedurally from the Exponax solver
    (\cref{sec:foundations:implementation:exponax}). A single trajectory in 1D is
    produced in milliseconds on a modern accelerator, making datasets of
    arbitrary size feasible without storage overhead. More importantly, procedural
    generation ensures that every experiment is exactly reproducible from a random
    seed and a set of PDE parameters, eliminating the ambiguity that arises when
    different studies subsample or preprocess the same downloaded dataset
    differently.

    \item[Solver as user-facing interface.] \label{sec:apebench:framework:principles:solver}
    APEBench exposes the numerical solver directly: users can modify PDE
    coefficients, spatial resolution $N$, time step $\Delta t$, and the
    initial-condition distribution rather than consuming opaque data files. This
    enables the controlled variation experiments to further investigate the 
    emulator-solver relationship discussed in \cref{ch:relationship}.

    \item[Differentiable physics.]
    \label{sec:apebench:framework:principles:diffphysics} Because the Exponax
    solver is implemented in JAX, it is automatically differentiable and can
    participate directly in the training loop. This enables differentiable
    physics losses and neural-hybrid correction tasks where the emulator is
    composed with the solver (see \cref{sec:emulators:hybrid}).

    \item[Statistical evaluation via seed parallelism.]
    \label{sec:apebench:framework:principles:seeds} Neural emulator performance
    is sensitive to the random seed, both through the initial-condition sampling
    and the network initialization. APEBench therefore runs each configuration
    across multiple seeds (by default 50 in 1D, 20 in 2D and 3D) and reports
    medians with interquartile ranges. This is made practical by JAX's
    \texttt{vmap} transformation, which vectorizes over seeds within a single
    program, saturating the GPU that a single-seed run would underutilize. The
    result is that a 50-seed sweep in 1D costs roughly the same wall-clock time
    as a single run.
\end{description}

\subsection{46 PDE Configurations with Normalized Identifiers}
\label{sec:apebench:framework:configurations}

APEBench provides 46 PDE configurations organized into three categories, probing
a wide range of physics phenomena while clearly communicating how the PDE system
is parameterized and solved numerically.

\begin{description}
    \item[Linear dynamics] \label{sec:apebench:framework:configurations:linear}
    cover advection, diffusion, dispersion, and hyper-diffusion, along with
    spatially mixed and anisotropic variants that only exist in two or more
    dimensions. Because linear dynamics on periodic boundaries preserve
    bandlimitedness, the ETDRK integrator solves them without spatial
    discretization error, which is useful for cleanly isolating emulator errors.

    \item[Nonlinear dynamics]
    \label{sec:apebench:framework:configurations:nonlinear} include the viscous
    Burgers equation, the Korteweg--de Vries (KdV) equation, the
    Kuramoto--Sivashinsky (KS) equation, and the incompressible Navier--Stokes
    equations. These problems can generate richer states during rollout.

    \item[Reaction-diffusion dynamics]
    \label{sec:apebench:framework:configurations:reaction} comprise the
    Fisher--KPP equation, the Gray--Scott model, and the Swift--Hohenberg
    equation. Unlike the fluid-like problems, their nonlinearities are spatially
    local polynomials. They tend to develop rich high-frequency patterns that,
    in most cases, converge to non-trivial steady states.
\end{description}

For a qualitative overview, see \cref{fig:apebench-configurations}. The absence
of certain problems like high-Reynolds-number fluid flows, the Euler equations,
is an inherent limitation of the pseudo-spectral
approach. Yet, APEBench exercises the core competencies (Laplacian, transport,
incompressibility, polynomial nonlinearities, dispersion, chaos) that underpin
most of continuum modelling. Since we can probe behavior in isolation and in
combination, the insights obtained likely translate to more sophisticated
setups.

We additionally classify each configuration by qualitative properties:
whether it is decaying, infinitely running, chaotic, multi-channel, or
converging to a patterned steady state. The full table with continuous forms,
classifications, and dimension availability is in the Appendix of the APEBench
paper in \cref{paper:apebench}.

The initial conditions (as per \cref{sec:emulators:training:data}) are
sampled procedurally. Different initial condition generators allow for various
spectral behaviors, including flat (i.e., white noise), polynomial (i.e.,
Gaussian random fields), and exponential power spectra. These can additionally
be clamped to certain frequency ranges. The choice of generator matters because
the initial spectrum shapes what the emulator sees during training. For linear
PDEs on periodic domains, the active modes never interact, so the initialization
distribution determines the long-term spectral content entirely. The emulator's
accuracy on modes not represented in the training distribution is then a form of
state-space generalization (\cref{sec:emulators:training:generalization}).

Most scenarios can be addressed through three interface modes that put the
normalized identifiers of \cref{sec:relationship:benchmarking:normalized} into
practice: \emph{difficulty mode} (DIFF) via $\gamma_j$, \emph{normalized mode}
(NORM) via $\alpha_j$, or \emph{physical mode} (PHYS) via the raw constitutive
parameters together with $N$, $\Delta t$, and $L$. Setting any one determines
the others automatically. These modes serve as an exchange protocol: specifying
$\gamma_j$ at a given $N$ is sufficient for exact reproducibility and meaningful
cross-study comparison\footnote{For further details, see the APEBench
documentation at \url{https://tum-pbs.github.io/apebench/}.}.

\subsection{The Training Taxonomy}
\label{sec:apebench:framework:taxonomy}

\begin{figure}
    \centering
    \begin{tikzpicture}[
        % Mini-schematic node styles
        ic/.style={circle, fill=TUMGray, draw=TUMGrayLight!60,
                   minimum size=2.5mm, inner sep=0pt},
        emu/.style={circle, fill=TUMBlue!30, draw=TUMBlue!70,
                    minimum size=2.5mm, inner sep=0pt},
        sol/.style={circle, fill=orange!30, draw=orange!70,
                    minimum size=2.5mm, inner sep=0pt},
        earrow/.style={-{Stealth[length=2pt]}, TUMBlue!60, very thin},
        sarrow/.style={-{Stealth[length=2pt]}, orange!50!black, very thin},
        loss/.style={densely dashed, TUMGrayLight, very thin},
        cellname/.style={font=\tiny\bfseries},
    ]

    % Grid parameters
    \def\cw{3.2}   % cell width
    \def\ch{2.2}   % cell height

    % === Axis labels ===
    \node[font=\small\bfseries, text=TUMBlueDark, rotate=90, anchor=south]
        at (-2.7, -\ch) {Rollout steps $T$};
    \node[font=\small\bfseries, text=TUMBlueDark]
        at (\cw, 1.8) {Branch length $B$};

    % Column headers
    \node[font=\small, text=TUMBlueDark] at (0, 1.4) {$B{=}1$};
    \node[font=\small, text=TUMBlueDark] at (\cw, 1.4) {$B{=}2$};
    \node[font=\small, text=TUMBlueDark] at (2*\cw, 1.4) {$B{=}3$};

    % Row headers
    \node[font=\small, text=TUMBlueDark] at (-2.0, 0) {$T{=}1$};
    \node[font=\small, text=TUMBlueDark] at (-2.0, -\ch) {$T{=}2$};
    \node[font=\small, text=TUMBlueDark] at (-2.0, -2*\ch) {$T{=}3$};

    % === Invalid cells (B > T): gray background ===
    % (T=1, B=2)
    \fill[TUMGray!20, rounded corners=2pt]
        (\cw-\cw/2+0.08, -\ch/2+0.08) rectangle (\cw+\cw/2-0.08, \ch/2-0.08);
    \node[font=\scriptsize, text=TUMGrayLight!50] at (\cw, 0) {$B > T$};
    % (T=1, B=3)
    \fill[TUMGray!20, rounded corners=2pt]
        (2*\cw-\cw/2+0.08, -\ch/2+0.08) rectangle (2*\cw+\cw/2-0.08, \ch/2-0.08);
    \node[font=\scriptsize, text=TUMGrayLight!50] at (2*\cw, 0) {$B > T$};
    % (T=2, B=3)
    \fill[TUMGray!20, rounded corners=2pt]
        (2*\cw-\cw/2+0.08, -\ch-\ch/2+0.08) rectangle (2*\cw+\cw/2-0.08, -\ch+\ch/2-0.08);
    \node[font=\scriptsize, text=TUMGrayLight!50] at (2*\cw, -\ch) {$B > T$};

    % =====================================================================
    % CELL (T=1, B=1): One-step supervised
    % =====================================================================
    \begin{scope}[shift={(0, 0)}]
        \fill[TUMBlue!6, rounded corners=2pt]
            (-\cw/2+0.08, -\ch/2+0.08) rectangle (\cw/2-0.08, \ch/2-0.08);
        \node[cellname, text=TUMBlue!80!black] at (0, 0.7) {One-step supervised};
        % Emulator: IC -> e1
        \node[ic] (a-ic1) at (-0.25, 0.2) {};
        \node[emu] (a-e1) at (0.25, 0.2) {};
        \draw[earrow] (a-ic1) -- (a-e1);
        % Solver: IC -> s1
        \node[ic] (a-ic2) at (-0.25, -0.35) {};
        \node[sol] (a-s1) at (0.25, -0.35) {};
        \draw[sarrow] (a-ic2) -- (a-s1);
        % Loss
        \draw[loss] (a-e1) -- (a-s1);
    \end{scope}

    % =====================================================================
    % CELL (T=2, B=1): Diverted chain
    % =====================================================================
    \begin{scope}[shift={(0, -\ch)}]
        \fill[purple!6, rounded corners=2pt]
            (-\cw/2+0.08, -\ch/2+0.08) rectangle (\cw/2-0.08, \ch/2-0.08);
        \node[cellname, text=purple!70!black] at (0, 0.7) {Diverted chain};
        % Emulator: IC -> e1 -> e2
        \node[ic] (b-ic) at (-0.5, 0.25) {};
        \node[emu] (b-e1) at (0, 0.25) {};
        \node[emu] (b-e2) at (0.5, 0.25) {};
        \draw[earrow] (b-ic) -- (b-e1);
        \draw[earrow] (b-e1) -- (b-e2);
        % Solver branches (1-step each, from previous state)
        \node[sol] (b-s1) at (0, -0.3) {};
        \node[sol] (b-s2) at (0.5, -0.3) {};
        \draw[sarrow] (b-ic) -- (b-s1);
        \draw[sarrow] (b-e1) -- (b-s2);
        % Losses
        \draw[loss] (b-e1) -- (b-s1);
        \draw[loss] (b-e2) -- (b-s2);
    \end{scope}

    % =====================================================================
    % CELL (T=3, B=1): Diverted chain
    % =====================================================================
    \begin{scope}[shift={(0, -2*\ch)}]
        \fill[purple!6, rounded corners=2pt]
            (-\cw/2+0.08, -\ch/2+0.08) rectangle (\cw/2-0.08, \ch/2-0.08);
        \node[cellname, text=purple!70!black] at (0, 0.7) {Diverted chain};
        % Emulator: IC -> e1 -> e2 -> e3
        \node[ic] (c-ic) at (-0.75, 0.25) {};
        \node[emu] (c-e1) at (-0.25, 0.25) {};
        \node[emu] (c-e2) at (0.25, 0.25) {};
        \node[emu] (c-e3) at (0.75, 0.25) {};
        \draw[earrow] (c-ic) -- (c-e1);
        \draw[earrow] (c-e1) -- (c-e2);
        \draw[earrow] (c-e2) -- (c-e3);
        % Solver branches (1-step each)
        \node[sol] (c-s1) at (-0.25, -0.3) {};
        \node[sol] (c-s2) at (0.25, -0.3) {};
        \node[sol] (c-s3) at (0.75, -0.3) {};
        \draw[sarrow] (c-ic) -- (c-s1);
        \draw[sarrow] (c-e1) -- (c-s2);
        \draw[sarrow] (c-e2) -- (c-s3);
        % Losses
        \draw[loss] (c-e1) -- (c-s1);
        \draw[loss] (c-e2) -- (c-s2);
        \draw[loss] (c-e3) -- (c-s3);
    \end{scope}

    % =====================================================================
    % CELL (T=2, B=2): Supervised unrolling
    % =====================================================================
    \begin{scope}[shift={(\cw, -\ch)}]
        \fill[orange!6, rounded corners=2pt]
            (-\cw/2+0.08, -\ch/2+0.08) rectangle (\cw/2-0.08, \ch/2-0.08);
        \node[cellname, text=orange!60!black] at (0, 0.7) {Supervised};
        % Emulator: IC -> e1 -> e2
        \node[ic] (d-ic1) at (-0.5, 0.25) {};
        \node[emu] (d-e1) at (0, 0.25) {};
        \node[emu] (d-e2) at (0.5, 0.25) {};
        \draw[earrow] (d-ic1) -- (d-e1);
        \draw[earrow] (d-e1) -- (d-e2);
        % Solver: IC -> s1 -> s2 (parallel, precomputed)
        \node[ic] (d-ic2) at (-0.5, -0.3) {};
        \node[sol] (d-s1) at (0, -0.3) {};
        \node[sol] (d-s2) at (0.5, -0.3) {};
        \draw[sarrow] (d-ic2) -- (d-s1);
        \draw[sarrow] (d-s1) -- (d-s2);
        % Losses
        \draw[loss] (d-e1) -- (d-s1);
        \draw[loss] (d-e2) -- (d-s2);
    \end{scope}

    % =====================================================================
    % CELL (T=3, B=2): General configuration
    % =====================================================================
    \begin{scope}[shift={(\cw, -2*\ch)}]
        \draw[TUMGrayLight!30, rounded corners=2pt]
            (-\cw/2+0.08, -\ch/2+0.08) rectangle (\cw/2-0.08, \ch/2-0.08);
        % Emulator: IC -> e1 -> e2 -> e3
        \node[ic] (e-ic) at (-0.75, 0.35) {};
        \node[emu] (e-e1) at (-0.25, 0.35) {};
        \node[emu] (e-e2) at (0.25, 0.35) {};
        \node[emu] (e-e3) at (0.75, 0.35) {};
        \draw[earrow] (e-ic) -- (e-e1);
        \draw[earrow] (e-e1) -- (e-e2);
        \draw[earrow] (e-e2) -- (e-e3);
        % Branch 1: from IC, 2 solver steps, targets e2
        \node[sol] (e-sa) at (-0.25, -0.1) {};
        \node[sol] (e-sb) at (0.25, -0.1) {};
        \draw[sarrow] (e-ic) -- (e-sa);
        \draw[sarrow] (e-sa) -- (e-sb);
        \draw[loss] (e-e2) -- (e-sb);
        % Branch 2: from e1, 2 solver steps, targets e3
        \node[sol] (e-sc) at (0.25, -0.55) {};
        \node[sol] (e-sd) at (0.75, -0.55) {};
        \draw[sarrow] (e-e1) -- (e-sc);
        \draw[sarrow] (e-sc) -- (e-sd);
        \draw[loss] (e-e3) -- (e-sd);
    \end{scope}

    % =====================================================================
    % CELL (T=3, B=3): Supervised unrolling
    % =====================================================================
    \begin{scope}[shift={(2*\cw, -2*\ch)}]
        \fill[orange!6, rounded corners=2pt]
            (-\cw/2+0.08, -\ch/2+0.08) rectangle (\cw/2-0.08, \ch/2-0.08);
        \node[cellname, text=orange!60!black] at (0, 0.7) {Supervised};
        % Emulator: IC -> e1 -> e2 -> e3
        \node[ic] (f-ic1) at (-0.75, 0.25) {};
        \node[emu] (f-e1) at (-0.25, 0.25) {};
        \node[emu] (f-e2) at (0.25, 0.25) {};
        \node[emu] (f-e3) at (0.75, 0.25) {};
        \draw[earrow] (f-ic1) -- (f-e1);
        \draw[earrow] (f-e1) -- (f-e2);
        \draw[earrow] (f-e2) -- (f-e3);
        % Solver: IC -> s1 -> s2 -> s3 (parallel, precomputed)
        \node[ic] (f-ic2) at (-0.75, -0.3) {};
        \node[sol] (f-s1) at (-0.25, -0.3) {};
        \node[sol] (f-s2) at (0.25, -0.3) {};
        \node[sol] (f-s3) at (0.75, -0.3) {};
        \draw[sarrow] (f-ic2) -- (f-s1);
        \draw[sarrow] (f-s1) -- (f-s2);
        \draw[sarrow] (f-s2) -- (f-s3);
        % Losses
        \draw[loss] (f-e1) -- (f-s1);
        \draw[loss] (f-e2) -- (f-s2);
        \draw[loss] (f-e3) -- (f-s3);
    \end{scope}

    % === Ellipsis indicators ===
    \node[font=\large, text=TUMGrayLight!60] at ({2*\cw+\cw/2+0.5}, -\ch) {$\cdots$};
    \node[font=\large, text=TUMGrayLight!60] at (\cw, {-2*\ch-\ch/2-0.3}) {$\vdots$};

    % === Legend ===
    \begin{scope}[shift={(0, {-2*\ch-\ch/2-0.8})}]
        \node[ic, label={[font=\tiny, xshift=1mm]right:Initial cond.}] at (-0.8, 0) {};
        \node[emu, label={[font=\tiny, xshift=1mm]right:Emulator}] at (0.9, 0) {};
        \node[sol, label={[font=\tiny, xshift=1mm]right:Solver}] at (2.4, 0) {};
        \draw[loss] (3.6, 0) -- ++(0.35, 0)
            node[right, font=\tiny, text=TUMGrayLight] {Loss};
    \end{scope}

    \end{tikzpicture}
    \caption{%
        The $(T, B)$ training taxonomy for autoregressive neural PDE emulators.
        $T$ denotes the number of autoregressive rollout steps, $B$ the branch
        length of the reference solver.
    }
    \label{fig:training-taxonomy}
\end{figure}
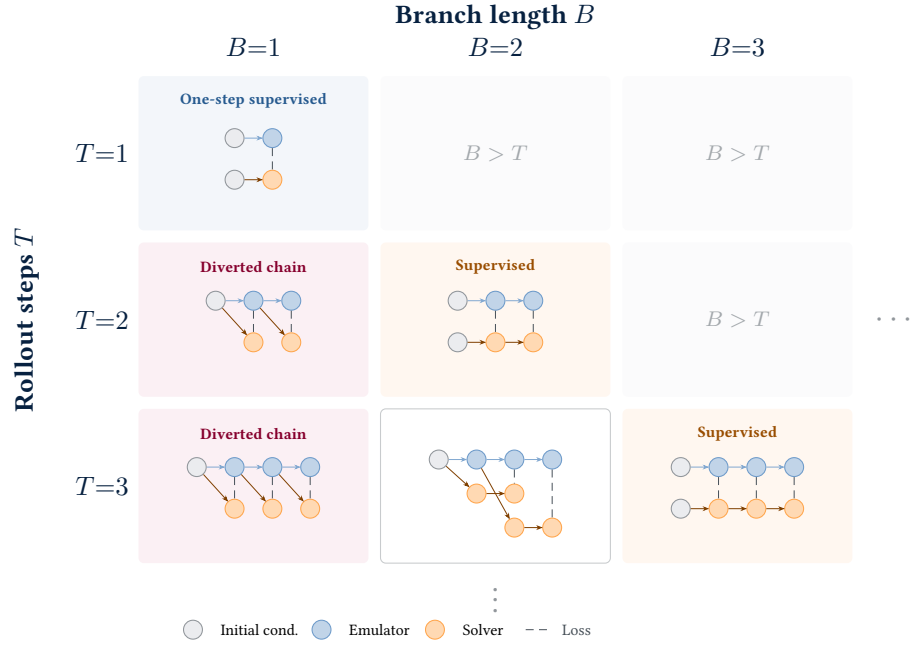

To systematically investigate the effect of unrolling and differentiable physics
during training, we introduced a unified objective parameterized by two integers:
the number of autoregressive rollout steps $T$ and the reference branch length
$B$ (with $B \leq T$). Given states $\mathbf{u}_h \sim \mathcal{D}_h$ sampled
from the training data, the objective reads
\begin{equation}\label{eq:objective-taxonomy}
    L(\theta) = \mathbb{E}_{\mathbf{u}_h \sim \mathcal{D}_h} \left[
        \sum_{n=0}^{T-B} \sum_{b=1}^{B}
        \zeta \left(
            f_\theta^{n+b}(\mathbf{u}_h), \;
            \mathcal{P}_h^{b}(f_\theta^{n}(\mathbf{u}_h))
        \right)
    \right],
\end{equation}
where $f_\theta^k$ denotes applying the emulator $k$ times autoregressively
(with $f_\theta^0$ being the identity) and $\zeta$ is a per-step loss, typically
the mean squared error. The emulator produces a \emph{main chain} of length $T$,
and at each position $n$ along this chain, a \emph{branch} of $B$ reference
solver steps provides the training target. \Cref{fig:training-taxonomy}
illustrates the resulting grid of configurations.

The key practical distinction is between $B = T$ and $B < T$. When $B = T$,
all reference targets can be precomputed from the initial condition, so the
solver need not be differentiable (or even present at training time). When $B <
T$, the solver sits inside the computational graph at each branch point and must
support gradient propagation. This is where APEBench's differentiable Exponax
solver becomes essential: the diverted chain configuration, in particular, is not
supported by any other PDE benchmark.

The taxonomy extends to neural-hybrid emulators. When $f_\theta$ is replaced by
a correction architecture $f_\theta = c_\theta \circ
\mathcal{P}_H$ that composes a corrector network with a coarse solver,
any unrolling with $T \geq 2$ requires the coarse solver $\mathcal{P}_H$
to also be differentiable since it is part of the main chain. APEBench supports
this by providing the coarse solver from the same ETDRK suite.

\subsection{Metrics and Reduction Orders}
\label{sec:apebench:framework:metrics}

\begin{figure}
    \centering
    \begin{minipage}[t]{0.60\textwidth}
        \centering
        \textbf{(a)} Three nRMSE reduction orders\par\medskip
        \begin{tikzpicture}[
            box/.style={rectangle, rounded corners=2pt, minimum width=2.6cm,
                minimum height=0.55cm, font=\scriptsize, align=center,
                inner sep=3pt},
            inputbox/.style={box, draw=TUMGrayLight, fill=TUMGray},
            linbox/.style={box, draw=TUMGrayLight!60, fill=white},
            nlbox/.style={box, draw=red!50!black, fill=red!8},
            chosenbox/.style={box, draw=green!50!black, fill=green!5,
                thick},
            outbox/.style={box, draw=TUMBlue!60, fill=TUMBlue!8,
                minimum width=1.6cm},
            arr/.style={-{Stealth[length=3pt]}, TUMGrayLight!70, thin},
            collabel/.style={font=\scriptsize\bfseries, text=TUMBlueDark},
        ]

        % === Column positions ===
        \def\colA{-3.3}
        \def\colB{0}
        \def\colC{3.3}

        % === Column headers ===
        \node[collabel] at (\colA, 0.3) {Per-channel, per-step};
        \node[font=\tiny, text=green!50!black] at (\colA, -0.05)
            {(APEBench)};
        \node[collabel] at (\colB, 0.3) {Global, per-step};
        \node[font=\tiny, text=TUMGrayLight] at (\colB, -0.05)
            {(e.g., PDEArena)};
        \node[collabel] at (\colC, 0.3) {Over trajectory};
        \node[font=\tiny, text=TUMGrayLight] at (\colC, -0.05)
            {(common in ML)};

        % === Row 1: Input ===
        \node[inputbox] (a1) at (\colA, -0.7)
            {$e^2, r^2$\; {\tiny$(T, C, X)$}};
        \node[inputbox] (b1) at (\colB, -0.7)
            {$e^2, r^2$\; {\tiny$(T, C, X)$}};
        \node[inputbox] (c1) at (\colC, -0.7)
            {$e^2, r^2$\; {\tiny$(T, C, X)$}};

        % === Row 2: First reduction (linear) ===
        \node[linbox] (a2) at (\colA, -1.65)
            {$\operatorname{mean}_X$\; {\tiny$(T, C)$}};
        \node[linbox] (b2) at (\colB, -1.65)
            {$\operatorname{mean}_{C,X}$\; {\tiny$(T,)$}};
        \node[linbox] (c2) at (\colC, -1.65)
            {$\operatorname{mean}_{T,C,X}$\; {\tiny scalar}};

        \draw[arr] (a1) -- (a2);
        \draw[arr] (b1) -- (b2);
        \draw[arr] (c1) -- (c2);

        % === Row 3: Nonlinear normalization ===
        \node[chosenbox] (a3) at (\colA, -2.6)
            {$\div\, r_{t,c},\;\sqrt{\cdot}$\; {\tiny$(T, C)$}};
        \node[nlbox] (b3) at (\colB, -2.6)
            {$\div\, r_t,\;\sqrt{\cdot}$\; {\tiny$(T,)$}};
        \node[nlbox] (c3) at (\colC, -2.6)
            {$\div\, r,\;\sqrt{\cdot}$\; {\tiny scalar}};

        \draw[arr] (a2) -- (a3);
        \draw[arr] (b2) -- (b3);
        \draw[arr] (c2) -- (c3);

        % === Row 4: Final reduction / output ===
        \node[outbox] (a4) at (\colA, -3.55)
            {$\sum_C$ $\to$ {\tiny\bfseries$(T,)$}};
        \node[outbox] (b4) at (\colB, -3.55)
            {\tiny\bfseries $(T,)$};
        \node[linbox, minimum width=1.6cm] (c4) at (\colC, -3.55)
            {\tiny\bfseries scalar};

        \draw[arr] (a3) -- (a4);
        \draw[arr] (b3) -- (b4);
        \draw[arr] (c3) -- (c4);

        % === Annotations below columns ===
        \node[font=\tiny, text=green!40!black, align=center]
            at (\colA, -4.15) {rollout metric\\[-1pt]weak channels visible};
        \node[font=\tiny, text=TUMGrayLight, align=center]
            at (\colB, -4.15) {rollout metric\\[-1pt]weak channels diluted};
        \node[font=\tiny, text=red!40!black, align=center]
            at (\colC, -4.15) {no temporal\\[-1pt]structure};

        \end{tikzpicture}
    \end{minipage}%
    \hfill
    \begin{minipage}[t]{0.37\textwidth}
        \centering
        \textbf{(b)} Temporal aggregation\par\medskip
        \begin{tikzpicture}[
            box/.style={rectangle, rounded corners=2pt,
                minimum height=0.55cm, font=\scriptsize, align=center,
                inner sep=3pt},
            inputbox/.style={box, draw=TUMBlue!60, fill=TUMBlue!8,
                minimum width=2.6cm},
            linbox/.style={box, draw=TUMGrayLight!60, fill=white,
                minimum width=1.6cm},
            chosenbox/.style={box, draw=green!50!black, fill=green!5,
                thick, minimum width=1.6cm},
            arr/.style={-{Stealth[length=3pt]}, TUMGrayLight!70, thin},
        ]

        % === Input (matches blue output from panel a) ===
        \node[inputbox] (tin) at (0, -0.7)
            {$\text{nRMSE}^{[t]}$\; {\tiny$(T,)$}};

        % === Branch ===
        \node[linbox] (arith) at (-1.3, -1.7)
            {$\frac{1}{T}\sum_t$};
        \node[chosenbox] (geom) at (1.3, -1.7)
            {$\bigl(\prod_t\bigr)^{1/T}$};

        \draw[arr] (tin.south) -- ++(0, -0.25) -| (arith.north);
        \draw[arr] (tin.south) -- ++(0, -0.25) -| (geom.north);

        % === Output ===
        \node[linbox, minimum width=1.2cm] (out1) at (-1.3, -2.6)
            {\tiny scalar};
        \node[chosenbox, minimum width=1.2cm] (out2) at (1.3, -2.6)
            {\tiny scalar};

        \draw[arr] (arith) -- (out1);
        \draw[arr] (geom) -- (out2);

        % === Labels ===
        \node[font=\tiny, text=TUMGrayLight, align=center]
            at (-1.3, -3.05) {arithmetic};
        \node[font=\tiny, text=green!40!black, align=center]
            at (1.3, -3.2) {geometric\\[-1pt](APEBench)};

        % === Motivation sketch: centered between the branches ===
        \begin{scope}[shift={(0, -5.0)}]
            % Mini axes
            \draw[-{Stealth[length=2pt]}, thin] (-1.3, 0) -- (1.6, 0)
                node[right, font=\tiny] {$t$};
            \draw[-{Stealth[length=2pt]}, thin] (-1.3, 0) -- (-1.3, 1.1);

            % Diverging nRMSE curve (exponential blowup)
            \draw[TUMGrayLight, thick]
                plot[smooth, tension=0.5] coordinates {
                    (-1.1, 0.04) (-0.7, 0.07) (-0.2, 0.12) (0.1, 0.18)
                    (0.4, 0.3) (0.7, 0.48) (0.9, 0.7) (1.1, 0.95)
                };
            % Divergence arrow
            \draw[TUMGrayLight, thick, -{Stealth[length=2pt]}]
                plot[smooth, tension=0.6] coordinates {
                    (1.1, 0.95) (1.25, 1.05)
                };

            % Arithmetic mean line (pulled very high by divergence)
            \draw[red!50!black, densely dashed, thin]
                (-1.3, 0.7) -- (1.4, 0.7)
                node[right, font=\tiny, text=red!50!black] {arith.};

            % Geometric mean line (much lower, balanced)
            \draw[green!50!black, thick]
                (-1.3, 0.22) -- (1.4, 0.22)
                node[right, font=\tiny, text=green!50!black] {geom.};

            % Annotation
            \node[font=\tiny, text=TUMGrayLight, align=center]
                at (0, -0.2) {arithmetic dominated by divergent};
            \node[font=\tiny, text=TUMGrayLight, align=center]
                at (0, -0.45) {tail and sensitive to cutoff $T$};
        \end{scope}

        \end{tikzpicture}
    \end{minipage}
    \caption{%
        Metric reduction order for the nRMSE. \textbf{(a)}~Three reduction
        strategies for the prediction tensor $(T, C, X)$. APEBench uses
        per-channel normalization (green), preserving sensitivity to weak
        channel. \textbf{(b)}~Given a rollout metric $(T,)$ from~(a), temporal
        aggregation via arithmetic vs.\ geometric mean. The geometric mean
        (APEBench choice, green) avoids domination by late-time divergent errors
        and is less sensitive to the choice of rollout horizon~$T$ (sketch
        below). Batch averaging ($\operatorname{mean}_B$) is applied last in all
        cases and omitted for clarity.}
    \label{fig:metric-reduction}
\end{figure}
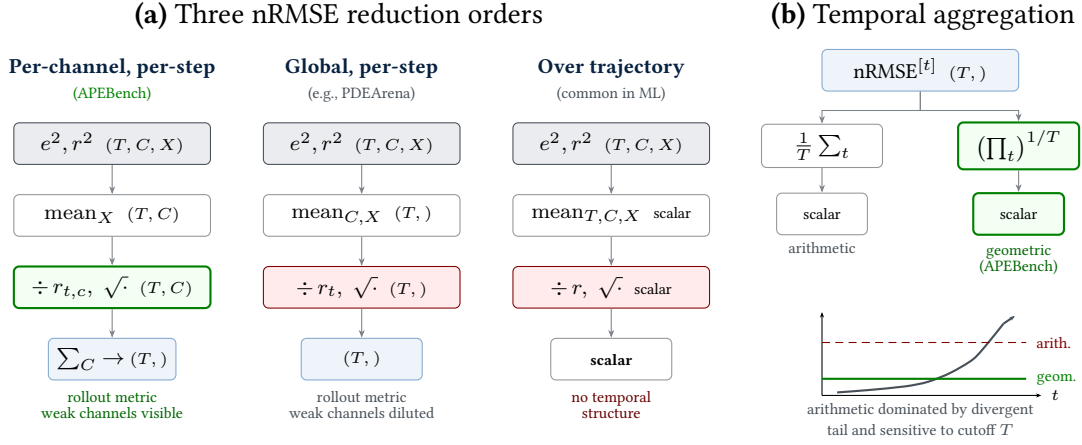

A metric for autoregressive PDE emulators must fully reduce the high-dimensional
prediction tensor $(B, T, C, X_1, \ldots, X_D)$ spanning batch, time, channels,
and spatial axes to a single scalar. Exponax
(\cref{sec:foundations:implementation:exponax}) implements a library of metric
functions that APEBench forwards to the user. These include metrics in state
space (MAE, MSE, RMSE and their normalized variants), in Fourier space (with
optional frequency band selection), Sobolev-type metrics that incorporate
derivative information, and Pearson correlations. The central subtlety is that
the \emph{order} in which axes are reduced matters whenever nonlinear operations
enter the pipeline.

If every operation in this reduction pipeline were linear, the
order of reduction would not matter: $\operatorname{mean}$ commutes with
$\operatorname{mean}$. The problem is that practical metric pipelines contain
nonlinear operations such as squaring in MSE, square roots in RMSE, division in
normalization, logarithms in geometric means. Nonlinear operations do not
commute with axis reduction. A metric is therefore fully specified only by its
\emph{reduction order}, i.e.,  the sequence of (axis, operation) pairs applied during
aggregation. Two metrics that reduce the same axes with the same operations but
in different orders are \emph{different metrics}, even if they share the same
name (see \cref{fig:metric-reduction}a).

% For further details on the metric framework, see the Exponax
% documentation\footnote{\url{https://fkoehler.site/exponax/examples/on_metrics_simple/}}
% and publication~\ref{paper:apebench}.

In practice, APEBench primarily reports \verb|mean_nRMSE|, which is the
per-channel normalized RMSE, aggregated over the batch via arithmetic mean.
Normalizing per channel ensures that weak channels are not masked by dominant
ones, and performing the batch average after normalization avoids conflating
sample-to-sample variation with scale differences. APEBench always returns the
full rollout error trajectory $e^{[n]}$. The geometric mean over time is only
used when a single scalar must be reported, since it is less sensitive to
late-time divergence than an arithmetic mean (\cref{fig:metric-reduction}b).
With multi-seed evaluation, the seed axis adds yet another dimension to the
reduction pipeline. APEBench always aggregates over seeds last, either per time
step (for rollout curves) or after temporal aggregation (for scalar summaries),
using the median with interquartile range as we observed this to be the most
robust for seed outliers.

\subsection{Software Ecosystem in JAX}
\label{sec:apebench:framework:software}

\begin{figure}
    \centering
    \begin{tikzpicture}[
        layer/.style={
            rectangle, rounded corners=3pt,
            minimum height=0.8cm, font=\small, align=center,
            inner sep=5pt,
        },
        jaxlayer/.style={layer, draw=TUMGrayLight, fill=TUMGray,
                         minimum width=13.5cm},
        extpkg/.style={layer, draw=TUMGrayLight!80, fill=TUMGray!30,
                       densely dashed, minimum width=5.0cm},
        solverpkg/.style={layer, draw=orange!60!black, fill=orange!10,
                          minimum width=3.2cm, minimum height=1.3cm,
                          font=\small\bfseries},
        emupkg/.style={layer, draw=TUMBlue!80, fill=TUMBlue!10,
                       minimum width=3.2cm, minimum height=1.3cm,
                       font=\small\bfseries},
        auxpkg/.style={layer, draw=orange!40!black, fill=orange!5,
                       densely dashed, minimum width=2.2cm,
                       minimum height=1.3cm, font=\small\bfseries},
        benchpkg/.style={layer, draw=purple!50, fill=purple!8,
                         minimum width=13.5cm, font=\small\bfseries},
        layerarrow/.style={-{Stealth[length=3pt]}, TUMGrayLight!60, thin},
    ]

    % === Layer 0: JAX ===
    \node[jaxlayer, minimum height=1.1cm] (jax) at (0, 0)
        {\textbf{JAX} \;\scriptsize\citep{jax2018github}\\[-1pt]
         \texttt{\scriptsize jit}\;\;
         \texttt{\scriptsize vmap}\;\;
         \texttt{\scriptsize grad}\;\;
         \texttt{\scriptsize lax.scan}\;\;
         \texttt{\scriptsize lax.while\_loop}};

    % === Layer 1: Equinox (external) + Chaotax (auxiliary, JAX-only) ===
    \node[extpkg] (equinox) at (-0.3, 1.6)
        {Equinox \;\scriptsize\citep{kidger2021equinox}};
    \node[font=\tiny, text=TUMGrayLight, anchor=west]
        at ([xshift=4pt]equinox.east) {\emph{external}};

    \node[auxpkg] (chaotax) at (5.5, 1.6)
        {Chaotax\\[-2pt]{\scriptsize\mdseries Lyapunov analysis}};

    % Layer 0 -> Layer 1 arrows
    \draw[layerarrow] (jax.north -| equinox.south) -- (equinox.south);
    \draw[layerarrow] (jax.north -| chaotax.south) -- (chaotax.south);

    % === Layer 2: Author contributions (three pillars + hot-swappable solver) ===
    % Three pillars of APEBench
    \node[solverpkg] (exponax) at (-2.2, 3.2)
        {Exponax\\[-2pt]{\scriptsize\mdseries Differentiable spectral}\\[-2pt]
         {\scriptsize\mdseries PDE solvers}};
    \node[emupkg] (pdequinox) at (1.5, 3.2)
        {PDEquinox\\[-2pt]{\scriptsize\mdseries Neural architectures}\\[-2pt]
         {\scriptsize\mdseries for emulators}};
    \node[emupkg] (trainax) at (4.8, 3.2)
        {Trainax\\[-2pt]{\scriptsize\mdseries Autoregressive}\\[-2pt]
         {\scriptsize\mdseries training methods}};

    % Hot-swappable solver alternative
    \node[solverpkg, minimum width=2.6cm] (picardax) at (-5.5, 3.2)
        {Picardax\\[-2pt]{\scriptsize\mdseries Finite-difference}\\[-2pt]
         {\scriptsize\mdseries solvers}};

    % Equinox -> Layer 2 fan-out arrows (all four Equinox-based packages)
    \draw[layerarrow] ([xshift=-2.0cm]equinox.north) -- (picardax.south);
    \draw[layerarrow] ([xshift=-0.7cm]equinox.north) -- (exponax.south);
    \draw[layerarrow] ([xshift=0.6cm]equinox.north) -- (pdequinox.south);
    \draw[layerarrow] ([xshift=1.8cm]equinox.north) -- (trainax.south);

    % Hot-swappable annotation: Picardax <-> Exponax
    \draw[orange!60!black, densely dashed, thick,
          {Stealth[length=3pt]}-{Stealth[length=3pt]}]
        (picardax.east) -- (exponax.west)
        node[midway, above, font=\tiny, text=orange!50!black]
            {\emph{hot swappable}};

    % === Layer 3: APEBench ===
    \node[benchpkg] (apebench) at (0, 4.8)
        {APEBench \;\scriptsize (Benchmarking Framework)};

    % Three pillars -> APEBench arrows
    \foreach \pkg in {exponax, pdequinox, trainax} {
        \draw[layerarrow] (\pkg.north) -- (\pkg.north |- apebench.south);
    }

    % === Layer 4: SuperAPE (specialized higher-level abstraction) ===
    \node[layer, draw=purple!40, fill=purple!5, densely dashed,
          minimum width=4.5cm, minimum height=0.9cm,
          font=\small\bfseries] (superape) at (4.0, 6.1)
        {SuperAPE\\[-2pt]{\scriptsize\mdseries Fast 1D superiority studies}};
    \draw[layerarrow] (apebench.north -| superape.south) -- (superape.south);

    \end{tikzpicture}
    \caption{%
        The JAX software ecosystem developed as part of this thesis.
        Equinox (dashed, external) supplies the module system on which all
        packages except Chaotax build. The three pillars of APEBench are
        Exponax (spectral PDE solvers), PDEquinox (neural architectures),
        and Trainax (training methodologies). Picardax provides
        finite-difference solvers with the same \texttt{equinox.Module}
        interface and is hot-swappable with Exponax. This interchangeability
        underpins both the PRDP and the emulator superiority work.
        Chaotax operates on pure JAX callables (independent of Equinox) to
        quantify dynamical properties such as Lyapunov exponents.
        SuperAPE is a higher-level abstraction above APEBench that uses
        \texttt{jax.lax.scan} for the training loop, enabling fast 1D
        superiority studies across solver types.
    }
    \label{fig:software-ecosystem}
\end{figure}
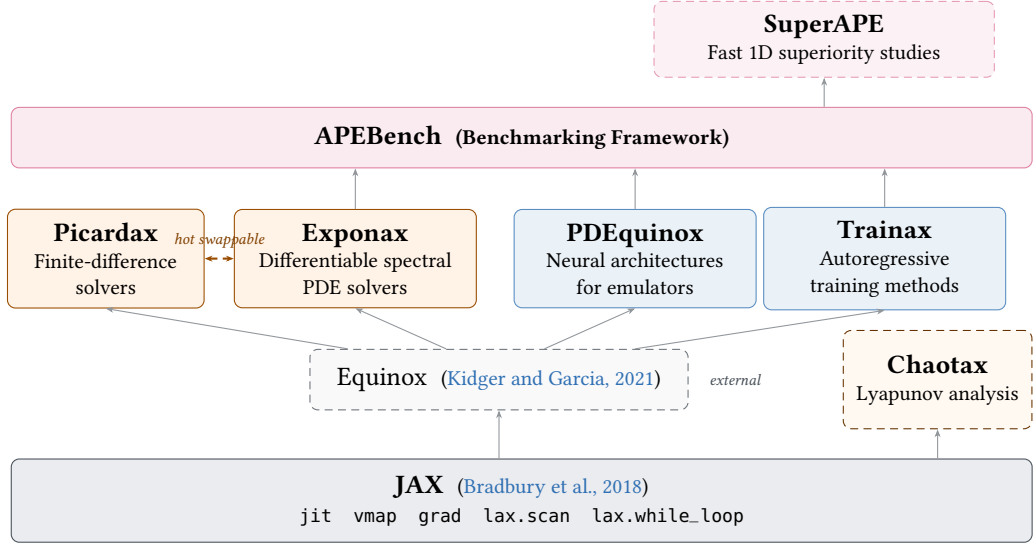

In contrast to PyTorch's monolithic approach \citep{paszke2019pytorch}, JAX
\citep{jax2018github} is designed as a lightweight numerical computing package.
As such, the deep learning stack is typically implemented using third-party
libraries, e.g., \verb|optax| \citep{deepmind2020jax} for gradient processing
(like the Adam optimizer \citep{kingma2014adam}). For general deep learning like
computer vision and natural language processing, Flax
\citep{flax2020github} has established itself as
the de facto standard for neural network building blocks. However, for
scientific applications, Equinox \citep{kidger2021equinox} is a popular
alternative. Its single-batch design is more natural when thinking of solving
ODEs and PDEs and helps with implementing numerical solvers. As such, it is the
basis for the solvers \verb|Exponax|
(\cref{sec:foundations:implementation:exponax}) and \verb|Picardax|
(\cref{sec:foundations:implementation:picardax}), discussed in \cref{ch:foundations}.
\Cref{fig:software-ecosystem} visualizes the layered ecosystem.

As part of this thesis, \verb|PDEquinox| was developed to provide a
collection of dimension- and boundary condition-agnostic architectures that can
be used to represent neural emulators $f_{\theta}$. Concretely, it includes
convolutional networks, residual networks, dilated residual networks, UNets, and
Fourier neural operators (\cref{sec:emulators:neural}). All are implemented as
\verb|equinox.Module| instances with a single-state call signature matching the
solver interface. The same architecture class works in 1D, 2D, and 3D by
adjusting the convolution dimension, and supports periodic, Dirichlet, and
Neumann boundaries via the padding mode. This uniformity enables the systematic
comparisons across 46 configurations
(\cref{sec:apebench:framework:configurations}) without per-scenario architecture
adjustments.

Moreover, we introduced \verb|Trainax| that abstracts the autoregressive
emulator learning pipeline. It implements the $(T, B)$ taxonomy of
\cref{sec:apebench:framework:taxonomy} as composable training configurations.
Each specifies how the emulator chain is unrolled, where the solver branches
attach, and how the loss is accumulated. This covers supervised, diverted chain,
and general configurations, as well as neural-hybrid setups where a coarse
solver is composed with a correction
network\footnote{An exhaustive visual catalog of training setups, including
gradient cut variants similar to \citep{list2025differentiability}, is available
at \url{https://fkoehler.site/predictor-learning-setups/}.}. By separating the
training logic from the architecture and the solver, switching between any $(T,
B)$ configuration requires no changes to the emulator or the scenario code.

In APEBench, these components are composed into \emph{scenario} objects. A
scenario bundles an \verb|Exponax| solver $\mathcal{P}_h$ (defining the PDE
system $\mathbb{P}$, the numerical method $\mathbb{M}$, and discretization
choices $N$ and $\Delta t$), a \verb|PDEquinox| architecture $f_\theta$
(defining the emulator), and a \verb|Trainax| configuration (defining the
training methodology via $T$ and $B$). Running an experiment reduces to
instantiating a scenario, calling its train and evaluate methods, and collecting
the metrics. Combined with \verb|jax.vmap| over random seeds, this is what makes
the one-line multi-seed experiments described in \cref{sec:apebench:framework:principles}
practical.

For the emulator superiority study (\cref{paper:superiority}), we developed
\verb|SuperAPE|, which extends APEBench by integrating the finite difference
solvers from \verb|Picardax|
(\cref{sec:foundations:implementation:picardax}) as an alternative solver
backend. This enables training on one solver type (e.g., finite differences) and
evaluating against another (e.g., spectral), which is precisely the setup needed
to study superiority across solver families. \verb|SuperAPE| uses
\verb|lax.scan| for the training loop, making fast 1D superiority sweeps
practical.

\section{Related Benchmarking Efforts}

APEBench builds upon previous efforts to provide synthetic
simulation data and evaluation resources. The datasets used in early works of
physics-based deep learning became popular references for subsequent works. This
includes the Burgers, Allen-Cahn, and Korteweg-de Vries datasets in 1D used
in \citet{raissi2019physics} and \citet{raissi2018hidden}. Moreover, the
datasets of the 2D Darcy Flow and 2D Kolmogorov Flow under low Reynolds numbers
of \citet{li2020fourier} became popular\footnote{It is important to note that
Burgers, Allen-Cahn, and Korteweg-de Vries datasets in these works were created
with \texttt{ChebFun} \citep{driscoll2014chebfun} which we consider a spiritual
predecessor for \texttt{Exponax} which is part of \texttt{APEBench}.}. The
unstructured datasets on flag motion, wake flow, plate deformation, and clothing
dynamics of \citet{pfaff2020learning} also remain popular as unstructured
benchmarks.

First explicit dataset and benchmarking approaches appeared for ordinary
differential equations and chaos with the work of \citet{gilpin2021chaos}. The
\verb|PDEBench| \citep{takamoto2022pdebench} became an instrumental resource for
most physics-based DL workflows, with \verb|PDEArena| \citep{gupta2022towards}
being released concurrently to it. Both have been extensively used in early
works of PDE foundation models \citep{mccabe2024multiple,hao2024dpot}. We view
\verb|APEBench| as complementary to these efforts, with a specific focus on
autoregressive generalization, fine-grained control via a fast differentiable
solver, and systematic training methodology coverage.

More specialized datasets and benchmark works appeared over the years. This
includes \citet{rasp2020weatherbench} and \citet{rasp2024weatherbench2} for
weather, \citet{hassan2023bubbleml} for boiling simulations, LagrangeBench
\citep{toshev2023lagrangebench} for SPH simulations, Constellaration
\citep{cadena2025constellaration} for stellarator designs, among others. While
orthogonal to autoregressive emulators, \verb|Pinnacle| \citep{hao2024pinnacle}
presented a systematic benchmarking approach for Physics-Informed Neural
Networks.

The field of computational fluid dynamics spawned the widest array of datasets
and benchmarking works, mostly targeting external flow aerodynamics. This
includes the early works of \citet{bonnet2022airfrans} and
\citet{luo2023cfdbench}. By now, the datasets have reached significant fidelity
and scale, starting with the works of \citet{elrefaie2024drivaernet++} and the
unprecedented \verb|DrivearNet| \citep{ashton2024drivaer}, as well as recently
the \verb|Shift-SUV| \citep{shift-suv} and \verb|Shift-Wing| \citep{shift-wing}
series.

For research in foundation models, large-scale datasets of problems of medium
difficulty have been produced and released. This includes \verb|The Well|
\citep{ohana2024well}. The pretraining corpus of Poseidon was released as
\verb|PDEGym| \citep{herde2024poseidon}. A popular data source is also the Johns
Hopkins turbulence dataset
\citep{li2008public,perlman2007data,kohl2026benchmarking}. The \verb|Exponax|
solver has also been used to scrape TeraBytes of simulation data in 2D
\citep{holzschuh2025pde} and in 3D \citep{holzschuh2026p3d}. Its PyTorch
translation \citep{liu_2025_15350210} is now being used to supply data online
(on-the-fly in a streaming fashion) to reduce memory burden in foundation model
training \citep{liu2026tadpole}.

\section{Systematic Investigation of the Emulation Pipeline}
\label{sec:apebench:studies}

The APEBench framework enables systematic investigation along multiple
experimental axes simultaneously.
We highlight key findings here. The full experiments, numeric tables, and
ablations are in the paper (\cref{paper:apebench}).

The entry point is the 1D advection experiment in which we scale the difficulty
$\gamma_1$ (i.e., the CFL number) while training different architectures.
Emulator accuracy degrades as $\gamma_1$ increases, and the degradation is
directly related to the mismatch between the architecture's receptive field and
the number of grid cells traversed per time step. Local architectures like
ConvNets and ResNets fail when this mismatch exceeds their receptive field,
while architectures with larger receptive fields (UNet, dilated ResNet) tolerate
higher difficulties. The FNO is essentially agnostic to $\gamma_1$ due to its
global receptive field in Fourier space, and the solution staying highly
bandlimited. This directly mirrors the CFL condition from classical numerics
(\cref{sec:foundations:numerical:temporal:explicit}): an emulator needs to
``see'' at least as many neighbors as the physics propagates per time step, just
as a finite difference stencil must cover the domain of dependence. In
borderline cases, unrolled training helped noticeably by extending the temporal
receptive field seen during optimization.

The broad architecture comparison across all 46 PDEs confirms that there is no
universally best architecture. ResNets offer the most consistent performance:
rarely the single best on any given problem, but almost never catastrophic. The
FNO excels on problems with global spatial structure, most notably the (low
Reynolds-number) Navier--Stokes Kolmogorov flow case, but struggles with
reaction-diffusion dynamics where energy concentrates in high-frequency patterns
that lie beyond its active modes. Local convolutional architectures perform best
on precisely those reaction-diffusion problems. In higher dimensions, the UNet
gains an advantage over the dilated ResNet, likely because its coarsening
hierarchy covers the receptive field more uniformly than axis-aligned dilations.

Training methodology provides a complementary lever. In the 2D advection
experiments, the diverted chain configuration ($B = 1$, $T > 1$) tended to
outperform supervised unrolling ($B = T$) for local architectures by combining
autoregressive exposure with one-step reference corrections.

The paper's appendix contains ablations over learning rate, training duration,
dataset size, and parameter count. A notable finding is that local
convolutional architectures are remarkably sample-efficient, converging with as
few as five training trajectories, while FNOs require more data. The temporal
horizon of the training data matters most for dynamics with multiple stages
(e.g., Burgers, KdV) where short horizons exclude important physical regimes,
but less so for dynamics already on their attractor (e.g.,
Kuramoto-Sivashinsky). The full experimental suite costs approximately
900~GPU-hours on 8$\times$~RTX 2080~Ti, enabled by procedural data generation
and seed parallelism via \texttt{jax.vmap}. Many additional axes remain to be
explored, including transfer learning, data augmentation, and scaling laws; we
discuss these in \cref{sec:outlook:limitations-and-questions}.

% =============================================================================
% CHAPTER 6: Discussion and Outlook
% =============================================================================
\chapter{Discussion and Outlook}
\label{ch:outlook}

\section{Cross-Cutting Findings}
\label{sec:outlook:findings}

Across all three publications, a recurring theme is that the surrogate pipeline
must be understood holistically. The numerical solver is not just a data source
but plays multiple roles (\cref{fig:solver-roles}), each with its own error
characteristics. Understanding these roles and their inherent numerical errors
is what enabled the superiority analysis, the PRDP savings, and the benchmarking
desiderata. Ultimately, we compare emulators against numerical methods, not just
in terms of accuracy but also in terms of speed. The infrastructure developed as
part of this thesis (Exponax, PDEquinox, Trainax, Picardax, Chaotax;
\cref{sec:foundations:implementation,sec:apebench:framework:software}) is what
made these systematic studies possible in the first place. Below, we distill the
cross-cutting insights that emerged from this perspective.

\subsection{The Spectral Lens as a Unifying Framework}

Fourier-spectral analysis is the methodological thread running through all three
publications. In the foundations, mode-wise multipliers characterize how
numerical schemes attenuate or amplify individual wavenumbers
(\cref{sec:foundations:spectral}). The same mode-wise perspective underpins the
ETDRK solvers in Exponax (\cref{sec:foundations:implementation:exponax}), which
treat the linear part of a PDE exactly in Fourier space, and it is precisely
what makes FNO layers interpretable as learned spectral time steppers
(\cref{sec:emulators:neural:fno}). In the superiority analysis, comparing
learned multipliers against analytical ones is what allowed us to prove in
closed form when and why an emulator exceeds its training source
(\cref{sec:relationship:superiority:spectral}). 

There is a broader lesson here. Spectral methods are rarely the method of choice
in practical engineering simulations, where complex geometries and non-periodic
boundaries rule them out~\citep{ferziger2020computational}. Yet historically,
they have been indispensable for developing \emph{understanding}: much of what
we know about turbulence was discovered through Fourier analysis in idealized
periodic settings \citep{pope2001turbulent}. The insights gained there informed
the design of practical methods for far more complicated problems. We believe
this thesis occupies a similar position. The restriction to semi-linear PDEs on
periodic grids is a genuine limitation
(\cref{sec:outlook:limitations-and-questions}), but it is also what made the
mode-wise analysis tractable and the resulting insights precise. The findings on
superiority, progressive refinement, and benchmarking desiderata do not depend
on periodicity per se; they depend on having a setting clean enough to isolate
the relevant mechanisms. We expect these mechanisms to transfer to more complex
settings, even where the spectral analysis itself does not.

\subsection{The Receptive Field--Domain of Dependence Connection}
\label{sec:outlook:findings:receptive}

Across all experiments in APEBench, the receptive field emerged as one of the
most important factors linking architecture choice to PDE difficulty. For
problems with a clear domain of dependence, like hyperbolic PDEs, local
convolutional architectures consistently performed the best, provided their
receptive field covered the domain of dependence. This mirrors the CFL condition
from classical numerics: a finite difference stencil must span at least as many
grid cells as the physics propagates per time step, and the same holds for the
emulator. Unlike a numerical scheme, however, the emulator needs some headroom
beyond the strict domain of dependence, and unrolled training can help by
extending the temporal receptive field seen during optimization
(\cref{sec:apebench:studies}).

\subsection{Representational Capacity Limits}

In our experiments, neural emulators are typically capped at
$\approx 10^{-2}$ nRMSE accuracy, with the best cases reaching
$\approx 10^{-3}$. This is fundamentally different from numerical methods
that, if set up correctly, can converge to machine precision. However, the
comparison is subtler than it appears: just because a solver $\mathcal{P}_h$
has converged to machine precision does not mean the underlying PDE system
$\mathbb{P}$ is accurately represented. What matters is which types of errors
are present, whether they are discretization errors in $\mathcal{P}_h$,
modelling errors in $\mathbb{P}$, or learned approximation errors in
$f_\theta$. This distinction is exactly what underpinned the superiority study
(\cref{sec:relationship:superiority}).

The theoretical work of \citet{de2022generic} shows that neural operator
approximation error (for certain classes of PDEs) scales polynomially with the
number of parameters, suggesting that sufficient capacity should, in principle,
yield arbitrary accuracy. Our experiments and those of other benchmarking
efforts \citep{takamoto2022pdebench,gupta2022towards} suggest that this
theoretical limit is not reached in practice, likely due to a combination of
spectral biases, distribution shifts from autoregressive application, and
optimization limitations. Both PRDP and the superiority analysis leveraged this
accuracy gap. The network does not need the solver to be fully converged because
it cannot exploit that extra precision anyway.

\subsection{Autoregressive Generalization and Distribution Shifts}

On any sufficiently challenging problem, autoregressive emulators will
eventually diverge.
Errors
compound at each step, and no amount of training can fully prevent this for
general dynamics. However, the severity depends strongly on the type of problem.
For periodic or statistically stationary dynamics, quite long stable rollouts
are achievable
\citep{kohl2026benchmarking,pedersen2025thermalizer,jiang2025hierarchical}. For
problems that converge to a steady state in reasonable time, emulators may
achieve effectively infinite stability since the target itself stops changing.
For Hamiltonian systems, structure-preserving architectures that exploit
conservation laws offer a path to long-term stability
\citep{greydanus2019hnn,zhou2025hamiltonian}.
In the APEBench experiments, rollout error trajectories varied widely across
architectures and PDEs. The superiority analysis
(\cref{sec:relationship:superiority}) showed that emulators can even accumulate
errors more favorably than their training solver, suggesting that the learned
dynamics can have implicit stability properties not present in the numerical
scheme.

The truly challenging regime is non-periodic, non-stationary dynamics. Even if
we chose a functional form for $f_\theta$ that is capable of unconditional
stability (as classical implicit schemes are), the data-driven optimization is
unlikely to fit the parameters with the same precision as a symbolically derived
scheme. This raises the practical question: for how long does the emulator
\emph{need} to be accurate? In many deployment scenarios, a finite prediction
horizon is sufficient, and the rollout length should be matched to that horizon
rather than pushed to infinity.

\section{Limitations and Open Questions}
\label{sec:outlook:limitations-and-questions}

The theoretical analyses and systematic experiments in this thesis rely on
semi-linear PDEs with periodic boundary conditions on uniform Cartesian grids.
This restriction was
fundamental to the Fourier-spectral framework that underpins both the
superiority analysis and the APEBench solver, but it excludes important problem
classes with Dirichlet or Neumann boundaries and complex geometries
(\cref{sec:outlook:future:unstructured}). We further restricted ourselves to settings
in which the reference solver is fast. In these regimes, surrogates are not
strictly necessary, but the fast iteration cycles enabled thorough hypothesis
testing that we believe can inform experiments on more challenging problems.

All evaluations in this thesis use full-field error metrics (nRMSE between
predicted and reference states). For many surrogate applications, accuracy on
derived quantities is more important: spatial derivatives amplify high-frequency
errors (e.g., stress fields from displacement predictions), and surface
integrals introduce additional quadrature errors (e.g., lift and drag
coefficients). Full-field accuracy at $\approx 10^{-2}$ does not guarantee
that these derived quantities are converged.

Each emulator $f_\theta$ in our experiments is trained for a single solver
$\mathcal{P}_h$, i.e., one PDE system $\mathbb{P}$ with one set of
discretization choices. The emulators are not \emph{fundamental} in the sense
of generalizing across PDE systems or constitutive parameters, a direction
addressed by the foundation model outlook
(\cref{sec:outlook:future:foundation}). The architecture comparisons in APEBench
cover ConvNets, ResNets, dilated ResNets, FNOs, and UNets, but not
Transformer-based architectures, which have since gained prominence
(\cref{sec:emulators:neural:transformer}). We also did not investigate
physics-informed loss terms or curriculum strategies on the unrolling length $T$,
both of which are natural extensions of the $(T, B)$ taxonomy.

Several questions remain actionable with the existing APEBench infrastructure.
The optimal unrolling length for a given PDE and deployment context is not yet
understood; more generally, the interplay between training unrolling length and
testing rollout horizon remains poorly understood. Mitigating rollout
instability and distribution shifts for non-periodic, non-stationary problems
remains an open challenge. The effect of inaccurate gradients on training, and
more broadly how decoupled loss and gradient landscapes affect optimization
\citep{schnell2024stabilizing}, deserves further investigation in the
differentiable physics setting. Finally, APEBench enables systematic studies
along axes not yet explored: transfer learning across PDEs, data augmentation
strategies, architecture search, and scaling laws.

\section{Outlook}
\label{sec:outlook:future}

\subsection{Foundation Models for PDEs}
\label{sec:outlook:future:foundation}

A natural next step is to amortize the data generation and model training into a
pretraining stage that produces a universal emulator which can be finetuned to
specific problems. Several works have begun exploring this direction, including
DPOT \citep{hao2024dpot}, Poseidon \citep{herde2024poseidon}, MPP
\citep{mccabe2024multiple}, Disco \citep{morel2025disco}, and Walrus
\citep{mccabe2025walrus}. Such pretraining blurs the five-role distinction of
\cref{sec:relationship:roles}: role~(1) fans out across many heterogeneous
solvers, each at its own fidelity, while the semantics of a single evaluation
reference~(2) and baseline comparator~(5) become ambiguous once the emulator
is intended to cover a whole family of PDE systems rather than a single
$\mathbb{P}$.

The idea that underpinned APEBench of using procedurally generated data was only
possible because the solver was fast. It \emph{precomputed} all necessary data
prior to training. This has since been extended for wider parameter regimes with
the pretraining corpora of \citet{holzschuh2025pde} and \citet{holzschuh2026p3d}.
However, the procedural generation also opens the door for \emph{online} data
generation, where the solver produces training trajectories on the fly during
training, which we investigate with \citet{liu2026tadpole}. In an even more
general setting, there could be a feedback signal from the training loop to the
solver to allow for active learning~\citep{musekamp2024active}.

\subsection{Compute-Optimal Data Generation}
\label{sec:outlook:future:generation}

Generating large-scale training data for PDE emulators is expensive. For
realistic CAE settings, a single high-fidelity dataset can cost thousands of
CPU-hours \citep{ashton2024drivaer}. The findings of this thesis suggest that
much of this cost may be unnecessary. If emulators cannot exploit solver
accuracy beyond $\approx 10^{-2}$ at current model capacities
(\cref{sec:outlook:findings}), and if they can implicitly correct for certain
structured solver errors (\cref{sec:relationship:superiority}), then generating
all training data at full fidelity is wasteful. While PRDP exploits this by
progressively increasing solver fidelity \emph{during} training (since the
solver sits in the gradient loop), the same insight applies more directly to
data generation: one can fix a fidelity distribution \emph{before} training and
generate the dataset upfront, mixing trajectories from solvers at varying levels
of convergence. The emulator can then be conditioned on the fidelity level,
allowing it to learn how solver accuracy affects the dynamics.

This raises a question that, to our knowledge, has no answer yet: what is the
compute-optimal trade-off between solver fidelity, number of trajectories,
trajectory length, and model capacity?
\citet{subramanian2023foundation_models_sciml} provide initial scaling curves
for neural operators but do not address the solver-side compute budget.
Characterizing this trade-off, analogous to how \citet{hoffmann2022training}
characterized the data-compute-size balance for language models, is where the
PRDP insights uniquely contribute. We are pursuing this direction in ongoing
work \citep{setinek2026compute}.

\subsection{Beyond Structured Grids}
\label{sec:outlook:future:unstructured}

How well do the findings of this thesis on simple PDEs and structured grids
translate to more complex physics and unstructured domains? There is reason to
believe that they do, since complicated physical systems are composed of the
same canonical effects that APEBench covers. But the field of deep learning has
shown repeatedly that surprising emergent effects appear at
scale~\citep{wei2022emergent}, and the spectral analysis framework that
underpins much of this thesis does not straightforwardly generalize to
non-periodic, non-uniform settings.

Even without unstructured solvers, the fast spectral solvers can bridge
structured and unstructured worlds. Because spectral interpolation can map
solution fields to arbitrary point distributions with negligible interpolation
error, structured solver data can be projected onto unstructured
representations for training~\citep{moeed2025morphed_thesis}.

Extending the APEBench methodology fully to unstructured grids requires fast
differentiable solvers on non-Cartesian domains. One path is a finite volume
extension to Picardax (\cref{sec:foundations:implementation:picardax}). However,
the ML frameworks that enabled fast prototyping of Exponax and Picardax are
optimized for dense tensor operations. Unstructured solvers map less naturally
onto these primitives, and frameworks like NVIDIA Warp~\citep{warp2022} or lower-level GPU
programming may be necessary, though this sacrifices the seamless autodiff and
vmap integration that made our JAX-based approach productive.

\subsection{Modern Differentiable Solvers and the Automation of Scientific Software}
\label{sec:outlook:future:ecosystem}

Part of the success of APEBench and its spectral solvers is that they bring
established numerical methods to modern accelerators and outfit them with
automatic vectorization and automatic differentiation. The Exponax and Picardax
solvers developed in this thesis are concrete examples: established ETDRK and
finite difference methods, reimplemented in JAX, thereby gaining autodiff, vmap,
and accelerator support. This not only enables fast data generation and
neural-hybrid models but could also benefit classical workflows in which inverse
problems require many costly forward evaluations that could be accelerated by
gradient information.

This vision extends well beyond JAX. Autodiff engines like Enzyme
\citep{moses2020instead} operate at the compiler level, making existing C,
C++, and Fortran codes differentiable without rewrites. Tesseract
\citep{haefner2025tesseract} enables pipeline-level autodiff across
framework boundaries.

A more radical perspective questions how much we need neural surrogates in the
first place. Neural surrogates can be viewed as an indirect way to port solvers
from CPUs to GPUs: sample the CPU solver representatively and then use efficient
ML primitives for fast inference. Modern software engineering agents could
automate the translation of existing solvers to accelerators directly, and could
even exhaustively try different simulation techniques (FVM, FEM, SPH) to find
which is optimal for a given scenario, a task that has traditionally required
extensive domain expertise. This is a well-constrained problem with a clear
objective: the existing CPU solver provides a reference to validate against, and
the goal is to make it orders of magnitude faster while adding modern features
like autodiff compatibility and vectorization. Early evidence supports this
direction: CodePDE \citep{li2026codepde} frames PDE solving as a code generation
task, and an agentic pipeline fully automated the translation of Fortran kernels
to the GPU \citep{gupta2025legacy}. Whether such tools can scale to full
simulation pipelines remains to be seen, but the trajectory is clear.

\section{Conclusion}
\label{sec:outlook:conclusion}

This thesis began with a simple observation: neural emulators of PDEs are
trained on data produced by numerical solvers, yet the two paradigms are rarely
examined in the same analytical frame. The work presented here shows employing a
common language reveals a deeper relationship.

The overarching lesson is that numerical solvers and neural emulators are more
alike than different. Once this correspondence is recognized, phenomena that
initially seem surprising become explainable: an emulator can exceed its
training data because its inductive bias implicitly regularizes solver errors, a
hybrid solver need not be fully converged because the network cannot exploit the
extra precision, and architectures that structurally resemble the numerical
methods suited to a given PDE type tend to perform best.

Perhaps the most far-reaching implication is methodological. The restriction to
semi-linear PDEs on periodic grids, which is a genuine limitation of this work,
was also its greatest asset. The controlled settings of APEBench and the
Fourier-spectral analysis enabled precise isolation of mechanisms that would be
confounded in more realistic scenarios. The superiority phenomenon, the
progressive refinement savings, and the benchmarking desiderata are the
consequences of the fundamental interplay between discretization error,
inductive bias, and optimization dynamics. There is good reason to expect them
to persist, in modified form, as the field moves toward unstructured meshes,
complex boundary conditions, and foundation models that span families of PDEs.

The title of this thesis was chosen to emphasize that the flow of insights is
not one-directional. Neural emulators benefit from the decades of knowledge
encoded in classical numerical methods, and classical methods gain new relevance
when their structure is recognized inside learned architectures. As both
paradigms continue to evolve, with solvers migrating to autodiff-compatible
implementations potentially using mixed-precision on modern accelerators and
emulators growing toward foundation models with increasing physical fidelity,
the feedback loop between them is likely to tighten rather than dissolve. The
work presented here offers one language in which that conversation can be
conducted precisely.

\chapter{Summary of Publications}
\label{ch:publications_summary}

This chapter summarizes the three publications that constitute this thesis, in
order of publication. For each paper, we state the main contributions, highlight
key experiments and findings, and clarify the author's contributions. The full
papers are reproduced in the appendix.

\section{APEBench: A Benchmark for Autoregressive Neural Emulators of PDEs}
\label{sec:summary:apebench}

Published at NeurIPS 2024 Datasets and Benchmarks
\citep{koehler2024apebench}. The full paper is reproduced in
\cref{paper:apebench}.

\subsection{Main Contributions}
\label{sec:summary:apebench:contributions}

APEBench is a benchmarking suite for autoregressive neural PDE emulators built
around a tightly integrated differentiable solver. Its core component is
Exponax (\cref{sec:foundations:implementation:exponax}), a fast pseudo-spectral
solver for semi-linear PDEs in JAX that serves both as data generator and as
differentiable training component. Alongside the solver, we developed PDEquinox,
a collection of dimension- and boundary condition-agnostic neural architectures,
and Trainax, a library that implements the $(T, B)$ training taxonomy
(\cref{sec:apebench:framework:taxonomy}) including differentiable physics
configurations like diverted chain that no other benchmark supports.

The benchmark covers 46 PDE scenarios across 1D, 2D, and 3D spanning linear,
nonlinear, and reaction-diffusion dynamics
(\cref{sec:apebench:framework:configurations}). To enable systematic comparison
across this space, we introduced normalized difficulty identifiers $\gamma_j$
(\cref{sec:relationship:benchmarking:normalized}) that compress the parameter
space into a minimal exchange protocol directly linked to CFL-like stability
criteria of classical numerical methods.

\subsection{Key Experiments and Findings}
\label{sec:summary:apebench:experiments}

The paper opens with a motivational experiment on 1D advection that bridges
classical numerics and machine learning. By scaling the difficulty $\gamma_1$,
we show how rollout metrics, training methodology, and the normalized
identifiers interact, and we establish a direct relationship between an
architecture's receptive field and the CFL number of the PDE: emulators fail
when their receptive field cannot cover the domain of dependence.

A broad comparison of five architectures (ConvNet, ResNet, UNet, Dilated
ResNet, FNO) across all 46 PDEs reveals that the best architecture for a given
problem tends to match the functional form of the corresponding best numerical
method. Local convolutional architectures excel on reaction-diffusion problems,
the FNO on problems with global spatial structure like Navier--Stokes, and the
ResNet offers the most consistent performance across all categories.

The paper further demonstrates the diverted chain training configuration, which
combines the rollout stability of unrolled training with the short-term accuracy
of one-step supervision. This configuration is uniquely enabled by the
differentiable solver and had not been studied in prior work. Large-scale
ablations on training data size, optimization duration, and parameter count
confirm the robustness of findings, with all results validated over multiple
seeds (50 in 1D, 20 in 2D and 3D).

\subsection{Impact and Adoption}
\label{sec:summary:apebench:impact}

Since publication, the Exponax solver has been translated to PyTorch
\citep{liu_2025_15350210} and is popularly used for data generation. It has
been scraped for the pretraining datasets of PDE-Transformer
\citep{holzschuh2025pde} and P3D \citep{holzschuh2026p3d}. The APEBench
framework also served as infrastructure for supervised theses on foundation
models \citep{jacob2024foundation_thesis} and architecture comparison
\citep{moeed2025evaluating_thesis}.

The full experimental suite of the paper costs approximately 900~GPU-hours on
8$\times$~RTX 2080~Ti, enabled by procedural generation and seed parallelism.
For the depth of insights obtained, this is remarkably cheap and supports many
follow-up studies. More broadly, working on simple, well-understood problems has
proven both empirically and educationally valuable
\citep{greydanus2024scaling,prince2023understanding}.

\subsection{Contribution Statement}
\label{sec:summary:apebench:statement}

I implemented the pseudo-spectral solver Exponax, the emulator architecture
library PDEquinox, the training methodology library Trainax, and the
benchmarking framework APEBench. I devised and carried out all experiments.
Simon Niedermayr wrote the web-GPU-based volume renderer and contributed to the
manuscript. R\"udiger Westermann provided feedback on the manuscript. Nils
Thuerey supervised the work and contributed to the manuscript.

\section{Progressively Refined Differentiable Physics}
\label{sec:summary:prdp}

Published at ICLR 2025 \citep{bhatia2025prdp}. This work originated from the
Master's thesis of Kanishk Bhatia \citep{bhatia2024unrolling_thesis}, which I
co-supervised. The full paper is reproduced in \cref{paper:prdp}.

\subsection{Algorithm Description}
\label{sec:summary:prdp:algorithm}

Differentiable physics training places iterative solvers inside the
computational graph, and differentiating through them becomes costly as the
number of iterations grows. PRDP addresses this by viewing the setup as a
bilevel optimization problem (\cref{sec:relationship:fidelity:bilevel}): the
outer problem optimizes the network parameters while the inner problem is the
iterative solver reaching convergence. The key insight is that full solver
convergence is unnecessary for accurate gradient computation, giving rise to two
types of savings. \emph{Progressive refinement} (PR) savings come from starting
training with few solver iterations $K \ll K_\epsilon$ and increasing $K$ only
as the outer optimization matures. \emph{Incomplete convergence} (IC) savings
come from the fact that the network reaches full accuracy at a refinement level
$K_{\max}$ that is significantly below the convergence threshold $K_\epsilon$.

PRDP detects $K_{\max}$ automatically by monitoring a validation metric during
training. When performance plateaus at the current refinement level, the
algorithm increases $K$. When a refinement step yields no measurable
improvement, it concludes that $K_{\max}$ has been reached and stops refining.
The approach is applicable to both unrolled differentiation (differentiating
through the iterations) and implicit differentiation (differentiating over the
converged solve); see \cref{sec:foundations:iterative:differentiation}.

\subsection{Results and Savings Achieved}
\label{sec:summary:prdp:results}

Across all tested scenarios, the final network accuracy under PRDP is
statistically indistinguishable from training with fully converged physics. In
the most challenging experiment, emulating the 2D Navier--Stokes equations with
a correction-based neural-hybrid solver, PRDP reduces training time by 62\%.
Across experiments, total solver iterations are reduced by up to 86\%.
% The IC savings also carry over to inference: deploying the learned model with
% the solver at $K_{\max}$ rather than $K_\epsilon$ reduces the inference cost
% proportionally.
The approach is validated on Poisson inverse problems
(1-parameter and 3-parameter), heat equation emulators in 1D and 2D, Burgers
equation emulators, and the Navier--Stokes neural-hybrid setting.

\subsection{Applicability and Limitations}
\label{sec:summary:prdp:applicability}

PRDP requires an iterative solver component in the computational graph and is
therefore not applicable to explicit solvers like ETDRK or to purely
data-driven training without a solver in the loop. Whether IC savings are
observed depends on the problem structure: in network-limited settings where
the network's representational capacity is the bottleneck, both PR and IC
savings occur. In double-convex settings like simple Poisson inverse problems,
where the solver accuracy directly limits the outer objective, only PR savings
are observed. The PR savings are universal and always present, consistent with
the bilevel optimization theory of \citet{pedregosa2016hyperparameter}. The
current implementation focuses on iterative linear solvers for sparsely
discretized differential operators; extension to nonlinear solvers (e.g.,
Newton) is straightforward in principle but not yet validated.

\subsection{Contribution Statement}
\label{sec:summary:prdp:statement}

This work originated from the Master's thesis of Kanishk Bhatia, which I
co-supervised. I devised and implemented the finite difference-based PDE solvers
(Picardax), including unrollable linear solvers with configurable convergence
criteria, and contributed theoretical derivations on different ways to
differentiate through linear solvers. Kanishk Bhatia investigated the bilevel
optimization perspective, developed the PRDP algorithm (plateau detection and
refinement scheduling), and carried out the main experiments. We both
contributed equally to the writing of the manuscript. Nils Thuerey supervised
the work and contributed to the manuscript.

\section{Neural Emulator Superiority}
\label{sec:summary:superiority}

Published at NeurIPS 2025 \citep{koehler2025neural}. The full paper is
reproduced in \cref{paper:superiority}.

\subsection{Theoretical Results}
\label{sec:summary:superiority:theory}

The theoretical analysis builds on the Fourier multiplier framework of
\cref{sec:foundations:spectral}. For each problem, we fit a simple linear
emulator ansatz whose functional form matches an explicit numerical scheme (a
first-order upwind form for advection, an FTCS form for diffusion, and a
second-order Laplacian form for Poisson) to data from an implicit or
unconverged solver. We derive the learned amplification factors in closed form
and show that the ansatz's inductive bias acts as implicit regularization: even
when trained on low-fidelity data, the learned multiplier is steered toward the
analytical solution. For advection and diffusion, we prove that $\xi < 1$ for a
considerable range of modes with $\phi > \psi$ (\emph{forward superiority}),
meaning the emulator outperforms its training solver in higher-frequency modes
where that solver's errors are largest. For the Poisson equation with an
unconverged Richardson solver, we find \emph{backward superiority}: the emulator
excels for $\phi < \psi$, i.e., in the low-frequency modes where the iterative
solver converges slowest.

We further distinguish two forms of the phenomenon. \emph{State-space
superiority} ($\xi^{[1]} < 1$) occurs already at the first time step when the
emulator is tested on initial conditions from a different distribution than used
for training. \emph{Autoregressive superiority} ($\xi^{[t]} < 1$ for $t \geq
2$) emerges during rollout from more favorable error accumulation, even when
training and test distributions are identical.

\subsection{Empirical Validation}
\label{sec:summary:superiority:empirical}

We validate the theoretical predictions with nonlinear neural architectures
(ConvNets, ResNets, FNOs, UNets, and Transformers) on both linear (diffusion,
advection) and nonlinear (Burgers) PDEs. Autoregressive superiority turns out
to be nearly universal: all tested architectures achieve it. State-space
superiority, on the other hand, is architecture-dependent. Architectures with
local receptive fields (ConvNet, ResNet) achieve it, while global architectures
(FNO, Transformer) do not under the same distribution shift. This is consistent
with the theoretical analysis: local architectures share the structural
inductive bias of finite difference schemes that drives the superiority
mechanism. Training on single-mode data produces emulators with stronger
state-space superiority than training on multi-mode data, because the
distributional mismatch at test time is larger.

\subsection{Implications for the Field}
\label{sec:summary:superiority:implications}

The superiority phenomenon challenges the common assumption that training data
fidelity is a hard ceiling for emulator accuracy. The student can exceed the
teacher. This has a direct consequence for benchmarking: evaluating an emulator
against its own training solver conflates the two error sources, and a benchmark
that does so cannot detect superiority. Emulator and solver error must be
separated by evaluating against high-fidelity references, which is precisely the
design principle underlying APEBench's use of spectral reference solutions
(\cref{sec:relationship:benchmarking:reference}). More broadly, the findings
show that the solver-emulator relationship is richer than a simple
student-teacher dynamic: the emulator's inductive biases interact with the
solver's error structure in ways that can be both characterized theoretically
and exploited in practice.

\subsection{Contribution Statement}
\label{sec:summary:superiority:statement}

I performed all theoretical derivations, conducted all experiments, and wrote
the manuscript. Nils Thuerey supervised the work and contributed to the
manuscript.

% Bibliography
\printbibliography[heading=bibintoc]

% =============================================================================
% APPENDICES
% =============================================================================
\appendix

\chapter{APEBench: Full Paper}
\label{paper:apebench}

\textbf{Title}
\\
APEBench: A Benchmark for Autoregressive Neural Emulators of PDEs
\\
\textbf{Abstract}
\\
We introduce the \textbf{A}utoregressive \textbf{P}DE \textbf{E}mulator
Benchmark (APEBench),  a comprehensive benchmark suite to evaluate
autoregressive neural emulators for solving partial differential equations.
APEBench is based on JAX and provides a seamlessly integrated differentiable
simulation framework employing efficient pseudo-spectral methods, enabling 46
distinct PDEs across 1D, 2D, and 3D. Facilitating systematic analysis and
comparison of learned emulators, we propose a novel taxonomy for unrolled
training and introduce a unique identifier for PDE dynamics that directly
relates to the stability criteria of classical numerical methods. APEBench
enables the evaluation of diverse neural architectures, and unlike existing
benchmarks, its tight integration of the solver enables support for
differentiable physics training and neural-hybrid emulators. Moreover, APEBench
emphasizes rollout metrics to understand temporal generalization, providing
insights into the long-term behavior of emulating PDE dynamics. In several
experiments, we highlight the similarities between neural emulators and
numerical simulators. The code is available at
\url{https://github.com/tum-pbs/apebench} and APEBench can be installed via
\verb|pip install apebench|.
\\
\textbf{Copyright Notice}
The copyright of this paper remains with the authors. The version included here
is reproduced with permission of the co-authors and in accordance with NeurIPS's
publication policy.

\includepdf[pages=-, pagecommand={\thispagestyle{plain}}]{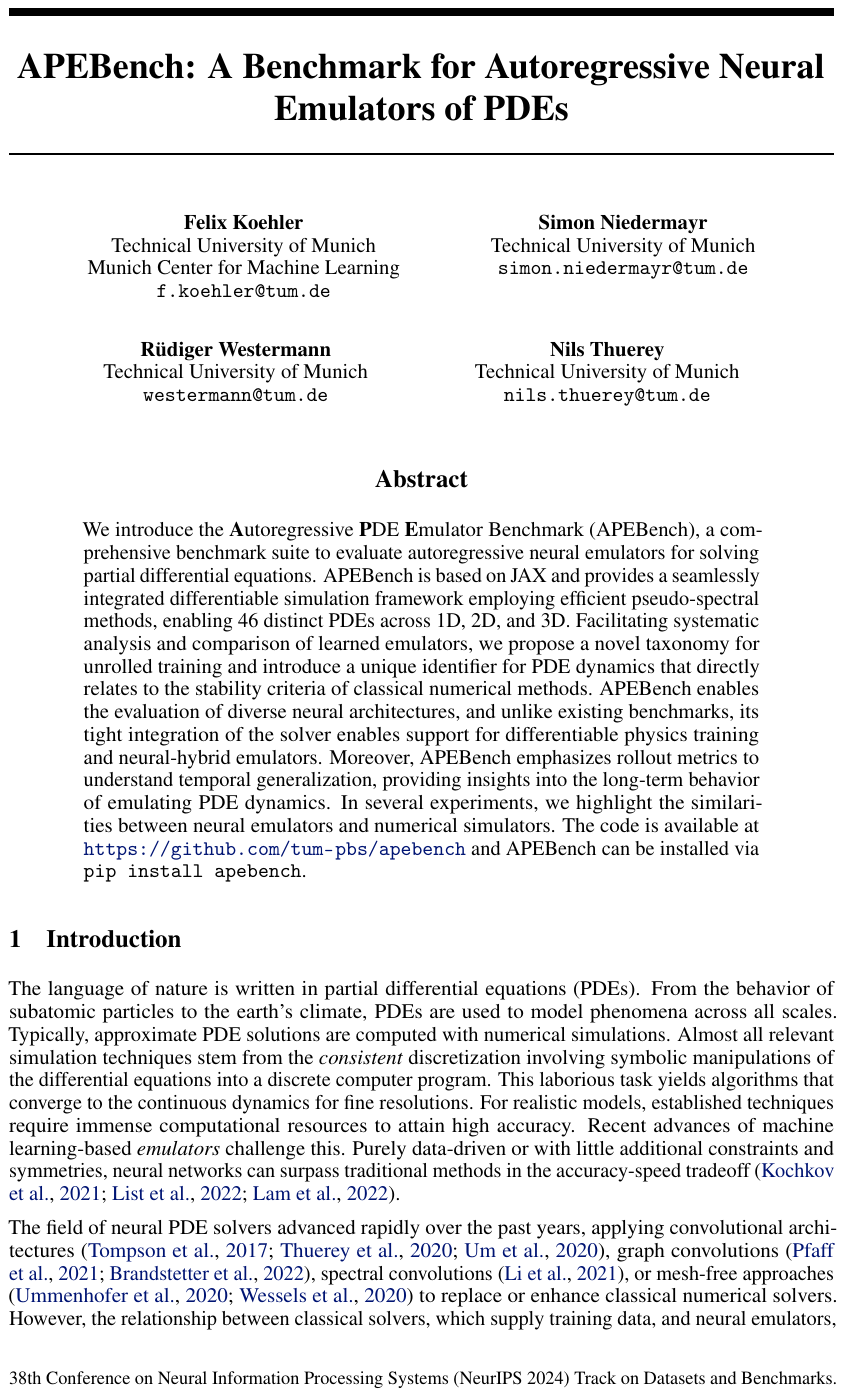}

\chapter{PRDP: Full Paper}
\label{paper:prdp}

\textbf{Title}
PRDP: Progressively Refined Differentiable Physics
\\
\textbf{Abstract}
The physics solvers employed for neural network training are primarily
iterative, and hence, differentiating through them introduces a severe
computational burden as iterations grow large. Inspired by works in bilevel
optimization, we show that full accuracy of the network is achievable through
physics significantly coarser than fully converged solvers. We propose
\emph{progressively refined differentiable physics} (PRDP), an approach that
identifies the level of physics refinement sufficient for full training
accuracy. By beginning with coarse physics, adaptively refining it during
training, and stopping refinement at the level adequate for training, it enables
significant compute savings without sacrificing network accuracy. Our focus is
on differentiating iterative linear solvers for sparsely discretized
differential operators, which are fundamental to scientific computing. PRDP is
applicable to both unrolled and implicit differentiation. We validate its
performance on a variety of learning scenarios involving differentiable physics
solvers such as inverse problems, autoregressive neural emulators, and
correction-based neural-hybrid solvers. In the challenging example of emulating
the Navier-Stokes equations, we reduce training time by 62\%. \\
\textbf{Copyright Notice}
The copyright of this paper remains with the authors. The version included here
is reproduced with permission of the co-authors and in accordance with ICLR's
publication policy.

\includepdf[pages=-, pagecommand={\thispagestyle{plain}}]{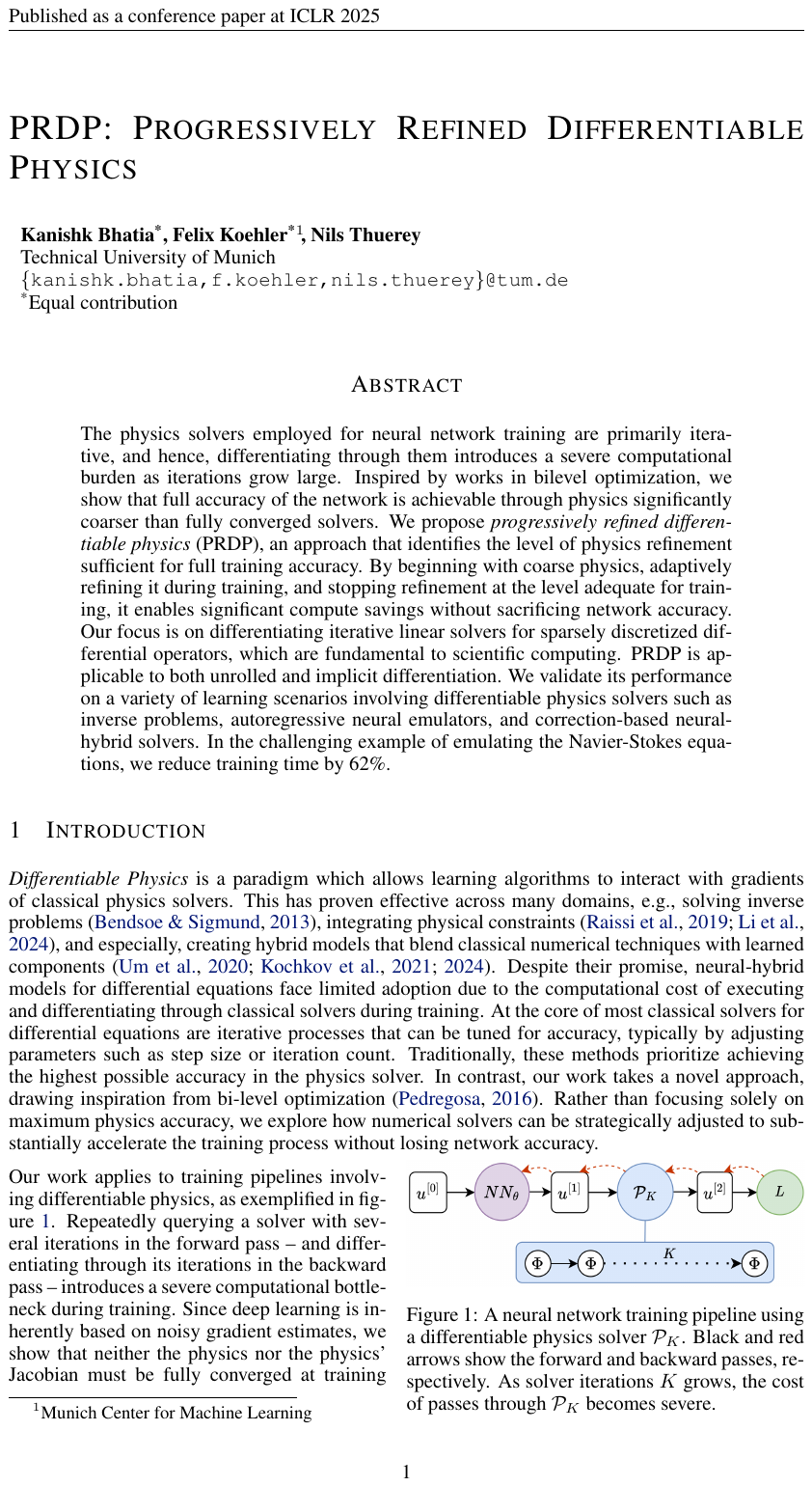}

\chapter{Emulator Superiority: Full Paper}
\label{paper:superiority}

\textbf{Title}
Neural Emulator Superiority: When Machine Learning for PDEs Surpasses its
Training Data
\\
\textbf{Abstract}
Neural operators or emulators for PDEs trained on data from numerical solvers
are conventionally assumed to be limited by their training data's fidelity. We
challenge this assumption by identifying "emulator superiority," where neural
networks trained purely on low-fidelity solver data can achieve higher accuracy
than those solvers when evaluated against a higher-fidelity reference. Our
theoretical analysis reveals how the interplay between emulator inductive
biases, training objectives, and numerical error characteristics enables
superior performance during multi-step rollouts. We empirically validate this
finding across different PDEs using standard neural architectures, demonstrating
that emulators can implicitly learn dynamics that are more regularized or
exhibit more favorable error accumulation properties than their training data,
potentially surpassing training data limitations and mitigating numerical
artifacts. This work prompts a re-evaluation of emulator benchmarking,
suggesting neural emulators might achieve greater physical fidelity than their
training source within specific operational regimes.

Project Page:
\href{https://tum-pbs.github.io/emulator-superiority}{tum-pbs.github.io/emulator-superiority}
\\
\textbf{Copyright Notice}
The copyright of this paper remains with the authors. The version included here
is reproduced with permission of the co-authors and in accordance with NeurIPS's
publication policy.

\includepdf[pages=-, pagecommand={\thispagestyle{plain}}]{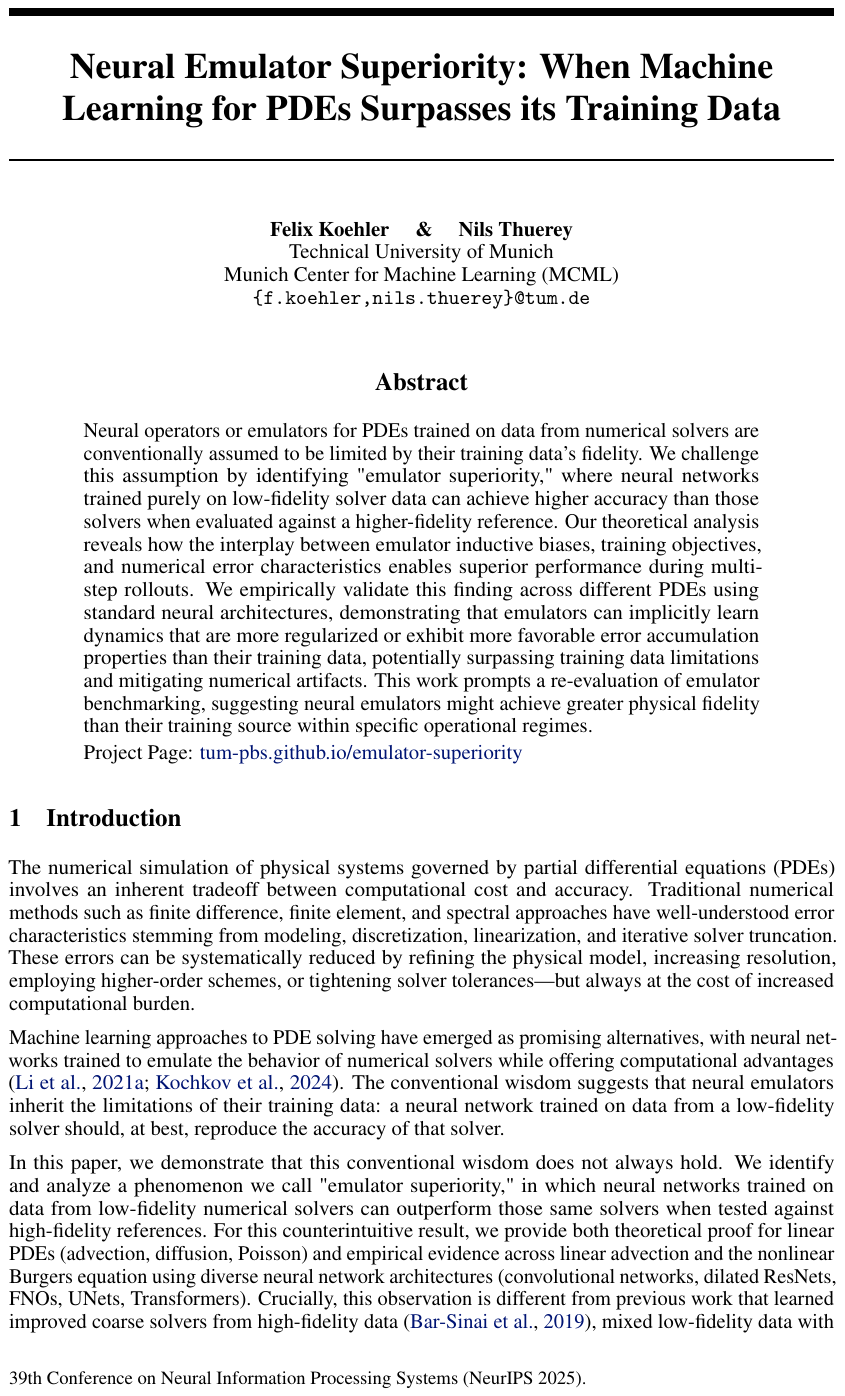}

\end{document}